\documentclass{ecai}  
\pdfoutput=1
\usepackage{graphicx}
\usepackage{latexsym}

\usepackage{hyperref}
\hypersetup{
    colorlinks=true,
  }

\usepackage{subcaption}
\usepackage{amsmath,amssymb,amsfonts}
\usepackage{bbm}
\graphicspath{ {./figures/} }

\usepackage{amsthm}
\newtheorem{theorem}{Theorem}
\newtheorem{corollary}[theorem]{Corollary}

\newcommand{\dfdx}{f^s}
\newcommand{\xc}{\mathbf{x}_c}

\newcommand{\xb}{\mathbf{x}}

\newcommand{\paragraphb}{\vspace{-0.25cm}\paragraph}

\begin{document}

\begin{frontmatter}

\title{RHALE: Robust and Heterogeneity-aware Accumulated Local Effects}

\author[A, B]{\fnms{Vasilis}~\snm{Gkolemis}} 
\author[B]{\fnms{Theodore}~\snm{Dalamagas}} 
\author[C]{\fnms{Eirini}~\snm{Ntoutsi}} 
\author[A]{\fnms{Christos}~\snm{Diou}} 

\address[A]{Harokopio University of Athens}
\address[B]{ATHENA RC}
\address[C]{Universitat der Bundeswehr Munchen}

\begin{abstract}
Accumulated Local Effects (ALE) is a widely-used explainability method for isolating the average effect of a feature on the output, because it handles cases with correlated features well. However, it has two limitations. First, it does not quantify the deviation of instance-level (local) effects from the average (global) effect, known as heterogeneity. Second, for estimating the average effect, it partitions the feature domain into user-defined, fixed-sized bins, where different bin sizes may lead to inconsistent ALE estimations. To address these limitations, we propose Robust and Heterogeneity-aware ALE (RHALE). RHALE quantifies the heterogeneity by considering the standard deviation of the local effects and automatically determines an optimal variable-size bin-splitting. In this paper, we prove that to achieve an unbiased approximation of the standard deviation of local effects within each bin, bin splitting must follow a set of sufficient conditions. Based on these conditions, we propose an algorithm that automatically determines the optimal partitioning, balancing the estimation bias and variance. Through evaluations on synthetic and real datasets, we demonstrate the superiority of RHALE compared to other methods, including the advantages of automatic bin splitting, especially in cases with correlated features.
\end{abstract}

\end{frontmatter}

\section{Introduction}
\label{sec:intro}

\begin{figure*}
    \centering
\begin{subfigure}{.33\textwidth}
  \centering
  \includegraphics[width=1\textwidth]{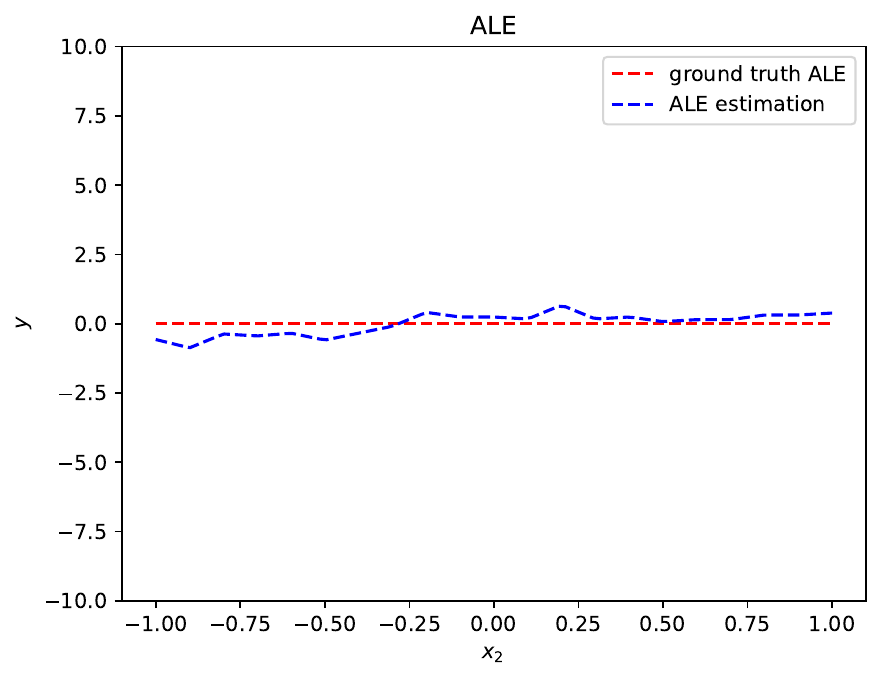}
  \caption{\(x_2\) ALE plot (20 bins)}
  \label{fig:concept-figure-0-subfig-1}
\end{subfigure}%
\begin{subfigure}{.33\textwidth}
  \centering
  \includegraphics[width=1\textwidth]{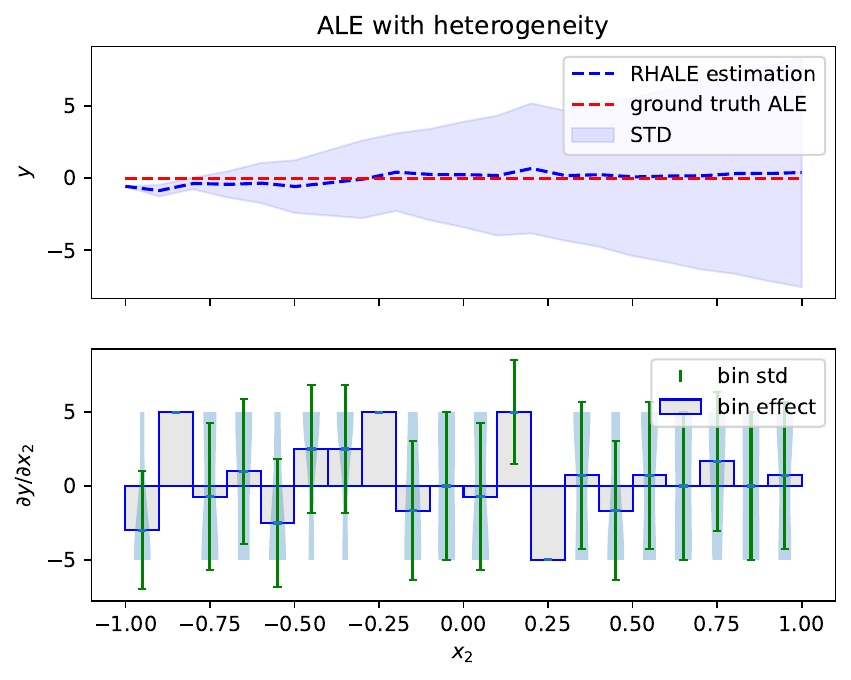}
  \caption{\(x_2\) ALE + heterogeneity (20 bins)}
  \label{fig:concept-figure-0-subfig-2}
\end{subfigure}
\begin{subfigure}{.33\textwidth}
  \centering
  \includegraphics[width=1\textwidth]{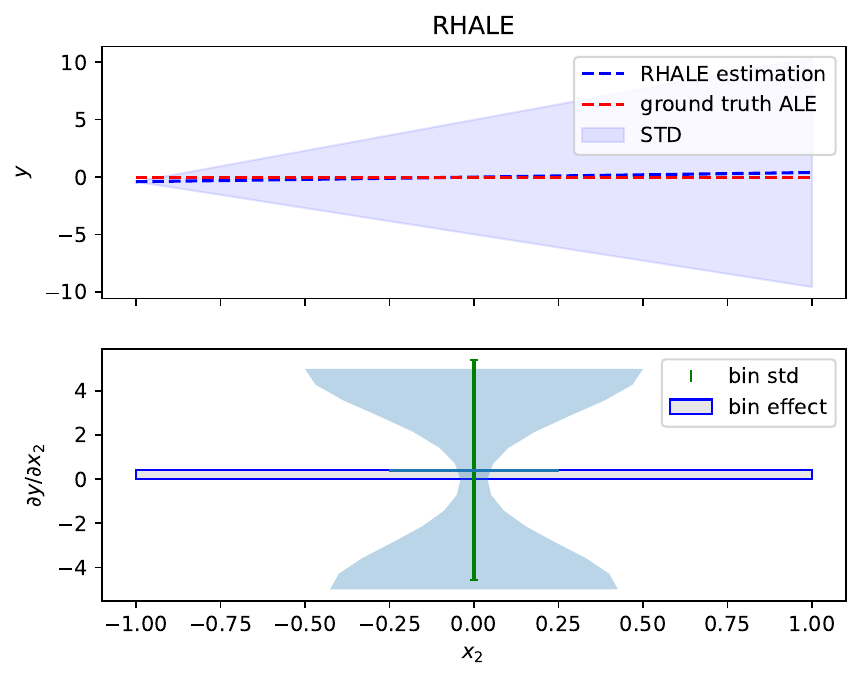}
  \caption{\(x_2\) RHALE (auto-binning)}
  \label{fig:concept-figure-0-subfig-3}
\end{subfigure}
\caption{Feature effect for \(x_2\) on the simple example of Eq.~(\ref{eq:concept-example-0}); (a) ALE incorrectly suggests that \(X_2\) does not relate to \(Y\), (b) ALE with heterogeneity using \(K=20\) fixed-size bins leads to significant approximation errors, (c) RHALE accurately estimates both the main effect and the heterogeneity, indicating that the zero average effect comes from opposite groups of instance-level effects.}
\label{fig:concept-figure-0}
\end{figure*}

\begin{figure}
  \centering
\begin{subfigure}{.23\textwidth}
  \centering
  \includegraphics[width=1\textwidth]{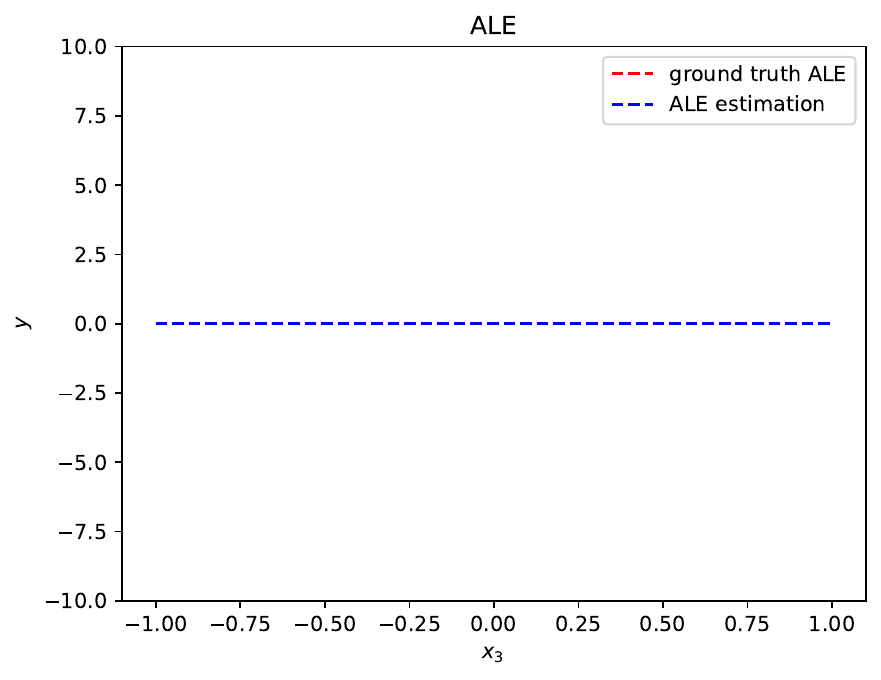}
  \caption{\(x_3\) ALE plot (20 bins)}
  \label{fig:concept-figure-1-subfig-0}
\end{subfigure}
\begin{subfigure}{.23\textwidth}
  \centering
  \includegraphics[width=1\textwidth]{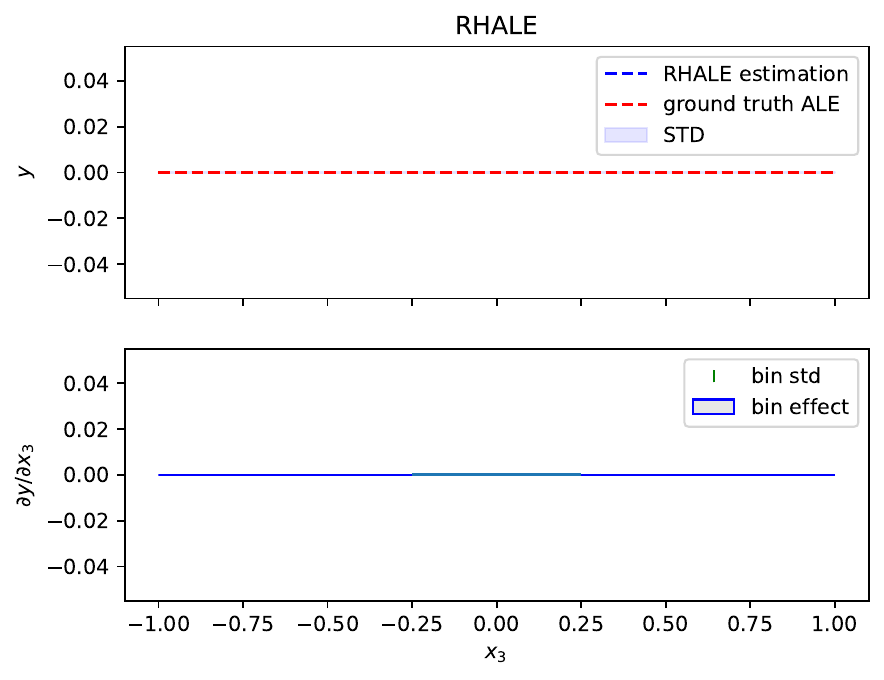}
  \caption{\(x_3\) RHALE plot (auto-binning)}
  \label{fig:concept-figure-1-subfig-1}
\end{subfigure}
\caption{Feature effect for \(x_3\) on the simple example of Eq. (1). ALE plot suggests that \(X_3\) does not relate to \(Y\). However, as seen in Figure~\ref{fig:concept-figure-0-subfig-1}, this interpretation can be misleading. Only after noticing the zero heterogeneity (\(\mathtt{STD}\) and bin-std are zero) of RHALE plot in (b), we can confirm this claim.}
  \label{fig:ale-with-heter}
\end{figure}

Recently, Machine Learning (ML) has been adopted across multiple areas
of human activity, including mission-critical domains such as
healthcare and finance. In such high-stakes areas, it is important to
accompany predictions with meaningful explanations~\cite{wiens2019no,
  freiesleben2022scientific}.  For this reason, there is an increased
interest in Explainable AI (XAI)~\cite{ribeiro2016should,
  koh2017understanding}. XAI literature distinguishes between local
and global methods~\cite{molnar2020interpretable}. Local methods
provide instance-level explanations~\cite{casalicchio2019visualizing},
i.e., explain the prediction for a specific input, whereas global
methods explain the entire model behavior~\cite{kim2016examples}. Most
of the times, global methods aggregate the instance-level explanations
into a single interpretable outcome, usually a number or a plot.

Feature Effect (FE)~\cite{Gromping2020MAEP} is a class of global
explainability methods that quantify the average (across all
instances) partial relationship between one feature and the output.
The most popular FE methods are \emph{Partial Dependence Plots}
(PDP)~\cite{friedman2001greedy} and \emph{Accumulated Local Effects}
(ALE)~\cite{apley2020visualizing}. PDPs have been
criticized~\cite{baniecki2021fooling, molnar2022, molnar2022general} for providing
misleading explanations in problems with highly correlated features,
making ALE the only reliable solution in such cases. Nevertheless, ALE
has two significant limitations. Firstly, the way ALE formulates the
FE (\textit{ALE definition} of Eq.~(\ref{eq:ALE})), does not take into
account the heterogeneity of instance-level effects, a quantity that
is crucial for a complete interpretation of the average
effect. Secondly, the way ALE estimates the FE from the instances of
the training set (\textit{ALE approximation} at
Eq.~(\ref{eq:ALE_accumulated_mean_est})) relies on a user-defined
binning process that often results in poor estimations.  Therefore,
this paper presents RHALE (Robust and Heterogeneity-aware ALE), a FE
method build on-top of ALE that overcomes these issues. To better
understand the advantages of RHALE over ALE, consider the following
example, which was first introduced in~\cite{goldstein2015peeking}:
\begin{equation}
  \label{eq:concept-example-0}
  \begin{aligned}
  Y &= 0.2X_1 - 5X_2 + 10X_2 \mathbbm{1}_{X_3 > 0} + \mathcal{E}\\
  \mathcal{E} &\stackrel{i.i.d.}{\sim} \mathcal{N}(0,1), \quad X_1, X_2, X_3 \stackrel{i.i.d.}{\sim} \mathcal{U}(-1, 1)
  \end{aligned}
\end{equation}
where we draw \(N=100\) samples, i.e,
\(\mathcal{D} = \{(x^i, y^i)\}_{i=1}^{N}\). Given the knowledge of
Eq.~\eqref{eq:concept-example-0}, the FE of \(X_3\) is zero because
the term \(10X_2\mathbbm{1}_{X_3 > 0}\), where \(X_3\) appears, is
part of the effect of \(X_2\).  In contrast, \(X_2\) relates to \(Y\)
in two opposite ways, as \(-5X_2\) when \(X_3 < 0\) and as \(5 X_2\)
otherwise. Therefore, the zero average effect of \(X_2\) after
aggregating the two opposites effects, should not erroneously imply
that \(X_2\) does not affect \(Y\). However, as demonstrated in
Figure~\ref{fig:concept-figure-0-subfig-1} (for \(X_2\)) and
Figure~\ref{fig:concept-figure-1-subfig-0} (for \(X_3\)) \textit{ALE
  definition} erroneously indicates that both variables are not
associated with the output. This phenomenon, known as aggregation
bias~\cite{mehrabi2021survey, herbinger2022repid}, is a common issue
of global XAI methods.

RHALE addresses this issue by quantifying the heterogeneity based on
the standard deviation of the underlying instance-level effects. As
shown in Figure~\ref{fig:concept-figure-0-subfig-3}, although the
average \(X_2\) effect is zero, the presence of two opposing groups of
instance-level effects, namely, \(5X_2\) and \(-5X_2\), is revealed by
both (a) the shaded area in the top subfigure (the limits of the
shaded area are the lines \(5X_2\) and \(-5X_2\)) and (b) the violin
plots in the bottom subfigure (the distribution of the instance-level
effects has most of its mass at \(-5\) and \(5\)). In contrast, in
Figure~\ref{fig:concept-figure-1-subfig-1}, the zero heterogeneity
states that \(X_3\) is indeed not related to the output.

The second limitation is that \textit{ALE approximation} requires an
initial step where the feature axis is partitioned in \(K\)
non-overlapping fixed-size bins. Afterwards, an average effect
(bin-effect) is computed inside each bin, and ALE plot is the
aggregation of the bin effects. Since there is no clear indication of
an appropriate value for \(K\), users often rely on heuristics, such
as ensuring that each bin contains on average at least \(\tau\)
instances on average. In the example above, for \(\tau=5\), then
\(K = 20\), which, as we show in
Figure~\ref{fig:concept-figure-0-subfig-2}, results in significant
approximation errors. Specifically, the bin-effects and bin-std values
deviate significantly from their ground-truth, which are 0 and 5,
respectively. To overcome this limitation, RHALE \textit{automatically
  determines the optimal sequence of variable-size bins}. For example,
in Figure~\ref{fig:concept-figure-0-subfig-3}, RHALE automatically
finds that it is optimal to create a single bin, which leads to a good
approximation of both the average effect and the heterogeneity. The
optimal bin splitting depends on the underlying instance-level effects
(Section~\ref{subsec:bin-spliting}). Essentially, a wide bin reduces
the variance of the estimation by increasing the samples inside the bin
(better Monte Carlo approximation) but it may also introduce bias in
the estimation. Using this insight, we formulate an optimization
problem and propose an algorithm that minimizes the bias while
ensuring that each bin has a sufficient number of samples.
The main contributions of this paper are:

\begin{itemize}
\item A new feature effect method, RHALE, that addresses aggregation
  bias by providing information on the heterogeneity of instance-level
  effects.
\item A formulation of the selection of variable-sized bins to balance
  RHALE bias and variance as an optimization problem.
\item An algorithm that efficiently solves the optimal bin
  partitioning problem.
\item A thorough experimental evaluation of RHALE on synthetic and
  real datasets, demonstrating its superiority over other feature
  effect methods, both in terms of accuracy and efficiency.
\end{itemize}

The code for reproducing all experiments is provided at \href{https://github.com/givasile/RHALE}{https://github.com/givasile/RHALE}.

\section{Background and related work}
\label{sec:background}

Let \(\mathcal{X} \in \mathbb{R}^d\) be the \(d\)-dimensional feature
space, \(\mathcal{Y}\) the target space and
\(f(\cdot) : \mathcal{X} \rightarrow \mathcal{Y}\) the black-box
function.  We use index \(s \in \{1, \ldots, d\}\) for the feature of
interest and \(c = \{1, \ldots, d\} - s\) for the rest.  For
convenience, to denote the input vector, we use \((x_s, \xc)\) instead
of \((x_1, \cdots , x_s, \cdots, x_D)\) and, for random variables,
\((X_s, X_c)\) instead of \((X_1, \cdots , X_s, \cdots, X_D)\).  The
training set \(\mathcal{D} = \{(\xb^i, y^i)\}_{i=1}^N\) is sampled
i.i.d.\ from the distribution \(\mathbb{P}_{X,Y}\).  Finally,
\(f^{\mathtt{<method>}}(x_s)\) denotes how \(\mathtt{<method>}\)
defines the feature effect and \(\hat{f}^{\mathtt{<method>}}(x_s)\)
how it estimates it from the training set.

\subsection{Feature Effect Methods}
\label{subsec:feat-effect-meth}

The most popular feature effect methods are: \emph{Partial Dependence
  Plots} (PDP) and \emph{Accumulated Local Effects} (ALE).  PDP
defines the FE as an expectation over the marginal distribution
\(X_c\), i.e.,
\(f^{\mathtt{PDP}}(x_s) = \mathbb{E}_{X_c}[f(x_s,X_c)]\).  A variation
of PDP, known as Marginal Plots (MP), computes the expectation over
the conditional distribution \(X_c|x_s\), i.e.,
\(f^{\mathtt{MP}}(x_s) = \mathbb{E}_{X_c|x_s}[f(x_s, X_c)]\).  Both
methods suffer from misestimations when features are correlated.  PDP
integrates over unrealistic instances and MP computes aggregated
effects, i.e., attributes the combined effect of sets of features to a
single feature~\cite{apley2020visualizing}. ALE tackles these
limitations, using a three-step computation; (a) the local effect at
\((z, X_c)\),
\(f^s(z, X_c) = \frac{\partial f}{\partial x_s} (z, X_c)\), is
computed with the derivatives $\frac{\partial f}{\partial x_s}$ to
isolate the effect of $x_s$, (b) the expected effect at \(z\),
\(\mu(z) = \mathbb{E}_{X_c|z}\left [f^s (z, X_c)\right ]\), is taken
over $X_c|z$, and, (c) the accumulation, $\int \mu(z) dz $, retrieves
the main effect.  ALE definition is:
\begin{equation}
  \label{eq:ALE}
  f^{\mathtt{ALE}}(x_s) = \int_{x_{s,\min}}^{x_s} \underbrace{\mathbb{E}_{X_c|X_s=z}\left [f^s (z, X_c)\right ]}_{\mu(z)} \partial z
\end{equation}
where \(x_{s,\min}\) is the minimum value of the \(s\)-th feature.
In real ML problems, $p(\mathbf{X})$ is unknown, so~\cite{apley2020visualizing} proposed estimating ALE
from the training set with:
\begin{equation}
  \label{eq:ALE_accumulated_mean_est}
  \hat{f}^{\mathtt{ALE}}(x_s) = \sum_{k=1}^{k_x} \frac{1}{|\mathcal{S}_k|} \sum_{i:\mathbf{x}^{(i)} \in
    \mathcal{S}_k} \left [ f(z_{k}, \xc^{(i)}) - f(z_{k-1}, \xc^{(i)})) \right ]
\end{equation}
where \(k_x\) the index of the bin such that
\(z_{k_x-1} \leq x_s < z_{k_x} \) and \(\mathcal{S}_k\)
is the set of the instances of the \(k\)-th bin, i.e.
\( \mathcal{S}_k = \{ \xb^i : z_{k-1} \leq x^{(i)}_s < z_{k} \} \).
In Eq.~\eqref{eq:ALE_accumulated_mean_est} the axis of the $s$-th feature
is split in \(K\) equally-sized bins and the
average effect in each bin (bin effect) is estimated using synthetic instances,
where $x^{(i)}_s$ is set to the right $(z_k)$ and left $(z_{k-1})$ limits of the bin.
Recently, \cite{gkolemis22} proposed the Differential ALE (DALE)
that computes the local effects of differentiable models
without modifying the training instances:
\begin{equation}  \label{eq:DALE_accumulated_mean_est}
  \hat{f}^{\mathtt{DALE}}(x_s) = \Delta x \sum_{k=1}^{k_x} \frac{1}{|\mathcal{S}_k|} \sum_{i:\mathbf{x}^{(i)} \in
    \mathcal{S}_k} f^s(\mathbf{x}^i)
\end{equation}
Their method allows formulating large bins without creating
out-of-distribution instances and changing the bin size without the
need to recalculate the instance-level effects.  However, both
Eq.~(\ref{eq:ALE_accumulated_mean_est}) and
Eq.~(\ref{eq:DALE_accumulated_mean_est}) are limited to equal-width
partitioning, which has limitations. As discussed in the Introduction,
selecting between narrow and wide bins is challenging and, even more,
there are cases (Figure~\ref{fig:ale-different-bins}) where
neither narrow nor wide bins are appropriate. In these cases, it is
necessary to use variable bin sizes (Figure~\ref{fig:concept-figure-subfig-3}).

\subsection{Heterogeneity Of Local Effects}
\label{subsec:quant-heter-effects}

Relying only on the average effect may provide a misleading interpretation of the model. Thus, there is an increasing interest in quantifying the degree of divergence between local effects and the average effect, which is commonly referred to as heterogeneity of local effects. To measure heterogeneity, PDP has a local equivalent called Individual Conditional Expectation (ICE)~\cite{goldstein2015peeking}.
ICE, along with its variations like c-ICE and d-ICE~\cite{goldstein2015peeking}, creates one curve per instance, \(f^{\mathtt{ICE}}_i(x_s) = f(x_s, \xc^{(i)})\), on top of the average PDP plot, as seen in Figure~\ref{fig:concept-figure-subfig-2}. In this way, the user assesses the heterogeneity by visually inspecting the similarity between ICE curves. However, as demonstrated in Section~\ref{subsec:simulation-examples-1}, ICE plots have the same limitations as PDPs in cases of correlated features.
Based on the variance of the ICE plots, \cite{molnar2021relating} proposed a method to quantify the standard error around the PDP plot. Some other
methods\cite{herbinger2022repid, britton2019vine, molnar2020model}
attempt to address PDP-ICE failure in case of correlated features by
clustering ICE plots based on their similarity.  The focus of these
works, however, is on regional effects, i.e., subsets of the input
space with homogeneous effects, rather than global effects.
Approaches such as the H-Statistic~\cite{friedman2008predictive},
Greenwell's interaction index~\cite{greenwell2018simple}, and SHAP
interaction values~\cite{lundberg2018consistent} provide a metric that
quantifies the level of interactions between feature pairs but do not
provide insight into how interactions influence different parts of the
feature effect plot.  To the best of our knowledge, no existing method
quantifies heterogeneous effects for ALE.

\section{RHALE}
\label{sec:UALE}

\begin{figure}
    \centering
    \begin{subfigure}{.23\textwidth}
      \centering
      \includegraphics[width=1\textwidth]{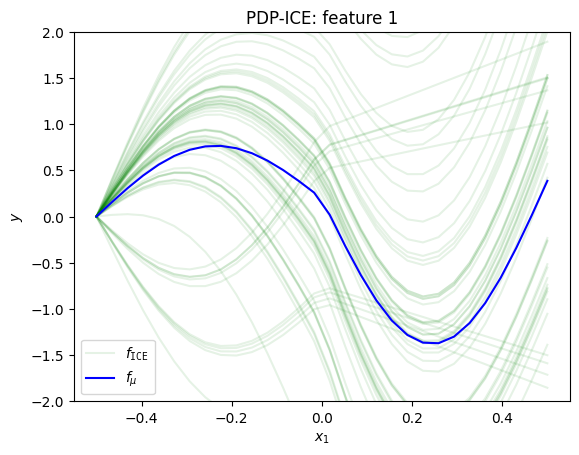}
      \caption{PDP-ICE plot}
      \label{fig:concept-figure-subfig-2}
    \end{subfigure}
    \begin{subfigure}{.23\textwidth}
      \centering
      \includegraphics[width=1\textwidth]{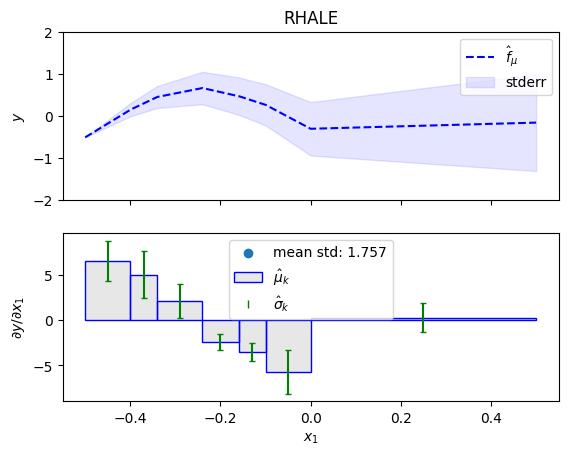}
      \caption{RHALE plot}
      \label{fig:concept-figure-subfig-3}
    \end{subfigure}
    \caption{Feature effect for $x_1$ on the example of Eq.~\ref{eq:target_function}. Due to feature correlations, only RHALE provides a robust estimation of the main effect and the heterogeneity.}
\label{fig:concept-figure}
\end{figure}

RHALE visualizes the feature effect with a plot as illustrated in
Figure~\ref{fig:concept-figure-subfig-3}. The plot includes
(a) $\hat{f}_{\mu}^{\mathtt{RHALE}}(x_s)$, the robust estimation of ALE that shows the average effect (\textit{RHALE estimation}),
(b) $\mathtt{STD}(x_s)$, the standard deviation of the ALE effect that shows the heterogeneity of the instance level effects (\textit{STD}),
(c) $\hat{\mu}_k \forall k$, the bin effects that show the average change on the output $y$ given a small change in \(x_s\) (\textit{bin effect}) and
(d) $\hat{\sigma}_k \forall k$ the bin standard deviations that quantify the heterogeneity inside each bin (\textit{bin std}).
In each bin, a violin plot on top of the bin effect shows the exact
distribution of the local effects. The variable-size
partitioning presented in Section~\ref{subsec:bin-spliting} leads to
an accurate estimation of these quantities.

To explain these four interpretable quantities and to highlight the
advantages of RHALE compared to PDP-ICE, we will use a running
example. We define a generative distribution
$p(\xb)= p(x_1)p(x_2)p(x_3|x_1)$ where $x_3$ is highly correlated with
$x_1$, while $x_2$ is independent from both.  Specifically, \(x_1\)
lies in \([-0.5, 0.5]\) with most samples inside the first half, i.e.
$p(x_1) = \frac{5}{6}\mathcal{U}(x_1;-0.5, 0) +
\frac{1}{6}\mathcal{U}(x_1;0, 0.5)$, \(x_3\) is almost equal to
\( x_1\) i.e., $p(x_3|x_1) = \mathcal{N}(x_3; 0, \sigma_3=0.01)$ and
$p(x_2) = \mathcal{N}(x_2; 0, \sigma_2=2)$. In the experiments, we use
60 samples drawn i.i.d. from \(p(\xb)\).  The predictive function is:
\begin{equation}
  \label{eq:target_function}
  f(\xb) = \sin(2 \pi x_1) (\mathbbm{1}_{x_1 < 0} - 2 \mathbbm{1}_{x_3 < 0}) + x_1 x_2 + x_2
\end{equation}
The simplicity of the toy example helps us isolate the effect of $x_1$,
which is $f(x_1) \approx -\sin(2 \pi x_1) \mathbbm{1}_{x_1<0}$.
This is because $x_3 \approx x_1$, so $(\mathbbm{1}_{x_1 < 0} - 2 \mathbbm{1}_{x_3 < 0} \approx - \mathbbm{1}_{x_1 < 0})$ and the effect of $x_1 x_2$ is $x_1 \mathbb{E}_{x_2} [x_2] = 0 $.
Furthermore, the only term that introduces heterogeneity is $x_1 x_2$,
due to $x_2 \sim \mathcal{N}(0, 2)$ that varies among instances.
Detailed derivations are provided at the Appendix B.1.

In Figure~\ref{fig:concept-figure-subfig-3} we show that a user can
interpret these effects from the RHALE plot. Specifically, from the
top subplot, a user can interpret that (a) the average effect of
\(x_1\) is
\(\hat{f}_{\mu}^{\mathtt{RHALE}}(x_1) \approx -\sin(2 \pi x_1)
\mathbbm{1}_{x_1<0}\), and is produced after aggregating (b) instance
level effects that vary in the region
\(-\sin(2 \pi x_1) \mathbbm{1}_{x_1<0} \pm 2x_1\). From the bottom
subfigure, the user can interpret the FE at the bin-level. For
example, for \(x_1 \in [-0.5, -0.4]\) (c) the average change on the
output is about \(\frac{\partial f}{\partial x_1} \approx 6\) units of
\(y\) and is produced after aggregating instance-level changes that
vary in \(6 \pm 2\). Furthermore, the violin plots show the exact
distribution of the instance-level changes.

\begin{figure}[t]
  \centering
  \includegraphics[width=.49\linewidth]{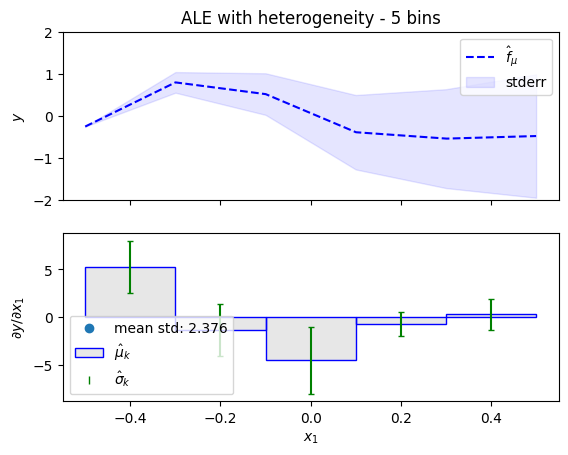}
  \includegraphics[width=.49\linewidth]{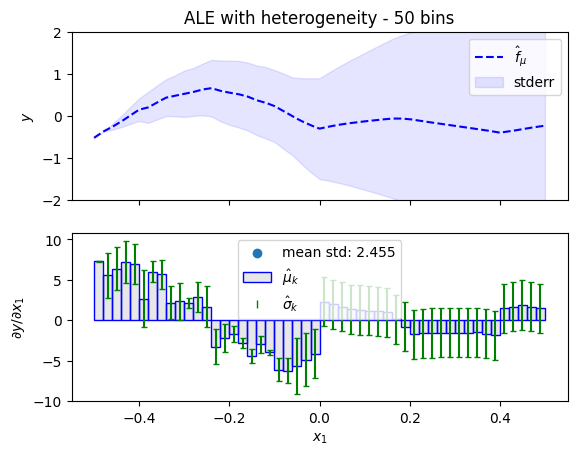}
  \caption{Estimation of the ALE effect, the standard error of ALE, the bin effect and the bin standard deviation using fixed-sized bins, $K=5$ (left) and $K=50$ (right).}
  \label{fig:ale-different-bins}
\end{figure}

For estimating the above quantities from the 60 available
samples, the optimized partitioning divides the sinusoidal
region in six bins (dense enough) and merges the constant region in a
single bin (robust estimation), balancing estimation variance and bias
(Section~\ref{subsec:bin-spliting}).
In contrast, Figure~\ref{fig:ale-different-bins} shows that \textit{all}
fixed-size splits result in poor estimations;
when using sparse bins ($K=5$) the estimation is biased,
as will be explained in Section~\ref{subsec:uale-definition-1},
and when using dense bins ($K=50$) the estimation has high variance.

A natural question that arises is whether we could come to the same
interpretation using the PDP-ICE plot. At
Figure~\ref{fig:concept-figure-subfig-2} we observe that PDP with
c-ICE, i.e., ICE curves centered to start from zero, lead to a
completely misleading interpretation. For example, in
$x_1 \in [0, 0.5]$, PDP shows a negative sinusoidal average effect and
c-ICE two heterogeneous effects; a negative sinusoidal when
$x_3^{(i)} \geq 0$ (for about $\frac{5}{6}$ of the instances), and a
linear when $x_3^{(i)} < 0$ (for about $\frac{5}{6}$ of the
instances). This is because PDP-based methods ignore the correlation
between $x_1$ and $x_3$.

\subsection{Definition}
\label{subsec:uale-definition-1}

We define the heterogeneity at \(x_s=z\) as the standard deviation $\sigma(z)$
of the instance-level effects, where:
\begin{equation}
  \label{eq:ALE_var}
  \sigma^2(z) = \mathbb{E}_{X_c|X_s=z}\left [ \left (\dfdx (z, X_c) - \mu(z) \right )^2 \right ]
\end{equation}
The variability is introduced by the implicit feature interactions.
If the black-box function does not have any interaction term, i.e.,
it can be written as $f(\xb) = f_s(x_s) + f_c(\xc)$ then the variability is zero.
For the interval-based formulation, we define the bin effect
\(\mu(z_1, z_2)\) and the bin standard deviation \(\sigma(z_1, z_2)\) as:
\begin{equation}
  \label{eq:mu_bin}
    \mu(z_1, z_2) = \mathbb{E}_{z \sim \mathcal{U}(z_1,z_2)} [\mu(z)]
    = \frac{\int_{z_1}^{z_2} \mu(z) \partial z}{z_2 - z_1}
\end{equation}
\begin{equation}
  \label{eq:var_bin}
  \sigma^2(z_1, z_2) = \mathbb{E}_{z \sim \mathcal{U}(z_1,z_2)} [\sigma^2(z)] =  \frac{\int_{z_1}^{z_2} \sigma^2(z)  \partial z}{z_2 - z_1}
\end{equation}
The bin effect and the bin standard deviation quantify the average effect and
the heterogeneity inside a bin, i.e., for a population
\(\xb^{(i)} = (z^{(i)}, \xc^{(i)})\), where
\(z^{(i)}\) is uniformly drawn from \(\mathcal{U}(z_1, z_2)\) and
$\xc$ from $X_c|z^{(i)}$.
Denoting as \(\mathcal{Z}\) the sequence of
\(K+1\) points that partition the axis of the \(s\)-th feature into \(K\) variable-size
intervals, i.e., \(\mathcal{Z} = \{z_0, \ldots, z_K\}\),
the interval-based formulation of RHALE is:
\begin{equation}
  \label{eq:ALE_2}
  \tilde{f}^{\mathtt{RHALE}}_{\mathcal{Z}}(x_s) = \sum_{k=1}^{k_x} \mu(z_{k-1}, z_k) (z_k - z_{k-1})
\end{equation}
where \(k_x\) is the index of the bin such that
\( z_{k_x - 1} \leq x_s < z_{k_x}\).
Eq.~\eqref{eq:ALE_2} is no more than a
piece-wise linear approximation of Eq.~\eqref{eq:ALE}.
The approximation of the bin effect and of the bin standard deviation is made from the
set \(\mathcal{S}_k\) of instances with the \(s\)-th feature
in the \(k\)-th bin, i.e.,
\( \mathcal{S}_k= \{ \mathbf{x}^i : z_{k-1} \leq x^{(i)}_s < z_k \}
\).
The bin effect is estimated with:
\begin{equation}
  \label{eq:mu_bin_approx}
  \hat{\mu}(z_{k-1}, z_k) = \frac{1}{|\mathcal{S}_k|}
  \sum_{i:\mathbf{x}^i \in \mathcal{S}_k} \left [ \dfdx(\mathbf{x}^i)
  \right ]
\end{equation}
which is an unbiased estimator of Eq.~\eqref{eq:mu_bin} (Appendix A.1).
The estimator of the bin deviation Eq.~\eqref{eq:var_bin} is:
\begin{equation}
  \label{eq:var_bin_approx}
  \hat{\sigma}^2(z_{k-1}, z_k) = \frac{1}{|\mathcal{S}_k| - 1}
\sum_{i:\mathbf{x}^i \in \mathcal{S}_k} \left ( \dfdx(\mathbf{x}^i) -
  \hat{\mu}(z_1, z_2) \right )^2
\end{equation}
At Appendix A.2, we show that \(\hat{\sigma}^2(z_1, z_2)\) is an unbiased estimator
of \(\sigma^2_*(z_1, z_2) = \frac{\int_{z_1}^{z_2} \mathbb{E}_{X_c|X_s=z}
  \left [ (f^s(z, X_c) - \mu(z_1, z_2) )^2 \right] \partial z}{z_2 -
  z_1} \) and in Theorem~\ref{sec:theorem-1} we prove that in the general
case, \(\sigma_*^2(z_1, z_2) \geq \sigma^2(z_1, z_2)\).
Therefore, without a principled bin-splitting strategy,
\(\hat{\sigma}^2(z_1, z_2)\) leads to an overestimation of the actual bin standard
deviation \(\sigma^2(z_1, z_2)\).
\begin{theorem}
  \label{sec:theorem-1}
  If we define (a) the residual \(\rho(z)\) as the difference between
  the expected effect at \(z\) and the bin effect, i.e,
  \(\rho(z) = \mu(z) - \mu(z_1, z_2)\), and (b)
  \(\mathcal{E}(z_1, z_2)\) as the mean squared residual of the bin,
  i.e.,
  \(\mathcal{E}(z_1, z_2) = \frac{\int_{z_1}^{z_2}\rho^2(z) \partial
    z}{z_2 - z_1}\), then it holds
\begin{equation}
    \label{eq:bin-uncertainty-proof}
 \sigma_*^2(z_1, z_2) = \sigma^2(z_1, z_2) + \mathcal{E}^2(z_1, z_2)
\end{equation}
\end{theorem}
\begin{proof}
The proof is at A.3 of the Appendix 
\end{proof}
We refer to \(\mathcal{E}^2(z_1, z_2)\) as bin error.
Based on
Theorem\ref{sec:theorem-1}, the estimation is unbiased only when
\(\mathcal{E}^2(z_1, z_2) = 0\).

\subsection{Automatic Bin-Splitting}
\label{subsec:bin-spliting}

RHALE approximation is affected by (a) the number of instances
(estimation variance) and (b) the error term \(\mathcal{E}(z_1, z_2)\)
(estimation bias), in each bin.  On the one hand, we favor wide bins
so that the estimation of
\(\hat{\mu}(z_1, z_2), \hat{\sigma}(z_1, z_2)\) comes from a
sufficient population of samples (low estimation variance).  On the
other hand, we want to minimize the accumulated bin error, i.e.,
\( \mathcal{E}^2_{\mathcal{Z}} = \sum_{k=1}^K\mathcal{E}^2(z_{k-1},
z_k) \Delta z_k\), where \(\mathcal{Z} = \{z_0, \cdots, z_K\}\) and
\(\Delta z_k = z_k - z_{k-1}\) (low estimation bias).  We search for a
partitioning that balances this trade-off.
\begin{corollary}
  \label{sec:coroll}
  If a bin-splitting \(\mathcal{Z}\) minimizes the accumulated error $\mathcal{E}^2_{\mathcal{Z}}$,
  then it also minimizes
  \(\sum_{k=1}^K\sigma_*^2(z_1, z_2) \Delta z_k \).
\end{corollary}
\begin{proof}
  ~\label{sec:coroll-1} The proof is based on the observation that
  \(\sum_{k=1}^K \sigma^2(z_{k-1}, z_k) \Delta z_k = \sigma^2(z_0,
  z_K) (z_K - z_0)\) which is independent of the bin-splitting.
  A detailed proof is provided at Appendix A4.
\end{proof}
Corollary~\ref{sec:coroll} shows that minimizing
\( \mathcal{E}^2_{\mathcal{Z}} \) is equivalent to minimizing
\(\sum_{k=1}^K \sigma_*^2(z_{k-1}, z_k)\Delta z_k\), which can be directly
estimated from \(\sum_{k=1}^K \hat{\sigma}^2(z_{k-1}, z_k) \Delta z_k \).
Based on that, we set-up the following optimization problem:
\begin{equation}
  \label{eq:opt}
\begin{aligned}
  \min_{ \mathcal{Z} = \{z_0, \ldots, z_K\}} \quad & \mathcal{L} = \sum_{k=1}^K \tau_k \hat{\sigma}^2(z_{k-1}, z_k) \Delta z_k \\
  \textrm{where} \quad & \Delta z_k = z_k - z_{k-1} \\
  & \tau_k = 1 - \alpha \frac{|S_k|}{N} \\
  \textrm{s.t.} \quad & |\mathcal{S}_k| \geq N_{\mathtt{PPB}}\\
                                     & z_0 = x_{s,min}\\
                                     & z_K = x_{s, max}
\end{aligned}
\end{equation}
The objective \(\mathcal{L}\) searches for a partitioning
\(\mathcal{Z}^* = \{ z_0^*, \ldots, z_K^* \} \) with a low accumulated
error \(\mathcal{E}_{\mathcal{Z}}^2\) and when many partitionings have
similar accumulated errors, the coefficient \(\tau_K\) favors the one
with wider bins (on average, more points per bin). The constraint of
at least \(N_{\mathtt{PPB}}\) points per bin sets the lowest-limit for
a \textit{robust} estimation.  The user can choose to what extent they
favor the creation of wide bins through the parameter \(\alpha\) that
controls the discount \(\tau_k\) and the parameter
\(N_{\mathtt{PPB}}\) that sets the minimum population per bin. A
typical choice for \(\alpha\) is \(0.2\), which means a discount
between range of \([0\%, 20\%]\) and for \(N_{\mathtt{PPB}}\) is
\(\frac{N}{20}\), which means at least \(\frac{N}{20}\) points in each
bin, where \(N\) is the dataset size.

For solving the optimization problem of~Eq.\ref{eq:opt} we discretize the solution space.
First, we set a threshold \(K_{\max}\) on the maximum
number of bins which, in turn, defines the minimum bin width,
i.e. \(\Delta x_{\min} = \frac{x_{s, \max} - x_{s,\min}}{K_{\max}}\).
Based on that, we restrict the bin limits to the multiples of the
minimum width, i.e.
\(z_k = k\cdot \Delta x_{\min}, \text{where } k \in \{0 , \cdots,
K_{\max}\} \).
In this discretized solution space, we find the global
optimum using Dynamic Programming.
To define the solution, we use two indexes;
index
\(i \in \{0, \ldots, K_{\max}\}\) denotes the limit of the $i$-th
bin \((z_i)\) and the index \(j \in \{0, \ldots, K_{\max}\} \)
denotes the \(j\)-th multiple of the minimum step, i.e.,
\(x_j = x_{s,\min} + j \cdot \Delta x_{\min}\).
The recursive cost
function \(T(i,j)\) computes the cost of setting \(z_i\) to \(x_j\):
\begin{equation}
  \label{eq:recursive_cost}
  \mathcal{T}(i,j) = \mathrm{min}_{l \in \{0, \ldots, K_{max}\}} \left [ \mathcal{T}(i-1, l) + \mathcal{B}(l, j) \right ]
\end{equation}
%
The term \(\mathcal{B}(l, j)\) is the cost of creating a bin with
limits \([x_l, x_j)\).  In our case, following Eq.~\eqref{eq:opt}, we
set it to \(\tau_k \hat{\sigma}^2(x_l, x_j) (x_j-x_l)\) if the bin is
valid, i.e., \(|\mathcal{S}_k| \geq N_{\mathtt{PPB}}\), and to
\(\infty\) otherwise.  The optimal partitioning $\mathcal{Z}^*$ is
given by solving \(\mathcal{L} = \mathcal{T}(K_{\max}, K_{\max})\) and
keeping track of the sequence of steps. Therefore, the main RHALE
effect is estimated as in Eq.~(\ref{eq:RHALE}) and its standard
deviation as in Eq.~(\ref{eq:SE}):

\begin{equation}
  \label{eq:RHALE}
  \hat{f}_{\mathcal{Z}^*}^{\mathtt{RHALE}}(x_s) = \sum_{k=1}^{k_x} \hat{\mu}(z_{k-1}, z_k)(z_k - z_{k-1})
\end{equation}

\begin{equation}
  \label{eq:SE}
  \mathtt{STD}(x_s) = \sqrt{\sum_{k=1}^{k_x} (z_k - z_{k-1})^2 \hat{\sigma}^2(z_{k-1}, z_k)}
  \end{equation}
The bin effects $\hat{\mu}_k$ are estimated using Eq.~\ref{eq:mu_bin_approx} and
and the heterogeneity by the standard deviation $\hat{\sigma}_k$ in each bin
using Eq.~\ref{eq:var_bin_approx}.

\paragraphb{Computational Complexity.}
The computational complexity of the DP solution is
\(\mathcal{O}(K_{\max}^3)\) because we use the DALE formula of
Eq.~\ref{eq:DALE_accumulated_mean_est}. This allows us to precompute
the instance-level effects once in the beginning, and then, the
bin-splitting algorithm simply reallocates them to different
partitionings without reevaluating \(f\) for each partitioning. As a
result, for up to roughly \(K_{\max} = 100\) bins, our algorithm runs
in a couple of seconds, regardless of the dataset size or the cost of
evaluating \(f\). On the other hand, a PDP-ICE plot needs to evaluate
\(f\) on \(t\) positions along the \(x_s\) axis for all \(N\) dataset
points, making it a much slower alternative. Additional details and
experimental results on the computational aspect can be found in
Appendix A5. Finally, it is worth noting that $K_{\max}$ only sets an
upper limit and the optimal sequence $\mathcal{Z}^*$ can range from
$1$ to $K_{\max}$.

\section{Simulation examples}
\label{sec:simulation-examples}

To formally evaluate RHALE, we rely on simulation examples as the
evaluation requires knowledge of the ground truth generating
distribution \(X\) and the black-box function \(f\). In contrast, in
Section~\ref{sec:real-world-example}, we showcase the applicability of
RHALE on a real-world dataset, but it impossible to conduct a formal
evaluation in this setting. The evaluation of RHALE on simulation
examples is two-fold. First, in
Section~\ref{subsec:simulation-examples-1} we conduct a formal
comparison between RHALE and PDP-ICE to verify that RHALE performs
well in cases with correlated features, which PDP-ICE struggles
with. Second, in Section~\ref{subsec:simulation-examples-2}, we compare
RHALE's automated bin-splitting approach against the traditional
fixed-size approximation. We demonstrate that RHALE's bin-splitting
technique produces more accurate estimations across various scenarios.

\subsection{RHALE vs PDP-ICE}
\label{subsec:simulation-examples-1}

\begin{figure}[h]
  \centering
  \includegraphics[width=.23\textwidth]{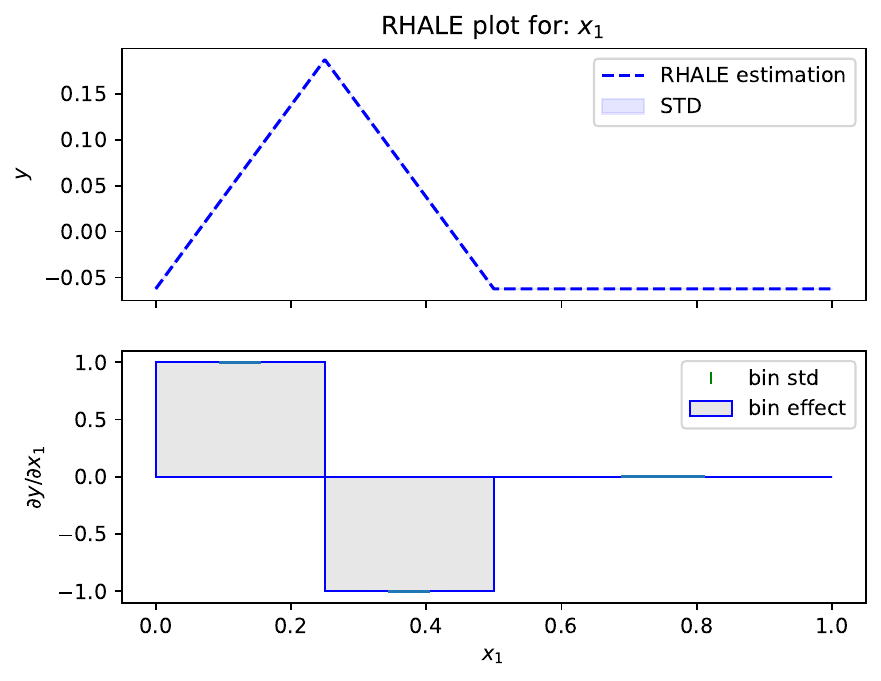}
  \includegraphics[width=.23\textwidth]{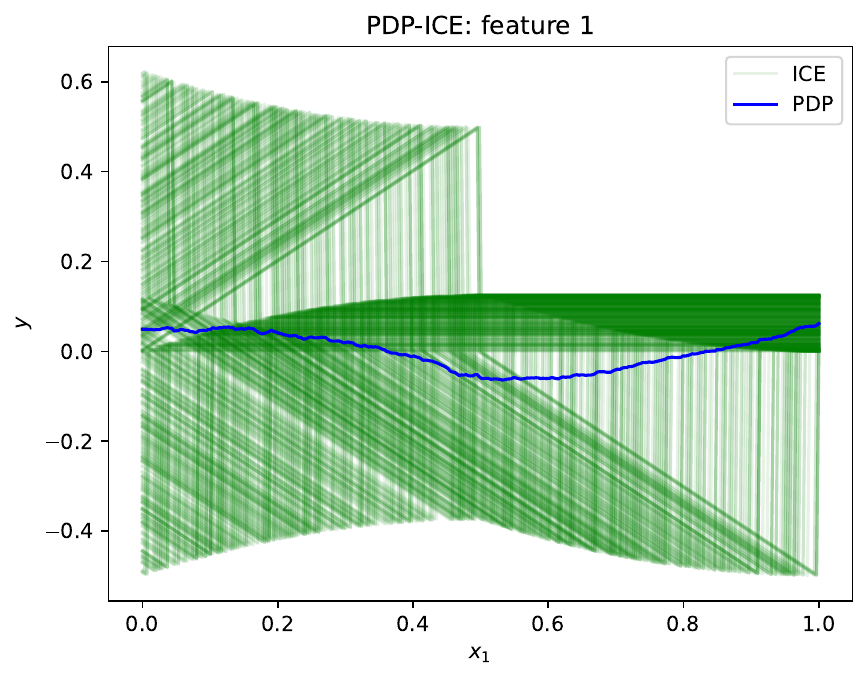}
  \caption{No interaction, Equal weights: Feature effect for \(x_1\)
    using RHALE (Left) and PDP-ICE (Right).}
  \label{fig:synth-ex-1-case-1}
\end{figure}

We consider a data generating distribution
\(p(\mathbf{x}) = p(x_3)p(x_2|x_1)p(x_1)\), where
\(x_1 \sim \mathcal{U}(0,1)\), \(x_2 = x_1 + \epsilon \) and
\(x_3 \sim \mathcal{N}\left(0, \sigma_3^2 = \frac{1}{4}\right)\). Here
$\epsilon \sim \mathcal{N}(0, 0.01)$ is a small additive noise.  The
predictive function is:
\begin{equation}
  \label{eq:synth-ex-1-function}
  f(\mathbf{x}) = \alpha f_2(\xb) + \underbrace{f_1(\xb) \mathbbm{1}_{f_1(\xb) \leq \frac{1}{2}}}_{g_1(\xb)} + \underbrace{(1 - f_1(\xb)) \mathbbm{1}_{\frac{1}{2} < f_1(\xb) < 1}}_{g_2(\xb)}
\end{equation}
where \(f_1(\mathbf{x}) = x_1 + x_2\) is the additive term and
\(f_2(\mathbf{x}) = x_1 x_3\) is the interaction term. We evaluate
RHALE and PDP-ICE when (a) there is no heterogeneity (\(\alpha=0\))
and (b) there is is heterogeneity implied by the interaction term
(\(\alpha > 0\)). We use this simple example for being able to
establish a ground truth for the main effect and the
heterogeneity. For the main effect, we use that due to
$x_2 \approx x_1$ we can determine the intervals where $g_1$ or $g_2$
are active. For the heterogeneity, we use the fact that under
no-interactions, \(a=0\), the heterogeneity must be zero and we
discuss separately the case with \(a>0\).  For detailed derivations
see Appendix B1.
\paragraphb{Case a: Interaction term disabled.}
Given that $x_2 \approx x_1$, when \( 0 \leq x_1 < \frac{1}{4}\), then
\(f_1(\xb) < \frac{1}{2}\), so the effect is $x_1$ and, similarly,
when $\frac{1}{4} \leq x_1 < \frac{1}{2}$ the effect is
$-x_1$. Therefore, the ground truth effect is
\(f^{\mathtt{GT}}(x_1) = x_1 \mathbbm{1}_{x_1 < \frac{1}{4}} +
\left(\frac{1}{4} - x_1\right) \mathbbm{1}_{\frac{1}{4} \leq x_1 <
  \frac{1}{2}}\).  Since \(x_1\) does not interact with any other
feature, the heterogeneity is zero.  In
Figure~\ref{fig:synth-ex-1-case-1}, we observe that PDP's main effect
is wrong and ICE plots show heterogeneous effects.  In contrast, RHALE
estimates correctly both the average effect and the heterogeneity.
Finally, we observe that RHALE's bin-splitting optimally creates three
wide bins,
\([0, \frac{1}{4}), [\frac{1}{4}, \frac{1}{2}), [\frac{1}{2}, 1)\), in
the regions with linear effect.

\begin{figure}[h]
  \centering
  \includegraphics[width=.23\textwidth]{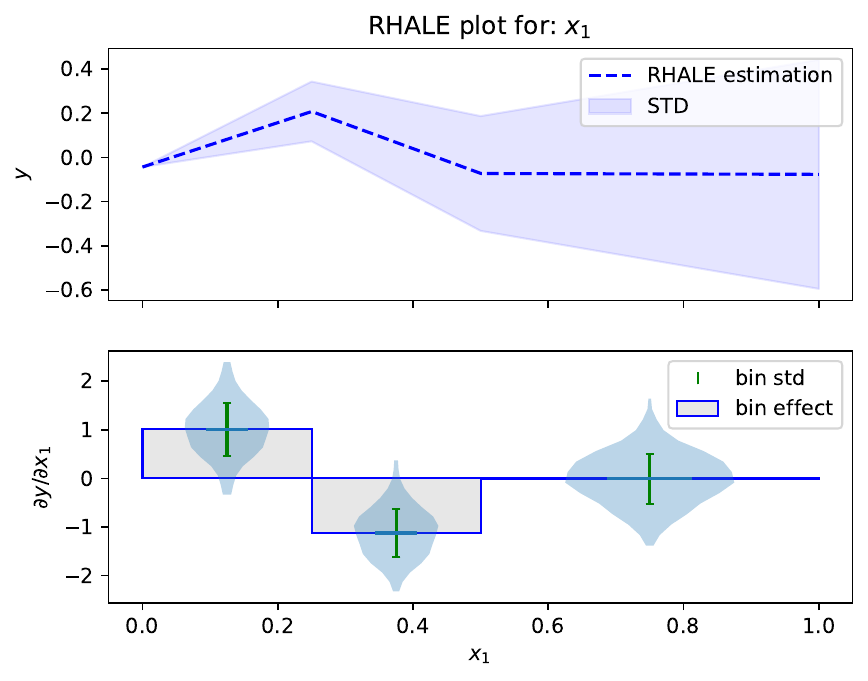}
  \includegraphics[width=.23\textwidth]{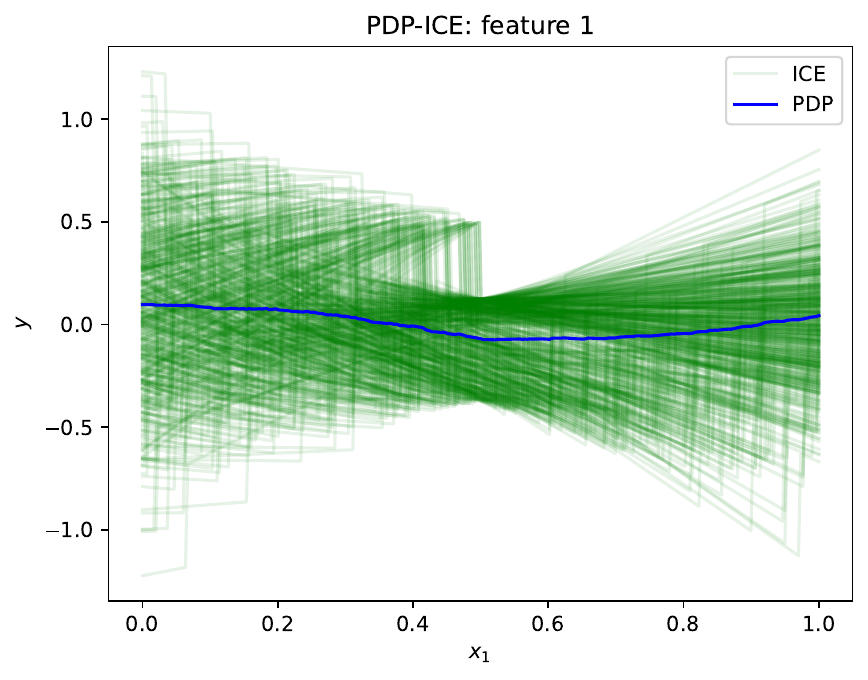}\\
  \includegraphics[width=.23\textwidth]{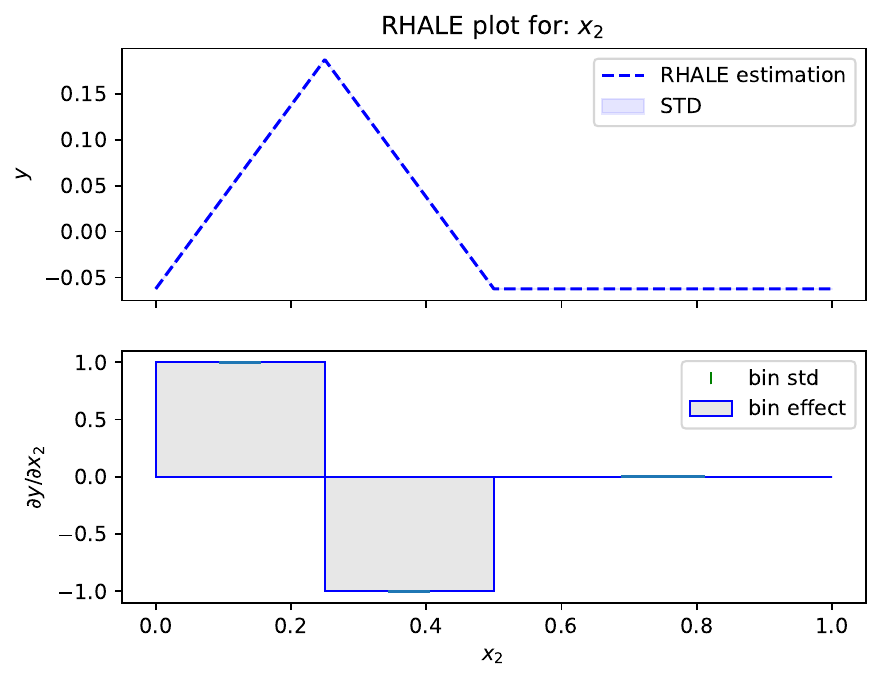}
  \includegraphics[width=.23\textwidth]{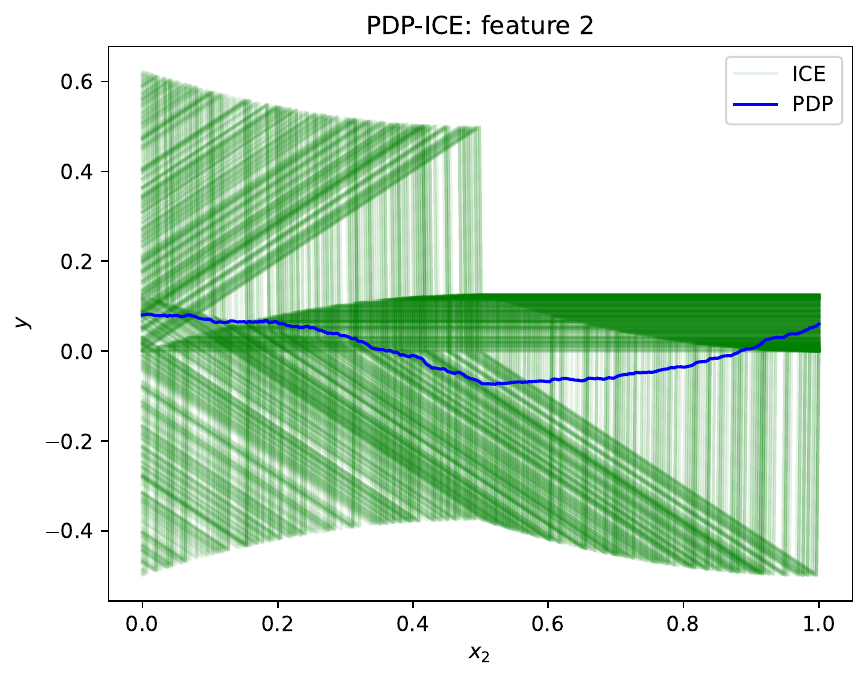}\\
  \includegraphics[width=.23\textwidth]{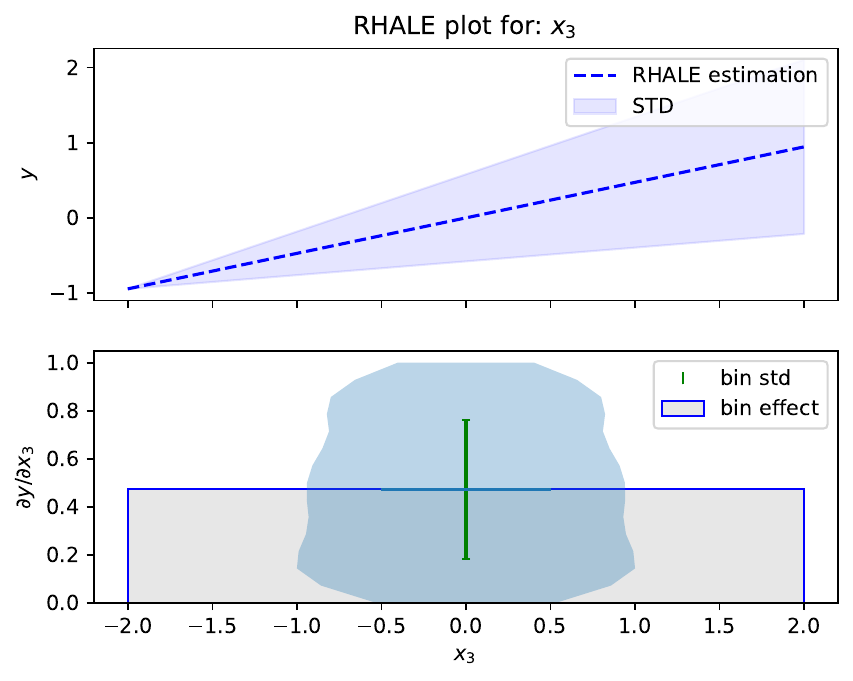}
  \includegraphics[width=.23\textwidth]{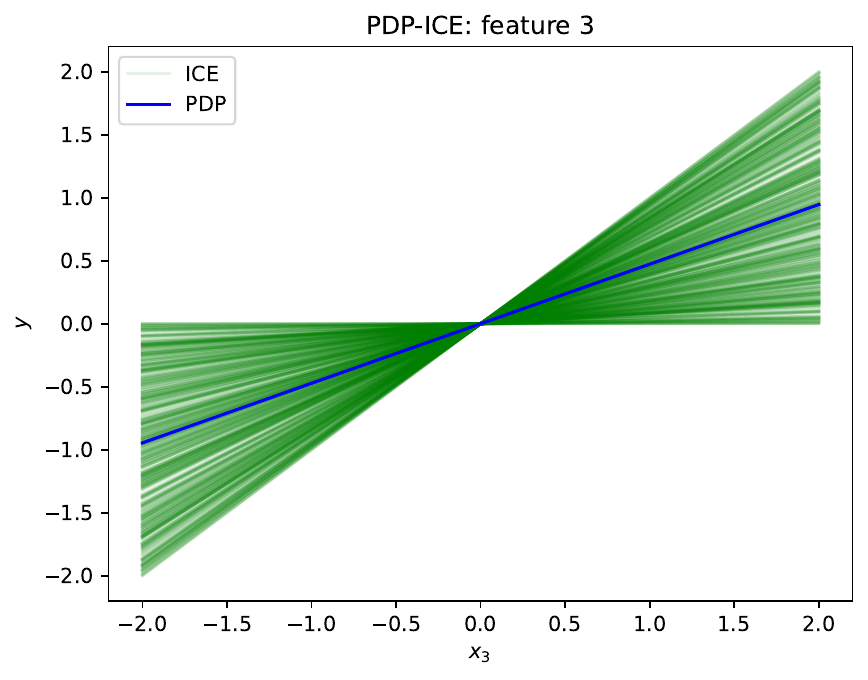}\\
  \caption{With interaction, equal weights: From top to bottom,
    feature effect for features \(\{x_1, x_2, x_3\}\) using RHALE (left
    column) and PDP-ICE (right column).}
  \label{fig:ex-synth-1-3}
\end{figure}

\paragraphb{Case b: Interaction term enabled.}
The main effects are
\(f^{\mathtt{GT}}(x_j) = x_j \mathbbm{1}_{x_j < \frac{1}{4}} +
\left(\frac{1}{4} - x_j\right) \mathbbm{1}_{\frac{1}{4} \leq x_j <
  \frac{1}{2}}\) for features $j=1,2$ and
\(f^{\mathtt{GT}}(x_3) = \frac{1}{2} x_3\) for feature $x_3$.  The
interaction term $x_1 x_2$ induces heterogeneous effects for features
$x_1$ and $x_3$, and since the two variables are independent, the
heterogeneity is \(\sigma_3 = \frac{1}{2}\) for $x_1$ and
\(\sigma_1 = \frac{1}{4}\) for $x_3$.  In
Figure~\ref{fig:ex-synth-1-3}, we observe that RHALE correctly
estimates the main effect and the heterogeneity of all features. In
contrast, PDP-ICE only estimates correctly only the effect and the
heterogeneity of $x_3$. This confirms our previous knowledge that
PDP-ICE performs well only when the interaction terms includes
non-correlated features, like the term \(f_2(\xb)\). For the
correlated features $x_1$ and $x_2$, both the average effect and the
heterogeneity are erroneously estimated by PDP-ICE.

\paragraphb{Discussion.}
The study shows RHALE's superiority under correlated features,
where, PDP and ICE plots can provide highly misleading results.
Additionally, RHALE's automatic bin splitting leads to a robust estimation of the average effect
and of the heterogeneity, favoring wider bins in regions with (near) constant effects.

\subsection{RHALE vs ALE}
\label{subsec:simulation-examples-2}

In this simulation, we compare the performance of RHALE's automatic
partitioning with ALE's fixed-size bin-splitting. To assess the
accuracy of these approximations, we first estimate the ground truth
average effect \(\mu\) and heterogeneity \(\sigma\) using a large
dataset (\(N=10^6\)) with dense fixed-size binning (\(K=10^3\)). We
then generate a smaller dataset (\(N=500\)) and compare the estimation
of $\hat{\mu}, \hat{\sigma}$, using (a) fixed-size bins for several
values of \(K\) against (b) RHALE's automatic partitioning. Our
objective is to show that RHALE provides better
estimates of \(\mu\) and \(\sigma\) compared to any fixed-size
alternative.

The dataset is generated by sampling from
\(p(\mathbf{x}) = p(x_2|x_1)p(x_1)\) where
\(x_1 \sim \mathcal{U}(0,1)\) and
\(x_2 \sim \mathcal{N}(x_1, \sigma_2^2=0.5)\).  RHALE's approximation
is denoted with \(\mathcal{Z^*}\) and the fixed-size with \(K\) bins
as \(\mathcal{Z^{\mathtt{K}}}\). The evaluation is based on the Mean
Absolute Error (MAE) of the bin effect \(\mu\) and of the
heterogeneity \(\sigma\) across bins, i.e.,
\begin{equation}
  \label{eq:eval_met_1}
  \mathcal{L}^{\mu} = \frac{1}{|\mathcal{Z}| - 1} \sum_{k \in
  \mathcal{Z}} | \mu(z_{k-1}, z_k) - \hat{\mu}(z_{k-1}, z_k) |
\end{equation}
\begin{equation}
  \label{eq:eval_met_2}
  \mathcal{L}^{\sigma} = \frac{1}{|\mathcal{Z}| -1} \sum_{k \in
    \mathcal{Z}} | \sigma(z_{k-1}, z_k) - \hat{\sigma}(z_{k-1}, z_k) |
\end{equation}
The ground truth bin effect, \(\mu(z_{k-1}, z_k)\), and heterogeneity,
\(\sigma(z_{k-1}, z_k)\) are obtained by averaging the dense
fixed-size bins within the interval \([z_{k-1}, z_k]\).
We also calculate the mean residual error
\(\mathcal{L}^{\rho} = \frac{1}{|\mathcal{Z}|} \sum_{k \in
  \mathcal{Z}} \mathcal{E}(z_{k-1}, z_k) \) to interpret cases
where the bin standard deviation is biased.

We compare RHALE vs ALE in two different scenarios; when
\(f\) is piecewise linear and \(f\) is non-linear. We
execute \(t = 30\) independent runs, using each time \(N=500\)
different samples, and report the mean values of the metrics.

\begin{figure}[h]
  \centering
  \includegraphics[width=.23\textwidth]{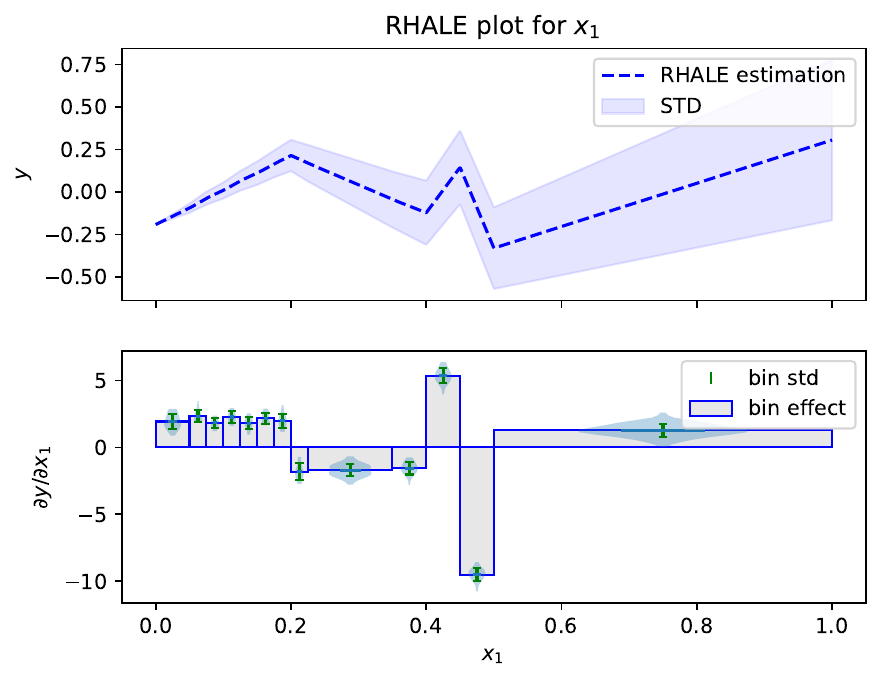}
  \includegraphics[width=.23\textwidth]{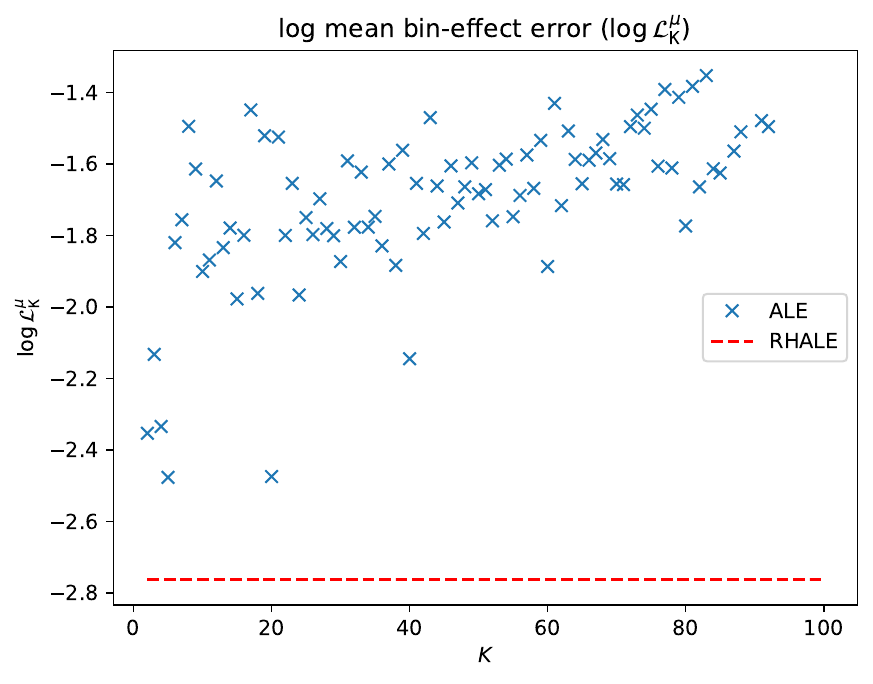}\\
  \includegraphics[width=.23\textwidth]{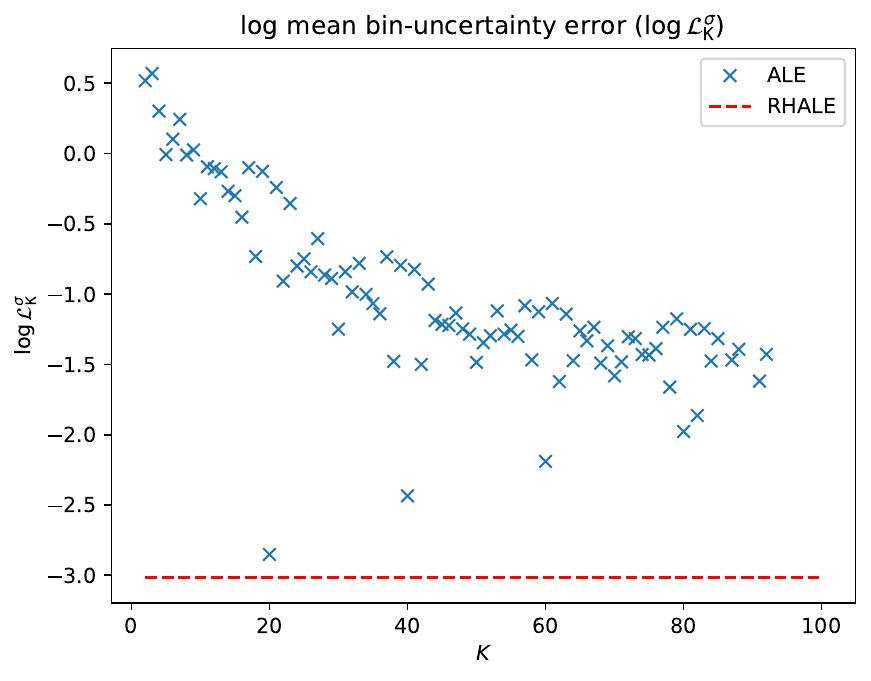}
  \includegraphics[width=.23\textwidth]{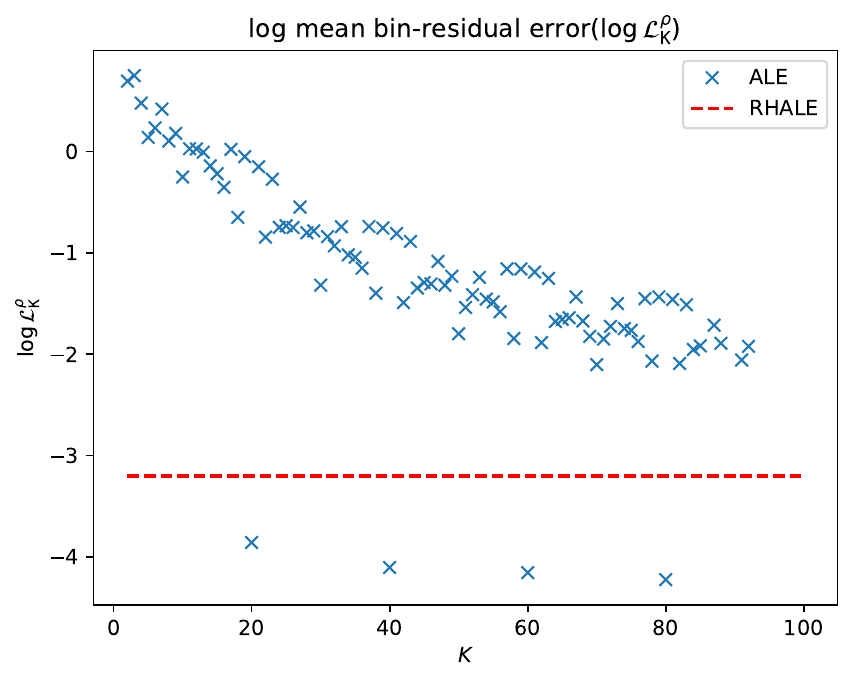}
  \caption{Bin-Splitting, piecewise linear function: RHALE's approximation
    (Top-Left). RHALE vs fixed-size approximations in terms of:
    \(\mathcal{L}^{\mu}\) (Top-Right), \(\mathcal{L}^{\sigma}\)
    (Bottom-Left), \(\mathcal{L}^{\rho}\) (Bottom-Right).}
  \label{fig:ex-synth-2-1}
\end{figure}

\paragraphb{Piecewise Linear Function.}
Here, the black-box function is \(f(\mathbf{x}) = a_1 x_1 + x_1 x_2\),
with \(5\) piecewise linear regions, i.e., \(a_1\) equals to
\(\{2, -2, 5, -10, 0.5\}\) in the intervals defined by the sequence
\(\{0, 0.2, 0.4, 0.45, 0.5, 1\}\).  The effect of \(x_1\)\ is
\(f^{\mathtt{GT}}(x_1) = a_1 x_1\) and the heterogeneity
\(\sigma_2 = \sqrt{0.5}\).  As we observe in the top left of
Figure~\ref{fig:ex-synth-2-1}, RHALE splits in fine-grained bins the
intervals \([0.4, 0.45]\), \([0.45, 0.5]\) and unites in a single bin
most of the constant-effect regions, e.g.~region \([0.5,
1]\). Therefore RHALE's estimation is better than any fixed-size
binning in terms of both \(\mathcal{L}^{\mu}\) and
\(\mathcal{L}^{\sigma}\).

\paragraphb{Non-Linear Function.}
\begin{figure}[h]
  \centering
  \includegraphics[width=.23\textwidth]{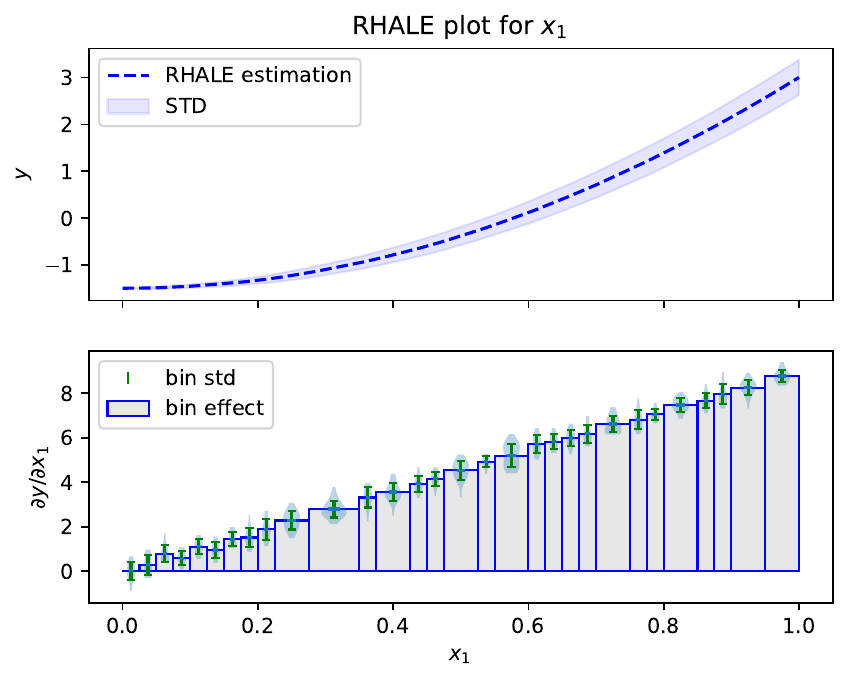}
  \includegraphics[width=.23\textwidth]{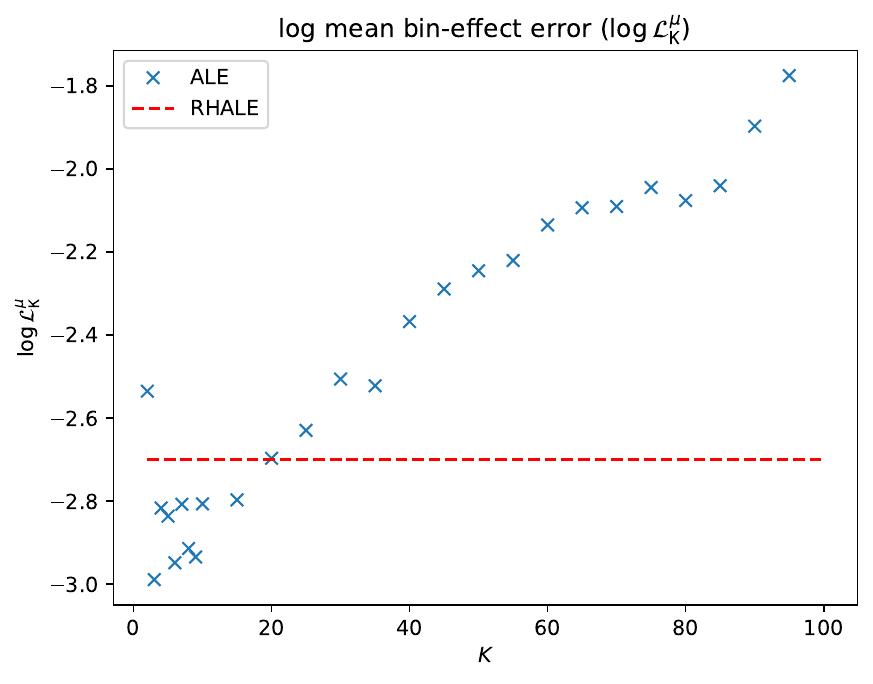}\\
  \includegraphics[width=.23\textwidth]{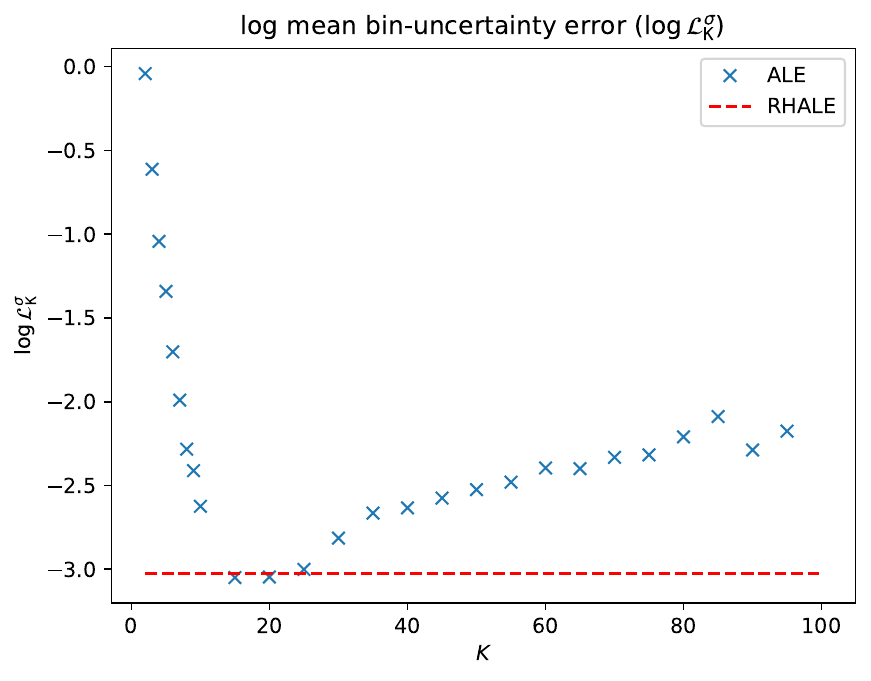}
  \includegraphics[width=.23\textwidth]{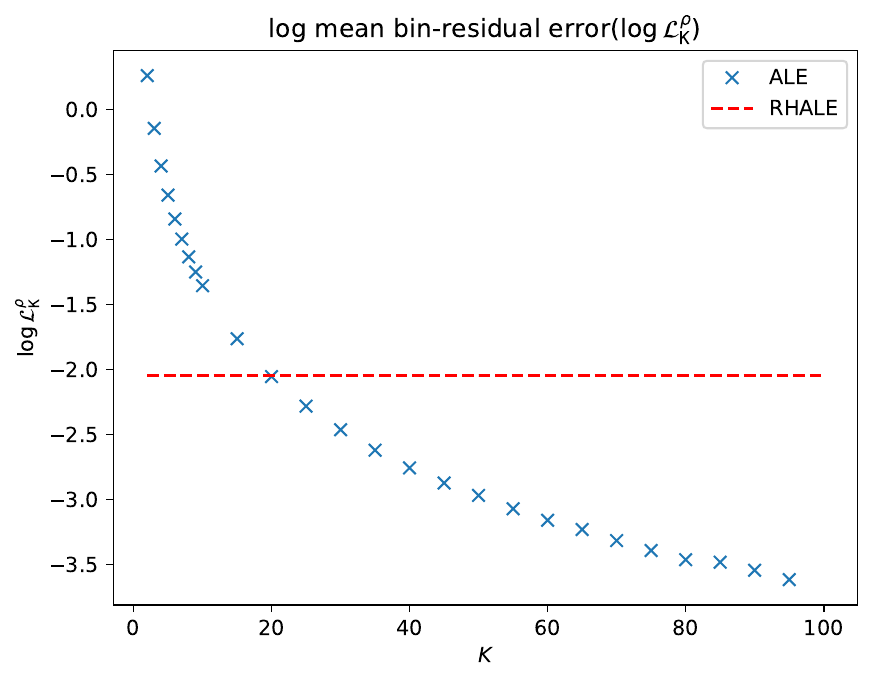}
  \caption{Bin-Splitting, non-linear function: RHALE's approximation
    (Top-Left). RHALE vs fixed-size approximations in terms of:
    \(\mathcal{L}^{\mu}\) (Top-Right), \(\mathcal{L}^{\sigma}\)
    (Bottom-Left), \(\mathcal{L}^{\rho}\) (Bottom-Right).}
  \label{fig:ex-synth-2-2}
\end{figure}
Here, the black-box function is
\(f(\mathbf{x}) = 4x_1^2 + x_2^2 + x_1 x_2\), so the effect of \(x_1\)
is \(f^{\mathtt{GT}}(x_1) = 4 x_1^2\) and the heterogeneity is
\(\sigma_2\). When using a wide binning (low \(K\)) there is an increase in the mean
residual error \(\mathcal{L}^{\rho}\) (bottom-right of
Figure~\ref{fig:ex-synth-2-2}), resulting in a biased approximation of
\(\sigma\). In contrast, narrow bins (high \(K\)) lead to a worse
approximation due to number of samples per bin. However, RHALE manages to
compromise these competing objectives and achieves an
(almost) optimal approximation of both \(\mu\) (top-right) and \(\sigma\)
(bottom-left), as illustrated in Figure~\ref{fig:ex-synth-2-2}.

\section{Real-world example}
\label{sec:real-world-example}

Here, since it is infeasible to access the
ground-truth FE, we simply demonstrate the usefulness of quantifying the
heterogeneity and the advantages of RHALE's approximation, on the
real-world California Housing dataset~\cite{pace1997sparse}.

\paragraphb{ML setup.}
The California Housing is a largely-studied dataset with approximately
\(20000\) training instances, making it appropriate for robust
approximation with large \(K\).  The dataset contains \(D=8\)
numerical features with characteristics about the building blocks of
California, e.g., latitude, longitude, population of the block or
median age of houses in the block.  The target variable is the median
value of the houses inside the block in dollars that ranges between
\([15, 500] \cdot 10^3\), with a mean value of
\(\mu_Y \approx 201 \cdot 10^3 \) and a standard deviation of
\(\sigma_Y \approx 110 \cdot 10^3\). We exclude instances with missing
or outlier values and we normalize all features to zero-mean and unit
standard deviation. We split the dataset into \(N_{tr} = 15639\)
training and \(N_{test} = 3910\) test examples (80/20 split) and we
fit a Neural Network with 3 hidden layers of 256, 128 and 36 units
respectively.  After 15 epochs using the Adam optimizer with learning
rate \(\eta = 0.02\), the model achieves a MAE of \(37 \cdot 10^3\)
dollars.

Below, we illustrate the feature effect for two features: latitude
\(x_2\) and median income \(x_8\).  The particular features cover the
main FE cases, e.g.~positive/negative trend and linear/non-linear
curve, and are therefore appropriate for illustration purposes.
Results for all features, along with details about the reprocessing,
training and evaluation parts are provided in the Appendix B2.

\paragraphb{Heterogeneity Quantification}
\begin{figure}[h]
  \centering
  \includegraphics[width=.23\textwidth]{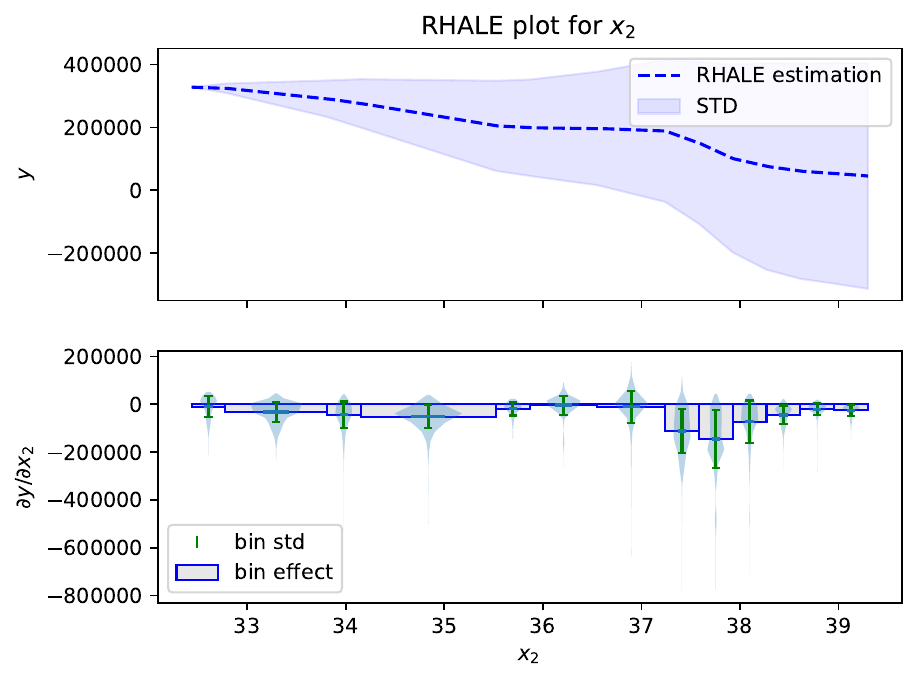}
  \includegraphics[width=.23\textwidth]{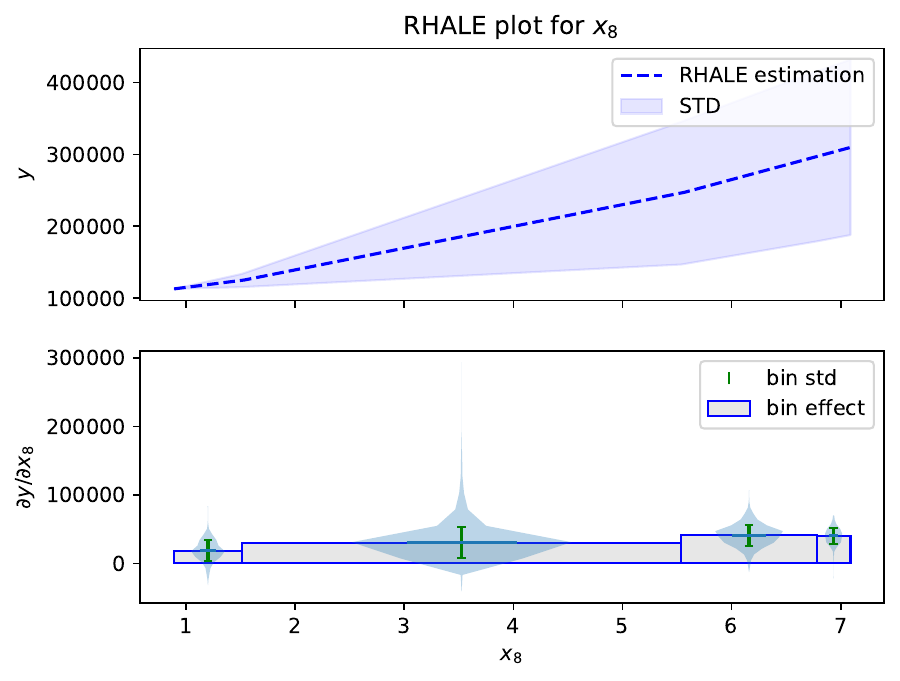}
  \caption{RHALE plot for features \(x_2\) (latitude) and \(x_6\)
    (median income). Apart from the average effects, i.e., negative
    for the \(x_2\) and positive for \(x_8\), the heterogeneity
    (\(\pm \mathtt{STD}\) and BIN-STD) shows that instance-level
    effects ara more heterogeneous on \(x_2\) case.}
  \label{fig:ex-real-1}
\end{figure}
Figure~\ref{fig:ex-real-1} illustrates the significance of RHALE's
heterogeneity quantification for comprehensive interpretation of
feature effects. We observe that both features exhibit significant
interactions with other features leading to high heterogeneity.
However, despite the high heterogeneity, we can confidently infer that
the (a) latitude of the house (\(x_2\)) negatively impacts the price,
and the (b) median income (\(x_8\)) has a positive influence on the
price, for almost all instances.

\begin{figure}[h]
  \centering
  \includegraphics[width=.23\textwidth]{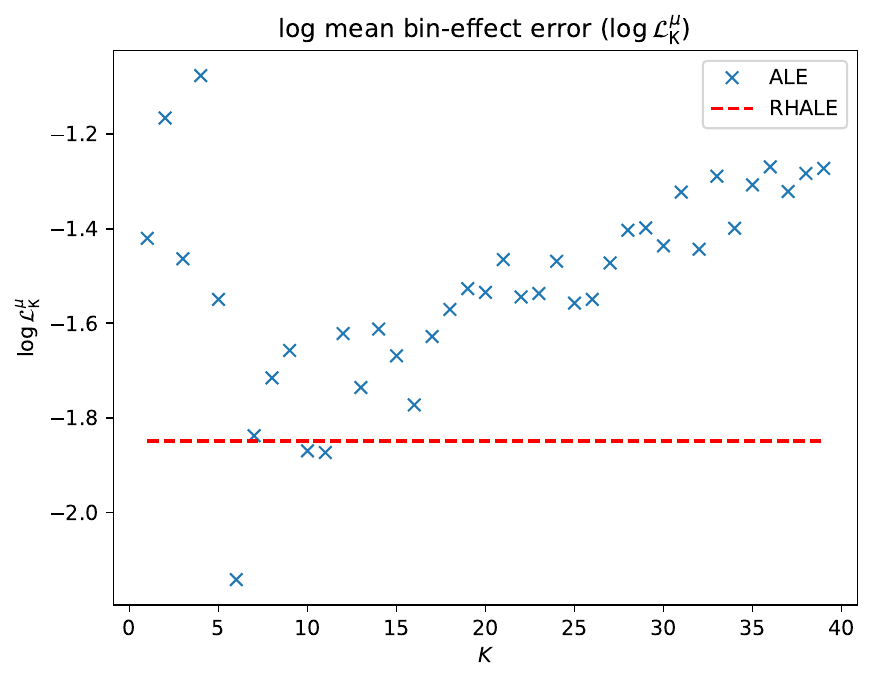}
  \includegraphics[width=.23\textwidth]{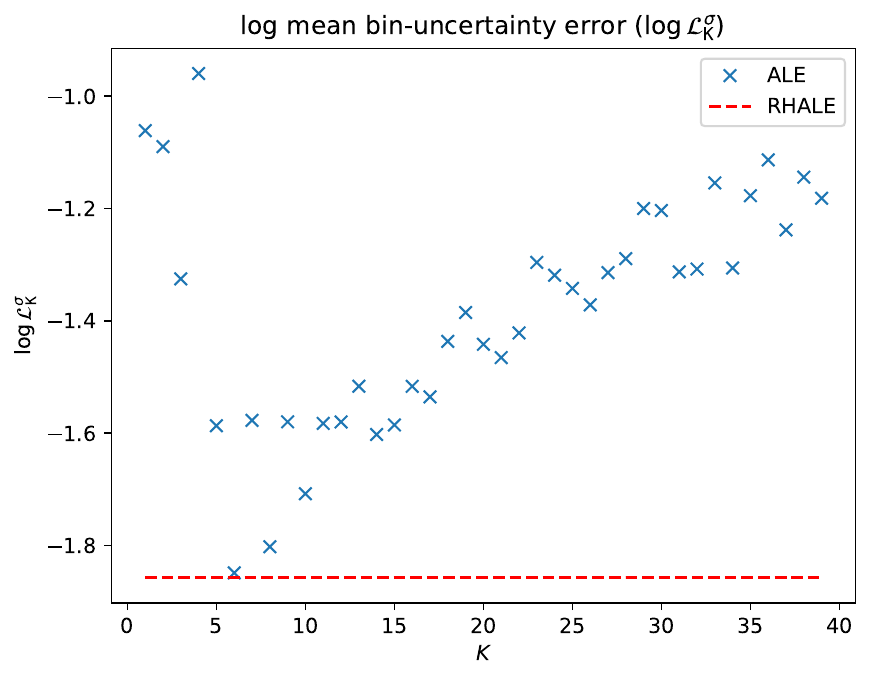}\\
  \includegraphics[width=.23\textwidth]{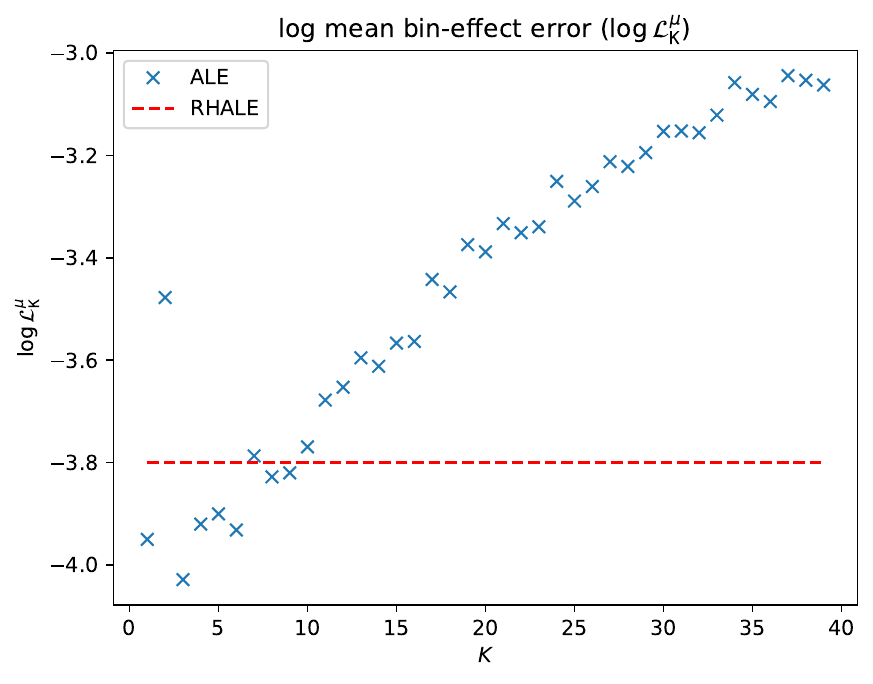}
  \includegraphics[width=.23\textwidth]{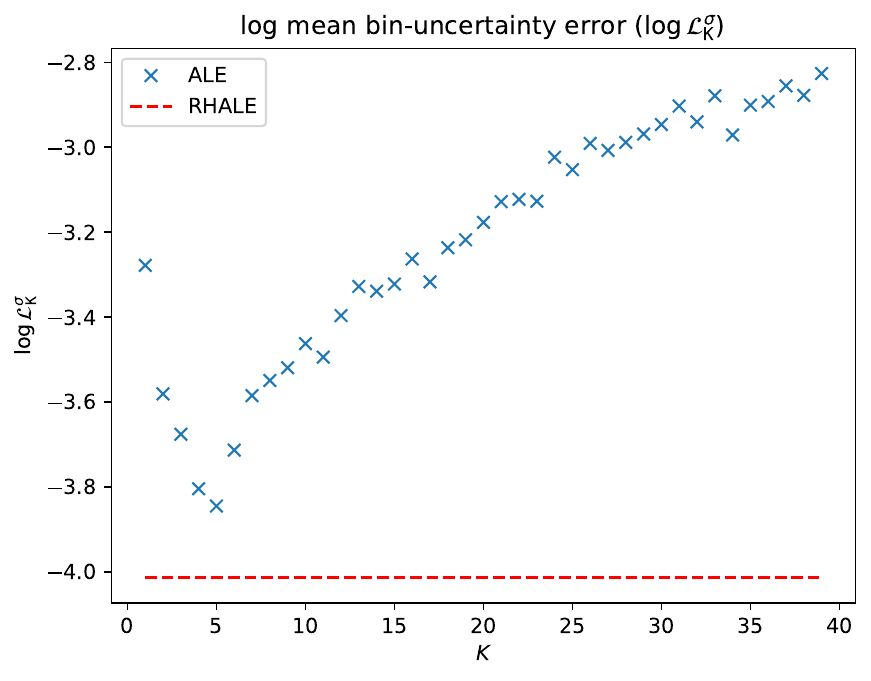}\\
  \caption{Lower is better. RHALE (red line) vs ALE fixed-size bins (blue crosses) in
    terms of \(\mathcal{L}^{\mu}\) (left column),
    \(\mathcal{L}^{\sigma}\) (right column) for features \(x_2\) (top)
    \(x_8\) (bottom). We observe that RHALE's estimation is better than (almost) any fixed-size alternative.}
  \label{fig:ex-real-2}
\end{figure}

\paragraphb{Bin Splitting}
We evaluate the robustness of RHALE approximation, following the same
methodology as described in
Section~\ref{subsec:simulation-examples-2}.  We consider as
ground-truth the effects computed on the entire training set
(\(N_{tr}=15639\)) with dense fixed-size bin-splitting
(\(K=80\)). Given the sufficient number of samples, we make the
hypothesis that the approximation with dense binning is close enough
to the ground truth. Next, we randomly select fewer samples,
\(N=1000\), and compare RHALE's approximation against fixed-size
approximation (for all \(K\)).  We repeat this process for \(t=30\)
independent runs and we report the mean values of
\(\mathcal{L}^{\mu}, \mathcal{L}^{\sigma}\).  In
Figure~\ref{fig:ex-real-2}, we observe that RHALE achieves accurate
approximations in all cases; \(\mathcal{L}^\mu\)
\(\mathcal{L}^\sigma\) are close to the best among the fixed-size
approximations.

\section{Conclusion and further work}
\label{sec:conclusion-and-further-work}

In this paper, we have introduced Robust and Heterogeneity-aware ALE (RHALE), a global feature effect method that addresses two major limitations of ALE. First, it quantifies the heterogeneity of local effects, which is essential for a complete interpretation of the feature effect. Second, it automates the bin-splitting process to improve the approximation of both the average effect and the heterogeneity. To achieve the latter, we proposed an automatic bin-splitting algorithm that balances estimation bias and variance by creating wider bins only when the underlying local effects are (near) constant. Our experiments on synthetic and real-world examples demonstrate RHALE's superiority over PDP-ICE, which struggles with correlated features, and traditional ALE, as automatic bin-splitting provides more accurate estimates than fixed-size splitting.

\paragraphb{Limitations.}
While the standard deviation of local effects is a good way to express the \textit{level} of heterogeneity, it is challenging to interpret the \textit{type} of heterogeneity. Therefore, we use violin plots to provide the distribution of local effects (type of heterogeneity), but their explanatory power is limited within each bin. At a global level, i.e., between the bins, the user can only determine the magnitude of heterogeneity. Finally, the automatic-binning algorithm comes with three hyperparameters, \(K_{\max}, \alpha, N_{\mathtt{PPB}}\). Although, their default values work well in most cases, on exceptional scenarios, such as a very small dataset, may need to be adjusted appropriately for an optimal bin splitting.

\clearpage

\ack

This work was supported by the XMANAI project (grant agreement No 957362), which has received funding by the European Regional Development Fund of the EU (EU 2020 Programme, ICT-38-2020 - Artificial intelligence for manufacturing).

\bibliography{bibliography}

\begin{thebibliography}{10}

\bibitem{apley2020visualizing}
Daniel~W Apley and Jingyu Zhu, `Visualizing the effects of predictor variables
  in black box supervised learning models', {\em Journal of the Royal
  Statistical Society: Series B (Statistical Methodology)}, {\bf 82}(4),
  1059--1086, (2020).

\bibitem{baniecki2021fooling}
Hubert Baniecki, Wojciech Kretowicz, and Przemyslaw Biecek, `Fooling partial
  dependence via data poisoning', {\em arXiv preprint arXiv:2105.12837},
  (2021).

\bibitem{britton2019vine}
Matthew Britton, `Vine: visualizing statistical interactions in black box
  models', {\em arXiv preprint arXiv:1904.00561}, (2019).

\bibitem{casalicchio2019visualizing}
Giuseppe Casalicchio, Christoph Molnar, and Bernd Bischl, `Visualizing the
  feature importance for black box models', in {\em Machine Learning and
  Knowledge Discovery in Databases: European Conference, ECML PKDD 2018,
  Dublin, Ireland, September 10--14, 2018, Proceedings, Part I 18}, pp.
  655--670. Springer, (2019).

\bibitem{freiesleben2022scientific}
Timo Freiesleben, Gunnar K{\"o}nig, Christoph Molnar, and Alvaro
  Tejero-Cantero, `Scientific inference with interpretable machine learning:
  Analyzing models to learn about real-world phenomena', {\em arXiv preprint
  arXiv:2206.05487}, (2022).

\bibitem{friedman2001greedy}
Jerome~H Friedman, `Greedy function approximation: a gradient boosting
  machine', {\em Annals of statistics},  1189--1232, (2001).

\bibitem{friedman2008predictive}
Jerome~H Friedman and Bogdan~E Popescu, `Predictive learning via rule
  ensembles', {\em The annals of applied statistics},  916--954, (2008).

\bibitem{gkolemis22}
Vasilis Gkolemis, Theodore Dalamagas, and Christos Diou, `Dale: Differential
  accumulated local effects for efficient and accurate global explanations',
  {\em arXiv preprint arXiv:2210.04542}, (2022).

\bibitem{goldstein2015peeking}
Alex Goldstein, Adam Kapelner, Justin Bleich, and Emil Pitkin, `Peeking inside
  the black box: Visualizing statistical learning with plots of individual
  conditional expectation', {\em journal of Computational and Graphical
  Statistics}, {\bf 24}(1),  44--65, (2015).

\bibitem{greenwell2018simple}
Brandon~M Greenwell, Bradley~C Boehmke, and Andrew~J McCarthy, `A simple and
  effective model-based variable importance measure', {\em arXiv preprint
  arXiv:1805.04755}, (2018).

\bibitem{Gromping2020MAEP}
Ulrike Grömping.
\newblock Model-agnostic effects plots for interpreting machine learning
  models, 03 2020.

\bibitem{herbinger2022repid}
Julia Herbinger, Bernd Bischl, and Giuseppe Casalicchio, `Repid: Regional
  effect plots with implicit interaction detection', in {\em International
  Conference on Artificial Intelligence and Statistics}, pp. 10209--10233.
  PMLR, (2022).

\bibitem{kim2016examples}
Been Kim, Rajiv Khanna, and Oluwasanmi~O Koyejo, `Examples are not enough,
  learn to criticize! criticism for interpretability', {\em Advances in neural
  information processing systems}, {\bf 29}, (2016).

\bibitem{koh2017understanding}
Pang~Wei Koh and Percy Liang, `Understanding black-box predictions via
  influence functions', in {\em International conference on machine learning},
  pp. 1885--1894. PMLR, (2017).

\bibitem{lundberg2018consistent}
Scott~M Lundberg, Gabriel~G Erion, and Su-In Lee, `Consistent individualized
  feature attribution for tree ensembles', {\em arXiv preprint
  arXiv:1802.03888}, (2018).

\bibitem{mehrabi2021survey}
Ninareh Mehrabi, Fred Morstatter, Nripsuta Saxena, Kristina Lerman, and Aram
  Galstyan, `A survey on bias and fairness in machine learning', {\em ACM
  Computing Surveys (CSUR)}, {\bf 54}(6),  1--35, (2021).

\bibitem{molnar2022}
Christoph Molnar, {\em Interpretable Machine Learning}, 2 edn., 2022.

\bibitem{molnar2020interpretable}
Christoph Molnar, Giuseppe Casalicchio, and Bernd Bischl, `Interpretable
  machine learning--a brief history, state-of-the-art and challenges', in {\em
  Joint European Conference on Machine Learning and Knowledge Discovery in
  Databases}, pp. 417--431. Springer, (2020).

\bibitem{molnar2021relating}
Christoph Molnar, Timo Freiesleben, Gunnar K{\"o}nig, Giuseppe Casalicchio,
  Marvin~N Wright, and Bernd Bischl, `Relating the partial dependence plot and
  permutation feature importance to the data generating process', {\em arXiv
  preprint arXiv:2109.01433}, (2021).

\bibitem{molnar2020model}
Christoph Molnar, Gunnar K{\"o}nig, Bernd Bischl, and Giuseppe Casalicchio,
  `Model-agnostic feature importance and effects with dependent features--a
  conditional subgroup approach', {\em arXiv preprint arXiv:2006.04628},
  (2020).

\bibitem{molnar2022general}
Christoph Molnar, Gunnar K{\"o}nig, Julia Herbinger, Timo Freiesleben, Susanne
  Dandl, Christian~A Scholbeck, Giuseppe Casalicchio, Moritz Grosse-Wentrup,
  and Bernd Bischl, `General pitfalls of model-agnostic interpretation methods
  for machine learning models', in {\em International Workshop on Extending
  Explainable AI Beyond Deep Models and Classifiers}, pp. 39--68. Springer,
  (2022).

\bibitem{pace1997sparse}
R~Kelley Pace and Ronald Barry, `Sparse spatial autoregressions', {\em
  Statistics \& Probability Letters}, {\bf 33}(3),  291--297, (1997).

\bibitem{ribeiro2016should}
Marco~Tulio Ribeiro, Sameer Singh, and Carlos Guestrin, `{"Why should i trust
  you?" Explaining the predictions of any classifier}', in {\em Proceedings of
  the 22nd ACM SIGKDD international conference on knowledge discovery and data
  mining}, pp. 1135--1144, (2016).

\bibitem{wiens2019no}
Jenna Wiens, Suchi Saria, Mark Sendak, Marzyeh Ghassemi, Vincent~X Liu, Finale
  Doshi-Velez, Kenneth Jung, Katherine Heller, David Kale, Mohammed Saeed,
  et~al., `Do no harm: a roadmap for responsible machine learning for health
  care', {\em Nature medicine}, {\bf 25}(9),  1337–1340, (2019).

\end{thebibliography}


\begin{thebibliography}{1}
\providecommand{\natexlab}[1]{#1}
\providecommand{\url}[1]{\texttt{#1}}
\expandafter\ifx\csname urlstyle\endcsname\relax
  \providecommand{\doi}[1]{doi: #1}\else
  \providecommand{\doi}{doi: \begingroup \urlstyle{rm}\Url}\fi

\bibitem[Apley and Zhu(2020)]{apley2020visualizing}
Daniel~W Apley and Jingyu Zhu.
\newblock Visualizing the effects of predictor variables in black box
  supervised learning models.
\newblock \emph{Journal of the Royal Statistical Society: Series B (Statistical
  Methodology)}, 82\penalty0 (4):\penalty0 1059--1086, 2020.

\end{thebibliography}
\end{document}


\maketitle



\appendix

\section{Theoretical Evidence}

In this Section, we provide proofs for the equations used in the main paper.

\subsection{Proof that \(\hat{\mu}(z_1, z_2)\) is an unbiased estimator of \( \mu(z_1, z_2)\)}
\label{sec:proof-1}

This proof is required for Theorem 1 (Section~\ref{sec:proof-2}).
We want to show that

\[\hat{\mu}(z_1, z_2) = \frac{1}{|\mathcal{S}|} \sum_{i: \mathbf{x}^i
  \in \mathcal{S}} f^s(\mathbf{x}^i) \]
is an unbiased estimator of:
\[\mu(z_1, z_2) = \frac{\int_{z_1}^{z_2} \mathbb{E}_{X_c|z} \left [
      f^s(z, X_c) \right ] \partial z}{z_2 - z_1} \]
under the assumptions that
(a) \(z\) follows a uniform distribution in
\([z_1, z_2]\), i.e., \(z \sim \mathcal{U}(z_1, z_2)\),
(b) \(\tilde{X}\) is a random variable with PDF
\(p(\tilde{\xb}) = p(\xc|z)p(z) = \frac{1}{z_2-z_1}p(\xc|z) \) and
(c) the points \(\mathbf{x}^i\) are i.i.d. samples from
\(p(\tilde{\xb})\).
We want to show that \(\mathbb{E}_{\tilde{X}} [\hat{\mu}(z_1, z_2)] = \mu(z_1, z_1)\).

\paragraph{Proof Description}
We show that (a) \(\mu(z_1, z_2) = \mathbb{E}_{\tilde{X}} [f^s(\tilde{X})]\) and
we use the fact that
(b) the population mean is an unbiased estimator of the expected value.

\paragraph{Proof}
\begin{align}
  \mu(z_1, z_2)
  = \frac{\int_{z_1}^{z_2} \mathbb{E}_{X_c|z} [f^s(z, X_c)] \partial z}{z_2 - z_1}
  = \mathbb{E}_{z \sim \mathcal{U}(z_1, z_2)}[\mathbb{E}_{X_c|z} [f^s(z, X_c)]]
  = \mathbb{E}_{\tilde{X}} [f^s(\tilde{X})]
  = \mathbb{E}_{\tilde{X}} [\hat{\mu}(z_1, z_2)]
\end{align}

\subsection{Proof that \(\hat{\sigma}^2(z_1, z_2)\) is an unbiased estimator of \(\sigma_*^2(z_1, z_2)\)}
\label{sec:proof-2}

This equation is used in Section 3.1 of the main paper.
We want to show that

\[\hat{\sigma}^2(z_1, z_2) = \frac{1}{|\mathcal{S}_k - 1|}
\sum_{i:\mathbf{x}^i \in \mathcal{S}_k} \left ( f^s(\mathbf{x}^i) -
  \hat{\mu}(z_1, z_2) \right )^2\]
is an unbiased estimator of

\[\sigma^2_*(z_1, z_2) = \frac{\int_{z_1}^{z_2} \mathbb{E}_{X_c|X_s=z}
  \left [ (f^s(z, X_c) - \mu(z_1, z_2) )^2 \right] \partial z}{z_2 -
  z_1} \]
under the assumptions that (a) \(z\) follows a uniform distribution
in \([z_1, z_2]\), i.e., \(z \sim \mathcal{U}(z_1, z_2)\), (b)
\(\tilde{X}\) is a random variable with PDF
\(p(\tilde{\xb}) = p(\xc|z)p(z) = \frac{1}{z_2-z_1}p(\xc|z) \) and (c)
the points \(\mathbf{x}\) are i.i.d. samples from
\(p(\tilde{\xb})\). We want to show that
\(\mathbb{E}_{\tilde{X}} [\hat{\sigma}^2(z_1, z_2)] = \sigma_*^2(z_1, z_1)\).

\paragraph{Proof Description}

We show (a) that
\(\sigma^2_*(z_1, z_2) = \mathbb{E}_{\tilde{X}} \left [ (f^s(\tilde{X}) - \mathbb{E}_{\tilde{X}}[\hat{\mu}(z_1, z_2)] )^2 \right]\) and
then (b) we use the fact that the sample variance is an unbiased
estimator of the distribution variance.

\paragraph{Proof}

\begin{align}
  \sigma^2_*(z_1, z_2) & = \frac{\int_{z_1}^{z_2} \mathbb{E}_{X_c|z}
                         \left [ (f^s(z, X_c) - \mu(z_1, z_2) )^2 \right] \partial z}{z_2 -
                         z_1} \\
                       & = \mathbb{E}_{z \sim \mathcal{U}(z_1, z_2)}\mathbb{E}_{X_c|z}
                         \left [ (f^s(z, X_c) - \mu(z_1, z_2) )^2 \right] \\
                       & = \mathbb{E}_{\tilde{X}} \left [ (f^s(\tilde{X}) - \mu(z_1, z_2) )^2 \right]\\
                       & = \mathbb{E}_{\tilde{X}} \left [ (f^s(\tilde{X}) - \mathbb{E}_{\tilde{X}}[\hat{\mu}(z_1, z_2)] )^2 \right]\\
                       & = \mathbb{E}_{\tilde{X}} \left [ \hat{\sigma}^2(z_1, z_2)  \right]
  \end{align}

\subsection{Proof Of Theorem 1}
\label{sec:proof-3}

\begin{customthm}{3.1}
  \label{sec:theorem-1-app}
  If we define (a) the residual \(\rho(z)\) as the difference between
  the expected effect at \(z\) and the bin effect, i.e,
  \(\rho(z) = \mu(z) - \mu(z_1, z_2)\), and (b)
  \(\mathcal{E}(z_1, z_2)\) as the mean squared residual of the bin,
  i.e.,
  \(\mathcal{E}(z_1, z_2) = \frac{\int_{z_1}^{z_2}\rho^2(z) \partial
    z}{z_2 - z_1}\), then it holds
\begin{equation}
    \label{eq:bin-uncertainty-proof}
 \sigma_*^2(z_1, z_2) = \sigma^2(z_1, z_2) + \mathcal{E}^2(z_1, z_2)
\end{equation}
\end{customthm}
We want to show that
\(\sigma_*^2(z_1, z_2) = \sigma^2(z_1, z_2) + \mathcal{E}^2(z_1,
z_2)\), where (a) the bin-error \(\mathcal{E}^2(z_1, z_2)\) is the
mean squared residual of the bin, i.e.
\(\mathcal{E}^2(z_1, z_2) = \frac{\int_{z_1}^{z_2}\rho^2(z) \partial
  z}{z_2 - z_1}\) and (b) the residual \(\rho(z)\) is the difference
between the expected effect at \(z\) and the bin effect, i.e
\(\rho(z) = \mu(z) - \mu(z_1, z_2)\).

\paragraph{Proof Description}

We use that \(\forall z \in [z_1, z_2]\), it holds that \(\mu(z_1, z_2) = \mu(z) - \rho(z)\) and then we split the terms appropriately to complete the proof.

\paragraph{Proof}

\begin{align}
  \sigma_*^2(z_1, z_2) &= \frac{1}{z_2 - z_1}\int_{z_1}^{z_2} \mathbb{E}_{X_c|z} \left [ \left( f^s( z, X_c) - \mu(z_1, z_2) \right)^2 \right] \partial z \\
                       &= \frac{1}{z_2 - z_1} \int_{z_1}^{z_2} \mathbb{E}_{X_c|z} \left [ \left ( f^s(z, X_c) - \mu(z) + \rho(z) \right )^2 \right] \partial z \\
                       &= \frac{1}{z_2 - z_1} \int_{z_1}^{z_2} \mathbb{E}_{X_c|z} \left [ (f^s(z, X_c) - \mu(z) )^2 + \rho(z)^2 + 2(f^s(z, X_c) -\mu(z))\rho(z) \right ] \partial z \\
                       &= \frac{1}{z_2 - z_1} \int_{z_1}^{z_2} \left (
                         \underbrace{\mathbb{E}_{X_c|z} \left [ (f^s(z, X_c) - \mu(z) )^2 \right ]}_{\sigma^2(z)}  +
                         \underbrace{\mathbb{E}_{X_c|z} \left [ \rho^2(z) \right]}_{\rho^2(z)} +
                         2 (\underbrace{\mathbb{E}_{X_c|z} \left [ (f^s(z, X_c)   \right ]}_{\mu(z)} - \mu(z)) \rho(z) )\right )  \partial z \\
                       &= \underbrace{\frac{1}{z_2 - z_1} \int_{z_1}^{z_2} \sigma^2(z) \partial z}_{\sigma^2(z_1, z_2)} + \underbrace{\frac{1}{z_2 - z_1} \int_{z_1}^{z_2} \rho^2(z) \partial z}_{\mathcal{E}^2(z_1, z_2)}\\
                       &= \sigma^2(z_1, z_2) + \mathcal{E}^2(z_1, z_2)
\end{align}

\subsection{Proof Of Corollary 2}

We want to show that, \textit{if a bin-splitting \(\mathcal{Z}\) minimizes the accumulated error, then it also minimizes
  \(\sum_{k=1}^K\sigma_*^2(z_1, z_2) \Delta z_k \)}.
In mathematical terms, we want to show that:

\[ \mathcal{Z}^* = \argmin_{\mathcal{Z}} \sum_{k=1}^K \sigma_*^2(z_{k-1}, z_k) \Delta z_k \Leftrightarrow \mathcal{Z}^* = \argmin_{\mathcal{Z}} \sum_{k=1}^K \mathcal{E}^2(z_{k-1}, z_k) \Delta z_k \]

\paragraph{Proof Description}

The key-point for the proof is that the term \(\sum_{k=1}^K \sigma^2(z_{k-1}, z_k) \Delta z_k \) is
independent of the bin partitioning \(\mathcal{Z}\). In Eq.(16) we use Eq.8 of the main paper.

\paragraph{Proof}

\begin{align}
  \mathcal{Z}^* &= \argmin_{\mathcal{Z}} \sum_{k=1}^K \sigma_*^2(z_{k-1}, z_k) \Delta z_k \\
                &= \argmin_{\mathcal{Z}} \left [ \sum_{k=1}^K (\sigma^2(z_{k-1}, z_k) + \mathcal{E}^2(z_{k-1}, z_k)) \Delta z_k \right ] \\
                & = \argmin_{\mathcal{Z}} \left [ \sum_{k=1}^K \left ( \frac{\Delta z_k}{\Delta z_k} \int_{z_{k-1}}^{z_k} \sigma^2(z) \partial z   + \mathcal{E}^2(z_{k-1}, z_k) \Delta z_k \right ) \right ] \\
                & = \argmin_{\mathcal{Z}} \left [ \underbrace{\int_{z_0}^{z_K} \sigma^2(z) \partial z}_{\text{independent of } \mathcal{Z}}   + \sum_{k=1}^K\mathcal{E}^2(z_{k-1}, z_k) \Delta z_k) \right ] \\
                & = \argmin_{\mathcal{Z}} \sum_{k=1}^K\mathcal{E}^2(z_{k-1}, z_k) \Delta z_k
  \end{align}

\subsection{Dynamic Programming}

We denote with \(i \in \{0, \ldots, K_{max}\}\) the index of point
\(x_i\), as defined at Section 3.2 of the main paper,
and with \(z_j\) and \(z_{j+1}\) the
chosen limits (out of the values \(x_i\)) for bin \(j\). The states of
the problem are then represented by matrices \(\mathcal{C}(i, j)\) and
\(\mathcal{I}(i,j)\). \(\mathcal{C}(i, j)\) is the cost of setting
\(z_{j+1}=x_i\), i.e., the cost of setting the right limit of the
\(j\)-th bin to \(x_i\), and is computed by the recursive function:

\begin{equation}
  \label{eq:recursive_cost_supp}
  \mathcal{C}(i,j) =
  \begin{cases}
    \mathrm{min}_{i \in \{0, \ldots, K_{max}\}} \left [ C(i, j-1) + \mathcal{B}(i, j) \right ], & \text{if } j > 0 \\
    \mathcal{B}(i, j) & \text{if } j = 0\\
    \end{cases}
\end{equation}

\(\mathcal{I}(i,j)\) is an index matrix indicating the selected values
\(z_j\), i.e., the values indicating the right limit of \(j-1\)
bins. In other words, \(z_{j} = x_{\mathcal{I}(i,j)}\).
The value of \(\mathcal{I}(i,j)\) is given by
\( \mathcal{I}(i,j) = \mathrm{argmin}_{i \in \{0, \ldots, K_{max}\}}
\left [ C(i, j-1) + \mathcal{B}(i, j) \right ]\).
Note that although
this procedures always selects \(K_{max} + 1\) values for \(z_j\),
some of them may be the same point corresponding to zero-width
bins.
These are dropped when choosing the optimal bin limits
\(\mathcal{Z}\). Algorithm~\ref{alg:cap} presents the use of dynamic programming to solve the optimization
problem of Eq.13.

\begin{algorithm}
\caption{Algorithm for solving the optimization problem with dynamic programming }\label{alg:cap}
\begin{algorithmic}
\Require $\mathcal{B}(i,j)$: function that gives the cost of bin \([x_i, x_j)\), \(K_{max}\): max number of bins
\Ensure $\mathcal{Z}$: the optimal partitioning
\State $\mathcal{C}(i, j)  = + \infty, \forall i,j $ \Comment{Initiate the cost matrix with \(+ \infty\)}
\State $\mathcal{I}(i, j)  = 0, \forall i,j $ \Comment{Initiate the index matrix with \(0\)}
\State $\mathcal{C}(i,0)  = \mathcal{B}(0, i) \forall i$ \Comment{Set cost of the first bin}
\For{\(j = 0, \ldots, K_{max}-1 \)}
    \For{\(i = 0, \ldots, K_{max} \)}
        \For{\(k = 0, \ldots, K_{max} \)}
            \State{\(L(k) = \mathcal{C}(k, j - 1) + \mathcal{B}(k, j)\)}
        \EndFor
        \State \(\mathcal{C}(i,j) = \min_k L(k) \)
        \State \(\mathcal{I}(i,j) = \argmin_k L(k) \)
    \EndFor
\EndFor
\State \(Z(j) = 0 \forall j = \{0, \ldots, K_{max}\}\) \Comment{Initialize list with limits}
\State \(Z(0) = 0\), \(Z(K_{max}) = K_{max}\),   \Comment{First and last limit are always the same}
\For{\(j = K_{max}-1, \ldots, 1 \)}
    \State \(Z(j) = \mathcal{I}(j, Z(j+1)) \) \Comment{ Follow the inverse indexes}
\EndFor
\State Invert \(Z\) and drop \(Z\) items that show to the same point
\State \(\mathcal{Z} \leftarrow {x_{min}} + Z(j) \Delta X_{min} \) \Comment{Convert indexes to points}
\end{algorithmic}
\end{algorithm}

\section{Empirical Evaluation}

\subsection{Running Example}
\label{sec:running-example}

In the running example, the data generating distribution is
$p(\xb) = p(x_1)p(x_2)p(x_3|x_1)$, where
$p(x_1) = \frac{5}{6} \mathcal{U}(x_1; -0.5, 0) + \frac{1}{6}\mathcal{U}(x_1; 0, 0.5)$,
$p(x_2) = \mathcal{N}(x_2; \mu_2 = 0, \sigma_2=2)$ and
$p(x_3) = \mathcal{N}(x_3; \mu_3 = x_1, \sigma_3=0.01)$.
So, $x_1$ is highly correlated with $x_3$, while $x_2$ is independent from both $x_1$ and $x_3$.
The black-box function is:
\begin{equation}
  \label{eq:target_function-1}
  f(\xb) = \underbrace{\sin(2 \pi x_1) (\mathbbm{1}_{x_1 < 0} - 2 \mathbbm{1}_{x_3 < 0})}_{g_1(\xb)}
  + \underbrace{x_1 x_2}_{g_2(\xb)} + \underbrace{x_2}_{g_3(\xb)}
\end{equation}

\paragraph{Ground truth effect.}

For $g_1(\xb)$, $x_1 \approx x_3$ so
$\mathbbm{1}_{x_1 < 0} - 2 \mathbbm{1}_{x_3 < 0} = - \mathbbm{1}_{x_1 < 0}$ and therefore
$g_1(x_1) = - \sin(2 \pi x_1) \mathbbm{1}_{x_1 < 0}$.
For $g_2(\xb)$, $x_2$ is independent from $x_1$,
so $\mathbb{E}_{x_2|x_1} [x_1x_2] = \mathbb{E}_{x_2} [x_1 x_2] = x_1 \mathbb{E}_{x_2} [x_2] = 0$
and therefore $g_2(x_1) = 0$.
For $g_3(\xb)$, it does not include $x_1$, so $g_3(x_1) = 0$.
Therefore, the ground truth feature effect is
\begin{equation}
    f^{\mathtt{GT}}(x_1) = - \sin(2 \pi x_1) \mathbbm{1}_{x_1 < 0}
\end{equation}

\paragraph{Ground truth heterogeneity.}

For the heterogeneity, it is not easy to compute the ground truth, because each method
defines and visualizes it in a different way.
However, we use the fact that the heterogeneity is induced by the variability of the interaction terms.
For $g_1(\xb)$, $x_1 \approx x_3$ so
$\mathbbm{1}_{x_1 < 0} - 2 \mathbbm{1}_{x_3 < 0} = - \mathbbm{1}_{x_1 < 0}$ and therefore
$g_1$ does not introduce variability.
The variability of $g_3(\xb)$ is also zero.
The only term with variability is $g_2(\xb) = x_1 x_2$.
Since $x_1, x_2$ are independent the effect of this term varies according to the variation of $x_2$
that has a standard deviation of $\sigma_2$.
Therefore, independently of how each method computes the heterogeneity, the user should be able to
understand a variation of $\sigma_2$ on the local effects.

\paragraph{RHALE.} We compute in an analytic form the feature effect $f^{\mathtt{RHALE}}(x_1)$ and the
heterogeneity $\sigma(z)$ for the RHALE method.

\begin{align}
  f^{\mathtt{RHALE}}(x_1) &=
     \int_{x_{1,min}}^{x_1} \mathbb{E}_{x_2, x_3|z} \left [ \frac{\partial f}{\partial x_1}(z, x_2, x_3) \right ] \partial z\\
  &= \int_{x_{1,min}}^{x_1} ( \mathbb{E}_{x_3|z} [ 2 \pi z \cos(2 \pi z)  (\mathbbm{1}_{z < 0} - 2\mathbbm{1}_{x_3 < 0}) ] + \underbrace{\mathbb{E}_{x_2|z} [x_2]}_{0} ) \partial z\\
  &= \int_{x_{1,min}}^{x_1} 2 \pi z \cos(2 \pi z) \mathbb{E}_{x_3|z} [(\mathbbm{1}_{z < 0} - 2\mathbbm{1}_{x_3 < 0}) ] \partial z \\
  &\approx \int_{x_{1,min}}^{x_1} \underbrace{2 \pi z \cos(2 \pi z) (-\mathbbm{1}_{z < 0})}_{\mu(z)} \partial z \\
  &\approx  - \sin(2 \pi x_1) \mathbbm{1}_{x_1 < 0}
\end{align}

\begin{align}
  \sigma^2(z) &= \mathbb{E}_{x_2, x_3|z}\left [ \left (\frac{\partial f}{x_1} (z, x_2, x_3) - \mu(z) \right )^2 \right ]\\
  &= \mathbb{E}_{x_2, x_3|z} \left [ \left (2 \pi z \cos(2 \pi z)  (\mathbbm{1}_{z < 0} - 2\mathbbm{1}_{x_3 < 0}) + x_2 - 2 \pi z \cos(2 \pi z) (-\mathbbm{1}_{z < 0}) \right )^2 \right ] \\
  &= \mathbb{E}_{x_2, x_3|z} \left [ \left (2 \pi z \cos(2 \pi z)  (2 \mathbbm{1}_{z < 0} - 2\mathbbm{1}_{x_3 < 0}) + x_2 \right )^2 \right ] \\
  &= (4 \pi z \cos(2 \pi z))^2 \mathbb{E}_{x_3|z} [ (\mathbbm{1}_{z < 0} - \mathbbm{1}_{x_3 < 0})^2 ] + \mathbb{E}_{x_2|z} [x_2^2] + \mathbb{E}_{x_2, x_3|z} [4 \pi z \cos(2 \pi z) (\mathbbm{1}_{z < 0} - \mathbbm{1}_{x_3 < 0}) x_2]\\
  &= (4 \pi z \cos(2 \pi z))^2 \mathbb{E}_{x_3|z} [ (\mathbbm{1}_{z < 0} - \mathbbm{1}_{x_3 < 0})^2 ] + \sigma_2^2 + \mathbb{E}_{x_2|z}[x_2] \underbrace{\mathbb{E}_{x_3|z} [4 \pi z \cos(2 \pi z) (\mathbbm{1}_{z < 0} - \mathbbm{1}_{x_3 < 0})]}_{0}\\
  &= (4 \pi z \cos(2 \pi z))^2 \mathbb{E}_{x_3|z} [ (\mathbbm{1}_{z < 0} + \mathbbm{1}_{x_3 < 0} - 2\mathbbm{1}_{z < 0} \mathbbm{1}_{x_3 < 0} ] + \sigma_2^2 \\
  &= (4 \pi x_1 \cos(2 \pi x_1))^2 (2 \mathbbm{1}_{z < 0} - 2\mathbbm{1}_{z < 0}) + \sigma_2^2 \\
    &= \sigma_2^2
\end{align}

\paragraph{PDP-ICE.} We compute in an analytic form the feature effect $f^{\mathtt{PDP}}(x_1)$ and the
heterogeneity heterogeneity visualized by $f^{\mathtt{ICE}}(x_1)$.

The PDP effect uses

\begin{align}
    \label{eq:toy-example-pdp-effect}
  f^{\mathtt{PDP}}(x_1) &= \mathbb{E}_{x_2, x_3}[f(\xb)] \\
  &= \sin(2 \pi x_1) \mathbb{E}_{x_3} \left [ \mathbbm{1}_{x_1 < 0} - 2\mathbbm{1}_{x_3 < 0} \right ] + \mathbb{E}_{x_2} [x_1 x_2] + \mathbb{E}_{x_2} [x_2] \\
  &= \sin(2 \pi x_1) (\mathbbm{1}_{x_1 < 0} - 2 \mathbb{E}_{x_3} [\mathbbm{1}_{x_3 < 0}]) + \underbrace{x_1 \mathbb{E}_{x_2} [x_2]}_{0} + \underbrace{\mathbb{E}_{x_2} [x_2]}_{0} \\
  &= \sin(2 \pi x_1) \left ( \mathbbm{1}_{x_1 < 0} - 2 \int_{-0.5}^{0.5} \mathbbm{1}_{x_3 < 0} p(x_3) \partial x_3 \right )\\
  &= \sin(2 \pi x_1) \left ( \mathbbm{1}_{x_1 < 0} - 2 \int_{-0.5}^{0} 2 \frac{5}{6} \mathbbm{1}_{x_3 < 0}  \partial x_3 + \int_{0}^{0.5} 2 \frac{1}{6} \mathbbm{1}_{x_3 < 0}  \partial x_3 \right ) \\
  &= \sin(2 \pi x_1) \left ( \mathbbm{1}_{x_1 < 0} - 2 \frac{5}{6} \right )
\end{align}

For the ICE plots:

\begin{align}
    \label{eq:toy-example-ice-effect}
  f^{\mathtt{ICE}}(x_1^i) &= \sin(2 \pi x_1) (\mathbbm{1}_{x_1 < 0} - 2\mathbbm{1}_{x_3^i < 0} ) + x_1 x_2^i + x_2^i \\
    &= \sin(2 \pi x_1) (\mathbbm{1}_{x_1 < 0} - 2\mathbbm{1}_{x_3^i < 0} ) + x_1 x_2^i + c
\end{align}

So if $x_3^i < 0$, which happens in almost $\frac{5}{6}$ of the instances, then $f^{\mathtt{ICE}}(x_1^i)(x_1) = - \sin(2 \pi x_1) + x_1 x_2^i + c$,
and in almost $\frac{1}{6}$ of the instances, $f^{\mathtt{ICE}}(x_1^i)(x_1) = \sin(2 \pi x_1) + x_1 x_2^i + c$.

\paragraph{Discussion.}

\begin{figure}
    \centering
\begin{subfigure}{.32\textwidth}
  \centering
  \includegraphics[width=1\textwidth]{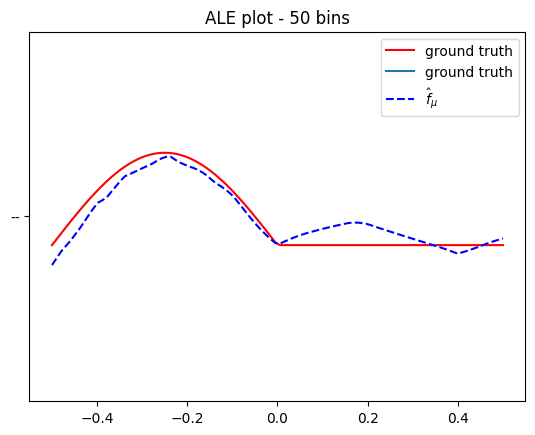}
  \caption{ALE plot (50 fixed-size bins)}
  \label{fig:concept-figure-subfig-1-app}
\end{subfigure}%
\begin{subfigure}{.32\textwidth}
  \centering
  \includegraphics[width=1\textwidth]{concept_figure/exp_1_pdp_ice_0}
  \caption{PDP-ICE plot}
  \label{fig:concept-figure-subfig-2-app}
\end{subfigure}
\begin{subfigure}{.32\textwidth}
  \centering
  \includegraphics[width=1\textwidth]{concept_figure/exp_1_rhale_0}
  \caption{RHALE plot}
  \label{fig:concept-figure-subfig-3-app}
\end{subfigure}
\caption{Feature effect of $x_1$ on the example defined by Equation~\ref{eq:target_function-1}.
ALE does not quantify the heterogeneity and fixed-size splitting leads to a bad estimation.
PDP-ICE plots fail in both main effect and heterogeneity, failing to capture feature correlations.
RHALE, on the other hand, provides a robust estimation of the main effect and the heterogeneity.}
\label{fig:concept-figure-app}
\end{figure}

The derivations above are reflected in Figure~\ref{fig:concept-figure-app}.
We observe that PDP and ICE provide misleading explanations which are \emph{not} due
to some approximation error, e.g., due to limited samples.
As shown by Equation~\ref{eq:toy-example-pdp-effect} and Equation~\ref{eq:toy-example-ice-effect}
PDP and ICE systematically produce misleading explanations~\citep{apley2020visualizing} for the feature effect and the heterogeneity
in cases of correlated features.
In contrast, we confirm our previous knowledge that ALE handles well these cases and
we observe that the deviation from the ground is only due to approximation issues,
which are addressed by RHALE.

\subsection{Simulation Study}
\label{sec:simulation-study}

The data generating distribution is
\(p(\mathbf{x}) = p(x_3)p(x_2|x_1)p(x_1)\), where
\(x_1 \sim \mathcal{U}(0,1)\), \(x_2 = x_1 + \epsilon \), where $\epsilon \sim \mathcal{N}(0, 0.01)$ is a small additive, noise and
\(x_3 \sim \mathcal{N}(0, \sigma_3^2 = \frac{1}{4})\). The predictive function is:
\begin{equation}
  \label{eq:synth-ex-1-function-app}
  f(\mathbf{x}) = \underbrace{\alpha f_2(\xb)}_{g_3(\xb)} + \underbrace{f_1(\xb) \mathbbm{1}_{f_1(\xb) \leq \frac{1}{2}}}_{g_1(\xb)} + \underbrace{(1 - f_1(\xb)) \mathbbm{1}_{\frac{1}{2} < f_1(\xb) < 1}}_{g_2(\xb)}
\end{equation}
where \(f_1(\mathbf{x}) = a_1 x_1 + a_2 x_2\) is a linear combination of $x_1, x_2$, and
\(f_2(\mathbf{x}) = x_1 x_3\) interacts the non-correlated features $x_1, x_3$.
We evaluate the effect computed by RHALE and PDP-ICE in three cases; (a)
without interaction (\(\alpha=0\)) and equal weights (\(a_1=a_2\)),
(b) without interaction (\(\alpha=0\)) and different weights
(\( a_1 \neq a_2 \)) and (c) with interaction (\(\alpha > 0\)) and
equal weights (\(a_1=a_2\)).

\paragraph{Ground truth for case (a)}

In this case, the weights are $a_1 = a_2 = 1$ and there is no interaction term $\alpha=0$).
Therefore:

\begin{equation}
  \label{eq:case-a}
  f(\mathbf{x}) = f_1(\xb) \mathbbm{1}_{f_1(\xb) \leq \frac{1}{2}} + (1 - f_1(\xb)) \mathbbm{1}_{\frac{1}{2} < f_1(\xb) < 1}
\end{equation}
where $f_1(\xb) = x_1 + x_2$.
For the ground truth feature effect, we use the fact that $x_1 \approx x_2$,
therefore knowing only the value of $x_1$ we can automatically infer the value of $x_2$
and therefore the value of $f_1(\xb)$.
For example, when $0 \leq x_1 \leq \frac{1}{4}$ then $0 \leq f_1(\xb) \leq \frac{1}{2}$ and,
therefore, $f_1(x_1) = a_1 x_1$.
In a similar way, we compute the effect of $x_2$.
The effect of $x_3$ is zero.
\begin{align}
    \label{eq:case-a-feat-1}
    f^{\mathtt{GT}}(x_1) &= x_1 \mathbbm{1}_{0 \leq x_1 \leq \frac{1}{4}} + \left ( \frac{1}{4} - x_1 \right ) \mathbbm{1}_{\frac{1}{4} < x_1 < \frac{1}{2}} \\
    f^{\mathtt{GT}}(x_2) &= x_2 \mathbbm{1}_{0 \leq x_2 \leq \frac{1}{4}} + \left ( \frac{1}{4} - x_2 \right ) \mathbbm{1}_{\frac{1}{4} < x_2 < \frac{1}{2}} \\
    f^{\mathtt{GT}}(x_3) &= 0
\end{align}
The heterogeneity is zero for all features because the heterogeneity is induced by the variability
of the interaction terms and, since, $x_1 \approx x_2$, the terms
$\mathbbm{1}_{f_1(\xb) \leq \frac{1}{2}}$ and $\mathbbm{1}_{\frac{1}{2} < f_1(\xb) \leq 1}$, do not vary.
\begin{figure}
    \label{fig:case-a}
    \includegraphics[width=0.3\linewidth]{example_1/dale_feat_0}
    \includegraphics[width=0.3\linewidth]{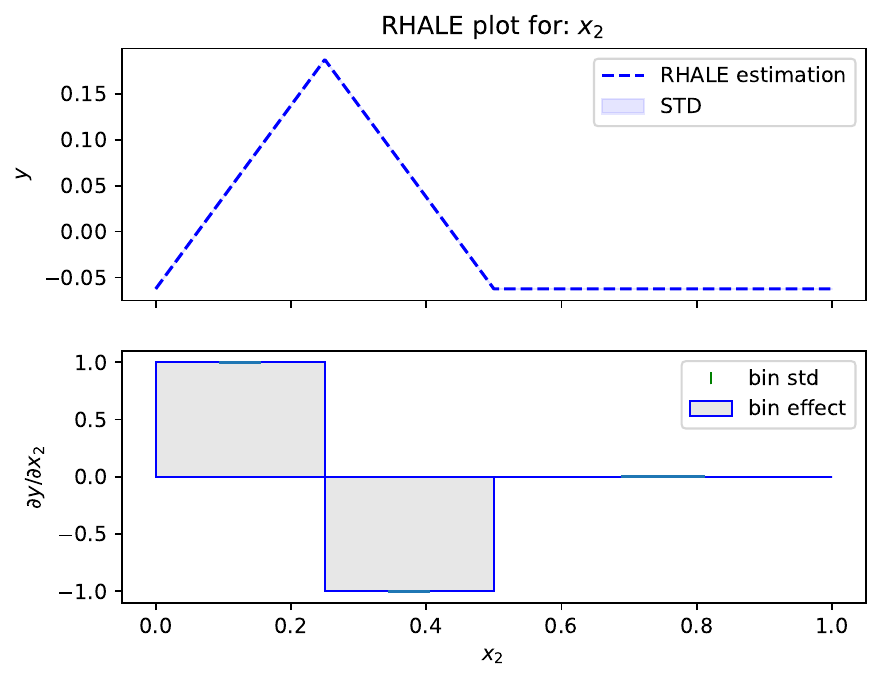}
    \includegraphics[width=0.3\linewidth]{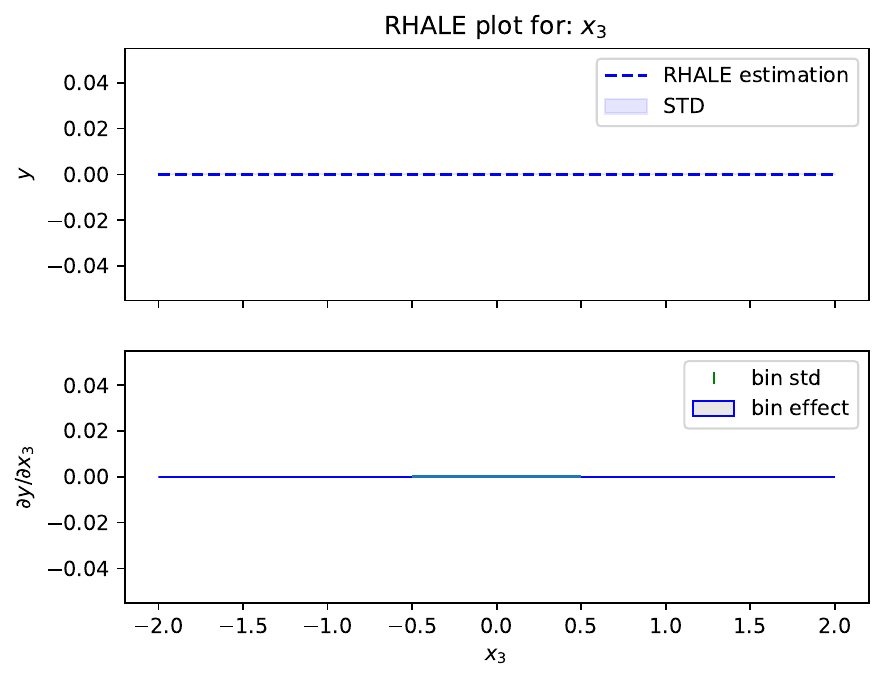}\\
    \includegraphics[width=0.3\linewidth]{example_1/pdp_ice_feat_0}
    \includegraphics[width=0.3\linewidth]{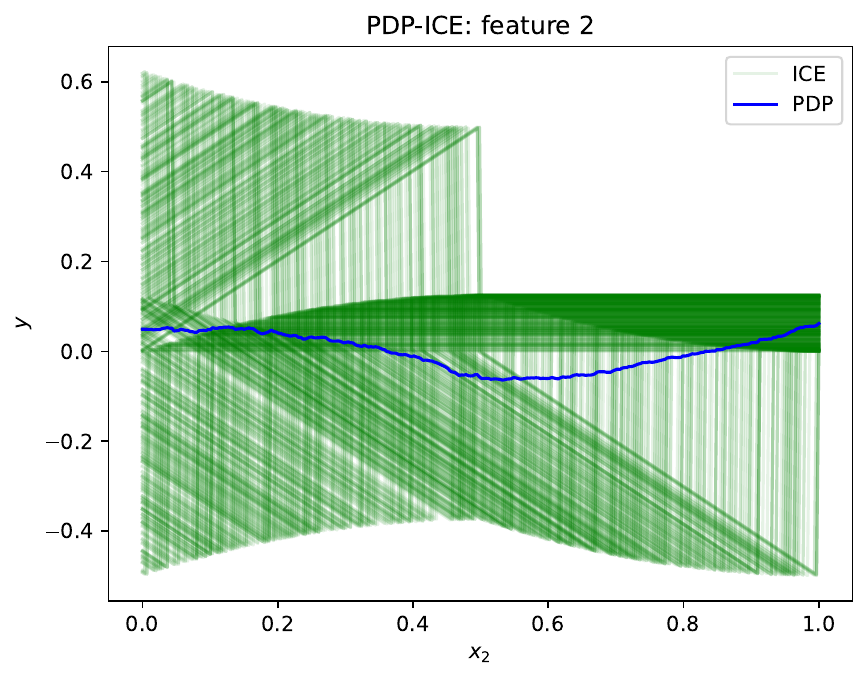}
    \includegraphics[width=0.3\linewidth]{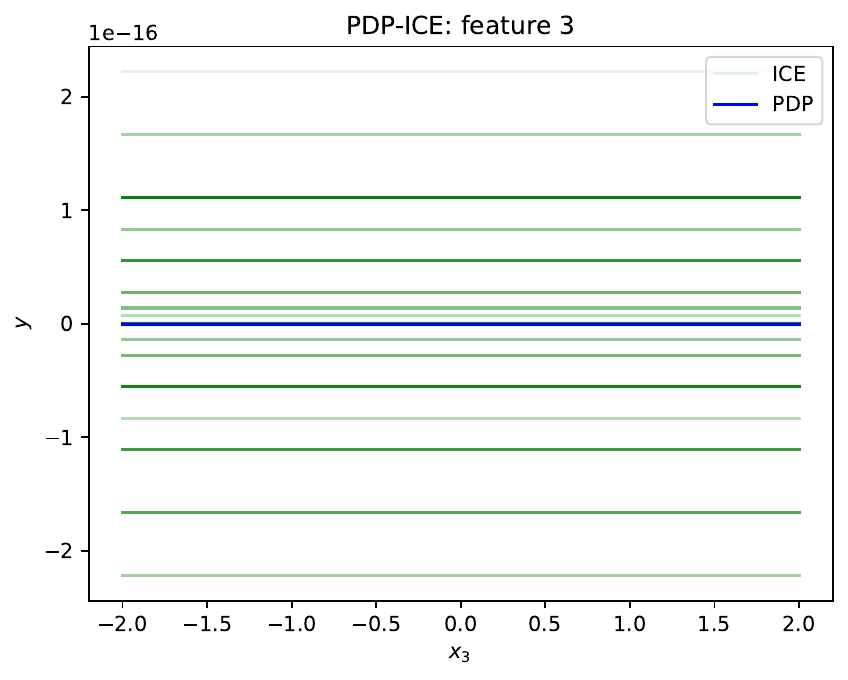}
    \caption{Case (a)}
\end{figure}

\paragraph{Ground truth for case (b)}

In this case, the weights are $a_1 = 2$ and $a_2 = \frac{1}{2}$ and there is no interaction
term $\alpha=0$.
Therefore:

\begin{equation}
  \label{eq:case-b}
  f(\mathbf{x}) = f_1(\xb) \mathbbm{1}_{f_1(\xb) \leq \frac{1}{2}} + (1 - f_1(\xb)) \mathbbm{1}_{\frac{1}{2} < f_1(\xb) < 1}
\end{equation}
where $f_1(\xb) = 2 x_1 + \frac{1}{2} x_2$.
As in case (a), we use again the fact that $x_1 \approx x_2$,
to compute the ground truth feature effect:
\begin{align}
    \label{eq:case-b-app}
    f^{\mathtt{GT}}(x_1) &= 2 x_1 \mathbbm{1}_{0 \leq x_1 \leq \frac{1}{5}} + (\frac{2}{5} - 2 x_1) \mathbbm{1}_{\frac{1}{4} < x_1 < \frac{2}{5}} \\
    f^{\mathtt{GT}}(x_2) &= 2 x_2 \mathbbm{1}_{0 \leq x_2 \leq \frac{1}{5}} + (\frac{2}{5} - 2 x_2) \mathbbm{1}_{\frac{1}{4} < x_2 < \frac{2}{5}} \\
    f^{\mathtt{GT}}(x_3) &= 0
\end{align}
The heterogeneity is zero for all features because the heterogeneity is induced by the variability
of the interaction terms and, since, $x_1 \approx x_2$, the terms
$\mathbbm{1}_{f_1(\xb) \leq \frac{1}{2}}$ and $\mathbbm{1}_{\frac{1}{2} < f_1(\xb) \leq 1}$, do not vary.
\begin{figure}
    \label{fig:case-b}
    \includegraphics[width=0.3\linewidth]{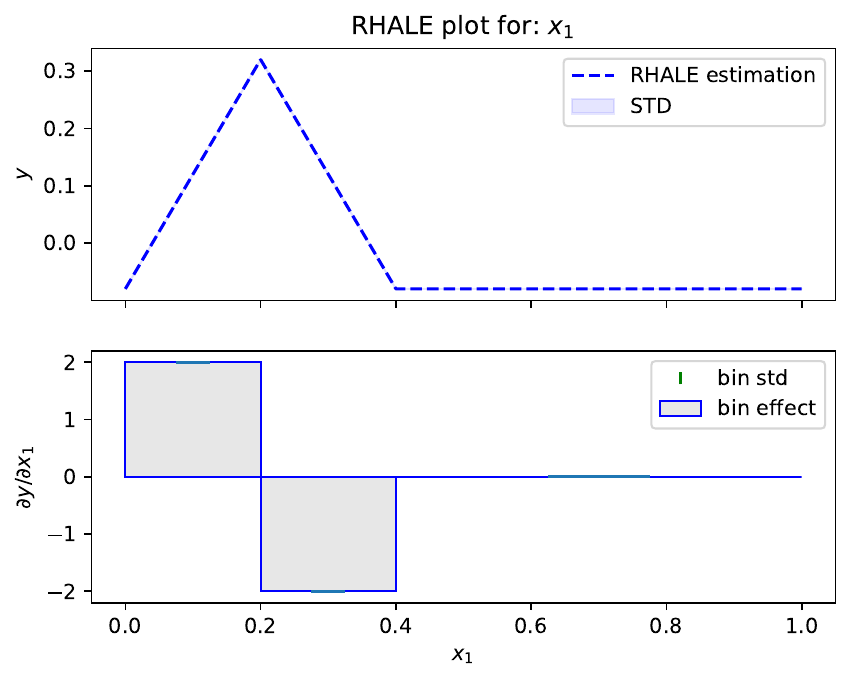}
    \includegraphics[width=0.3\linewidth]{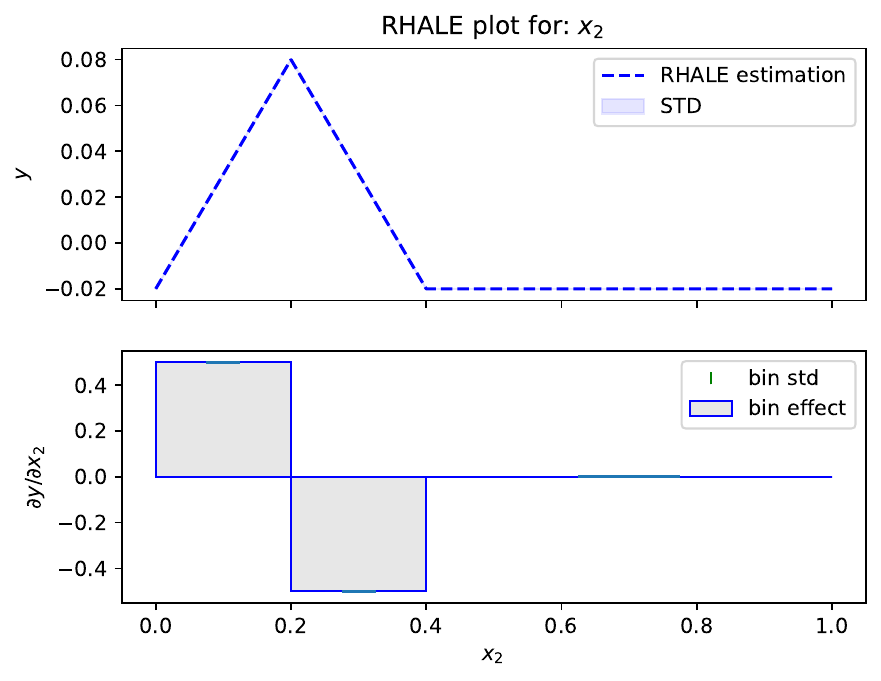}
    \includegraphics[width=0.3\linewidth]{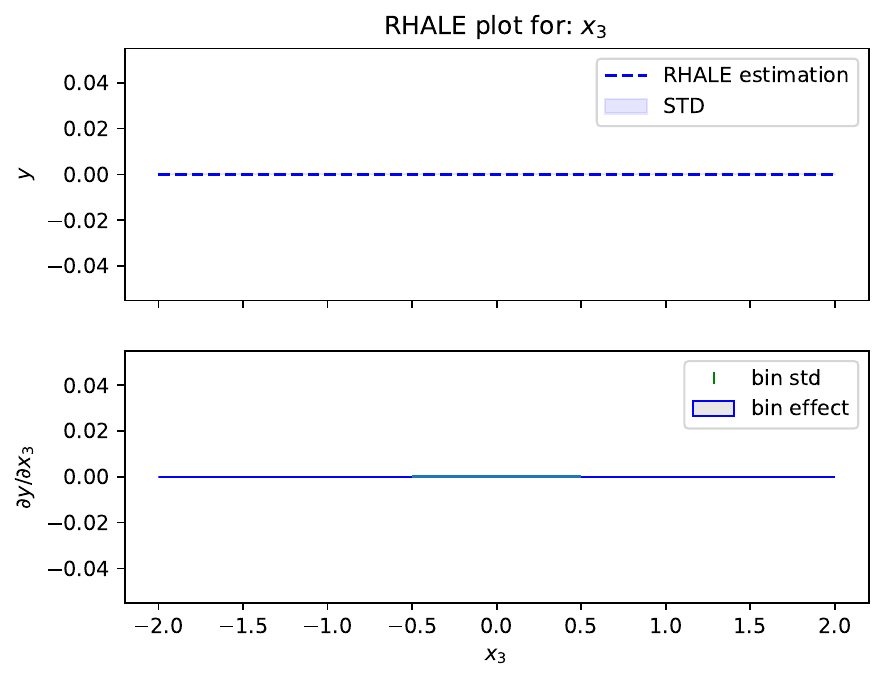}\\
    \includegraphics[width=0.3\linewidth]{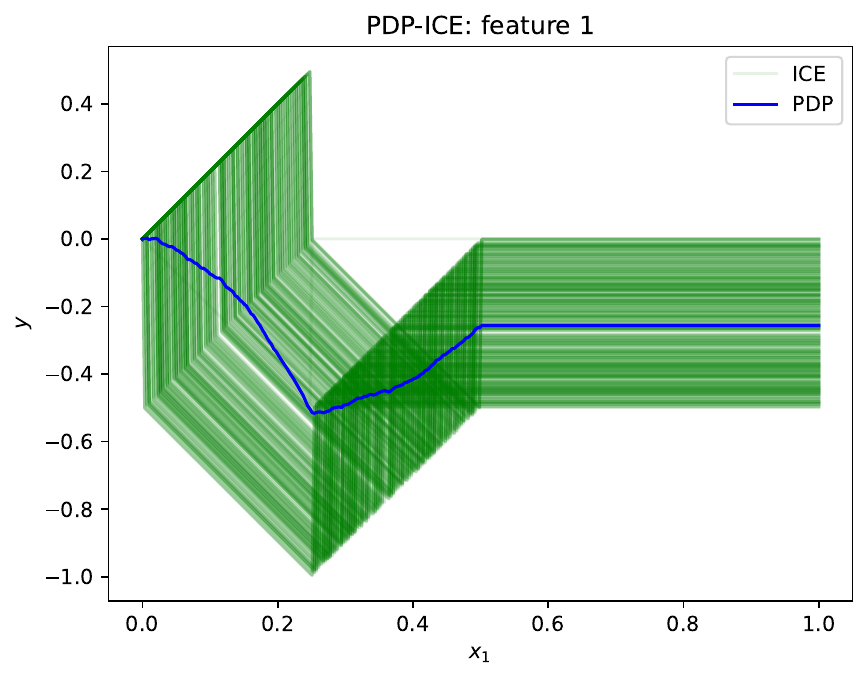}
    \includegraphics[width=0.3\linewidth]{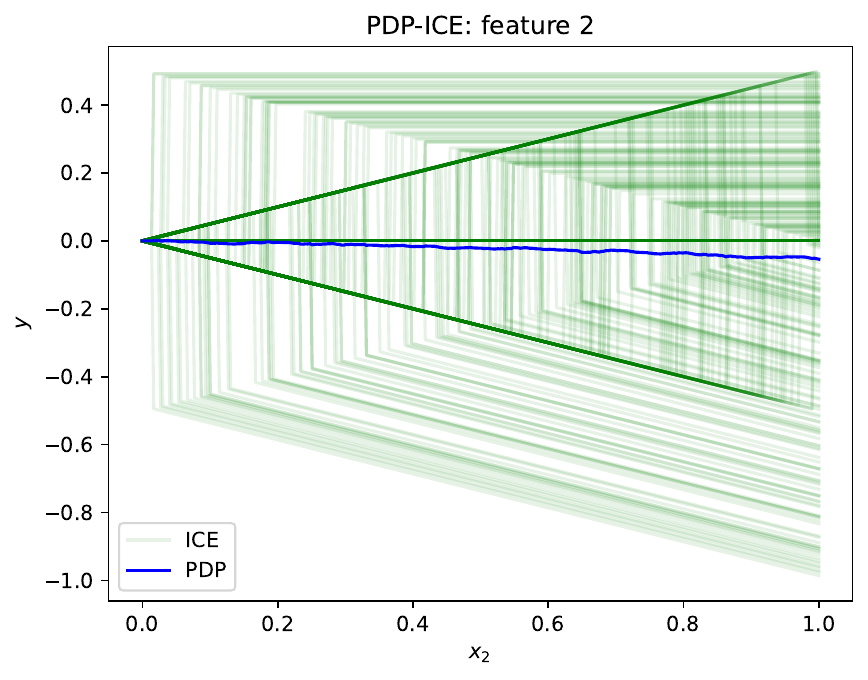}
    \includegraphics[width=0.3\linewidth]{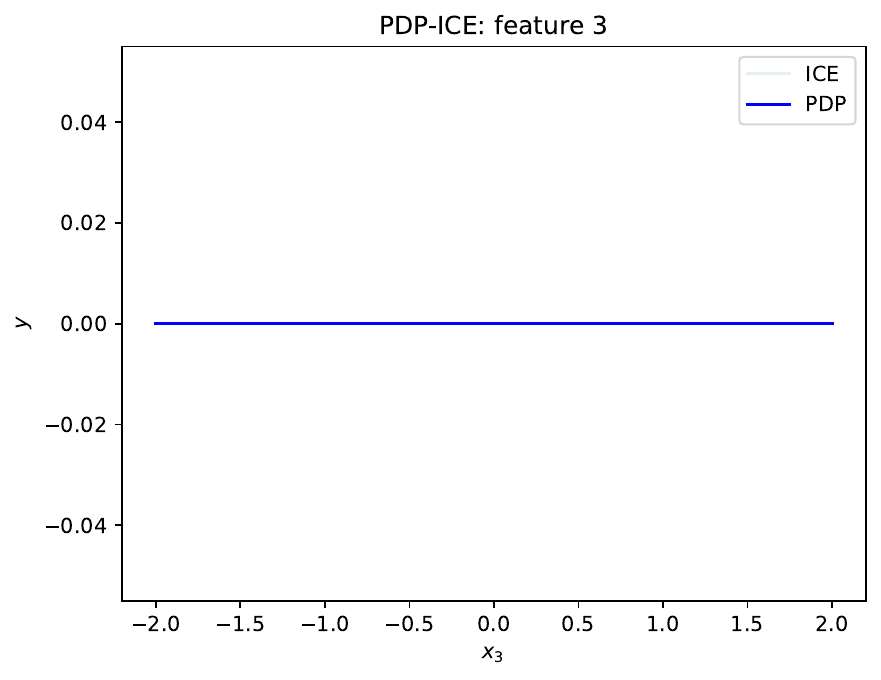}
    \caption{Case (b)}
\end{figure}

\paragraph{Ground truth for case (c)}

In this case, the weights are equal $a_1 = a_2 = 1$ and the is interaction
term is enabled $(\alpha=1)$.
Therefore:

\begin{equation}
  \label{eq:case-c}
  f(\mathbf{x}) = f_2(\xb) + f_1(\xb) \mathbbm{1}_{f_1(\xb) \leq \frac{1}{2}} + (1 - f_1(\xb)) \mathbbm{1}_{\frac{1}{2} < f_1(\xb) < 1}
\end{equation}
where $f_2(\xb) = x_1 x_3$ and $f_1(\xb) = x_1 + x_2$.
The feature effect of terms
$f_1(\xb) \mathbbm{1}_{f_1(\xb) \leq \frac{1}{2}} + (1 - f_1(\xb)) \mathbbm{1}_{\frac{1}{2} < f_1(\xb) < 1}$
are exactly the same with case (a).
The term $f_2(\xb) = x_1 x_2$.
For feature $x_1$ the effect is $\mathbb{E}_{x_3|x_1} [x_1 x_3] = x_1 \mathbb{E}_{x_3} [x_3] = 0$ and
for feature $x_2$ the effect is $\mathbb{E}_{x_1|x_3} [x_1 x_3] = x_3 \mathbb{E}_{x_1} [x_1] = 0.5 x_3$.
Therefore, the ground truth feature effect is:
\begin{align}
    \label{eq:case-c-app}
    f^{\mathtt{GT}}(x_1) &= x_1 \mathbbm{1}_{0 \leq x_1 \leq \frac{1}{4}} + \left ( \frac{1}{4} - x_1 \right ) \mathbbm{1}_{\frac{1}{4} < x_1 < \frac{1}{2}} \\
    f^{\mathtt{GT}}(x_2) &= x_2 \mathbbm{1}_{0 \leq x_2 \leq \frac{1}{4}} + \left ( \frac{1}{4} - x_2 \right ) \mathbbm{1}_{\frac{1}{4} < x_2 < \frac{1}{2}} \\
    f^{\mathtt{GT}}(x_3) &= \frac{1}{2} x_3
\end{align}
For the same reason with cases (a) and (b), the terms
$\mathbbm{1}_{f_1(\xb) \leq \frac{1}{2}}$ and $\mathbbm{1}_{\frac{1}{2} < f_1(\xb) \leq 1}$,
do not introduce heterogeneity.
Since $x_1, x_2$ are independent the effect of $x_1 x_3$ varies.
For feature $x_1$, it varies following the standard deviation of $x_3$, i.e. $\sigma_3=\frac{1}{2}$ and
for feature $x_3$, it varies following the standard deviation of $x_1$, i.e. $\sigma_1=\frac{1}{4}$.
\begin{figure}
    \label{fig:case-c}
    \includegraphics[width=0.3\linewidth]{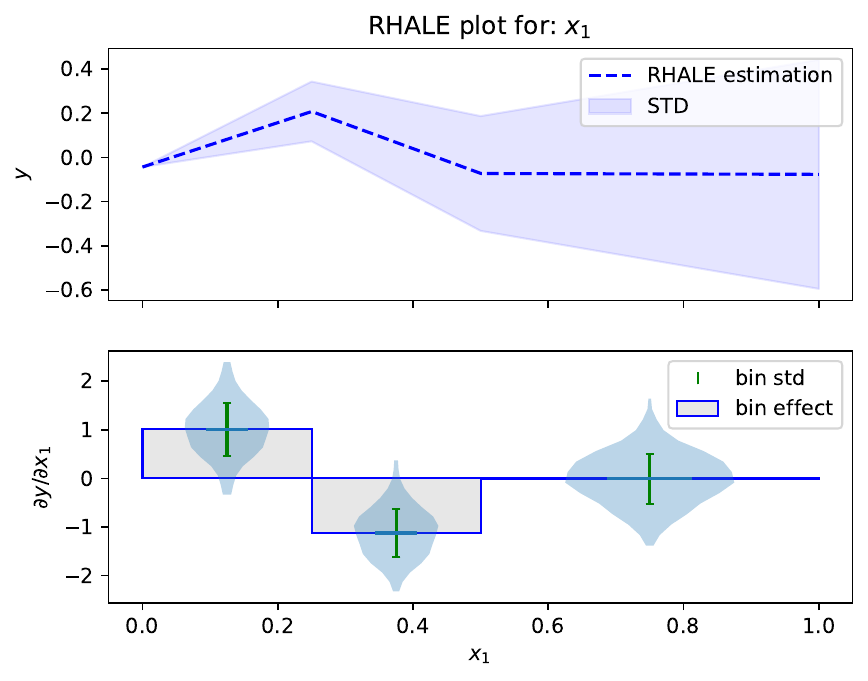}
    \includegraphics[width=0.3\linewidth]{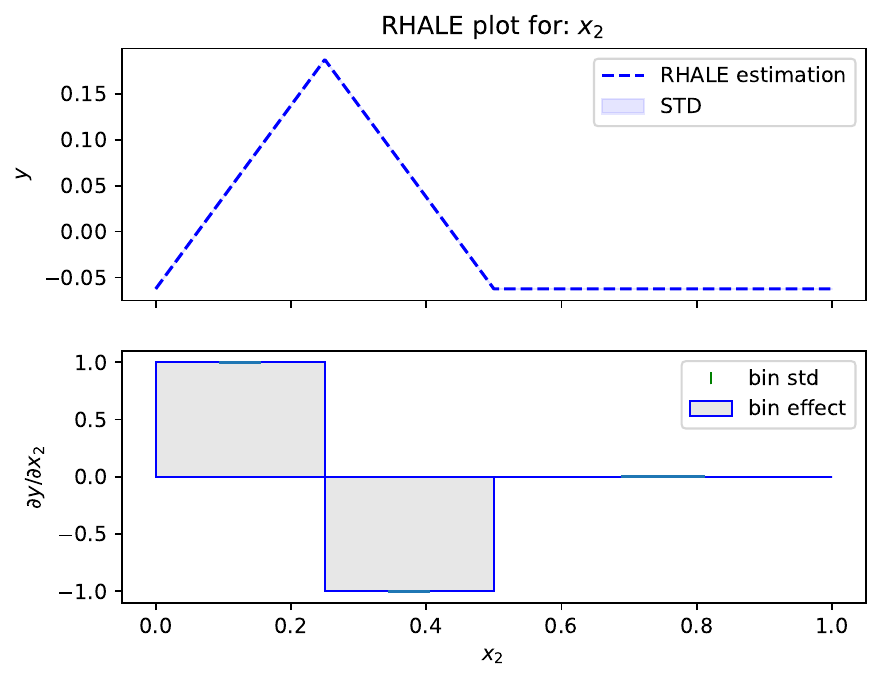}
    \includegraphics[width=0.3\linewidth]{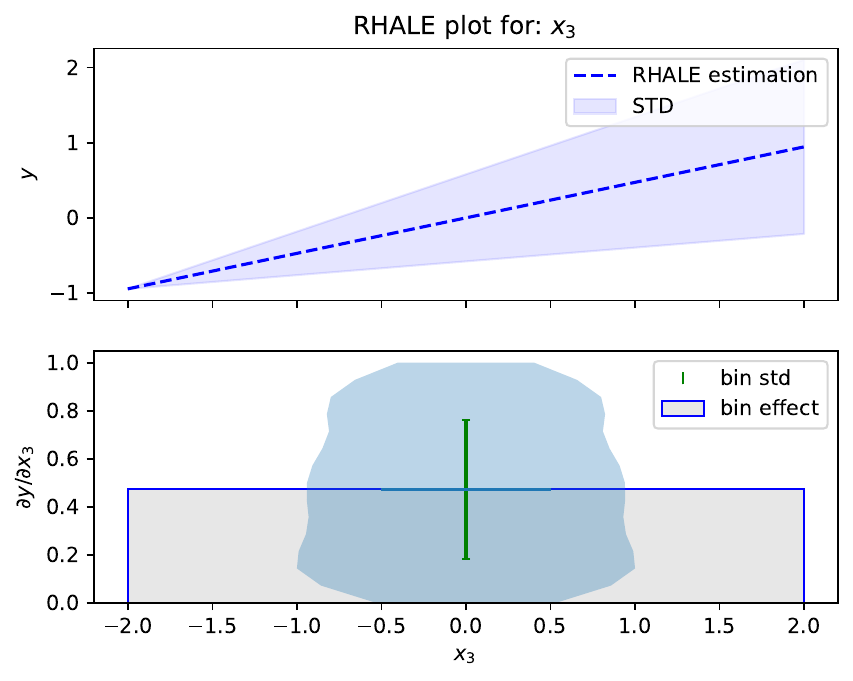}\\
    \includegraphics[width=0.3\linewidth]{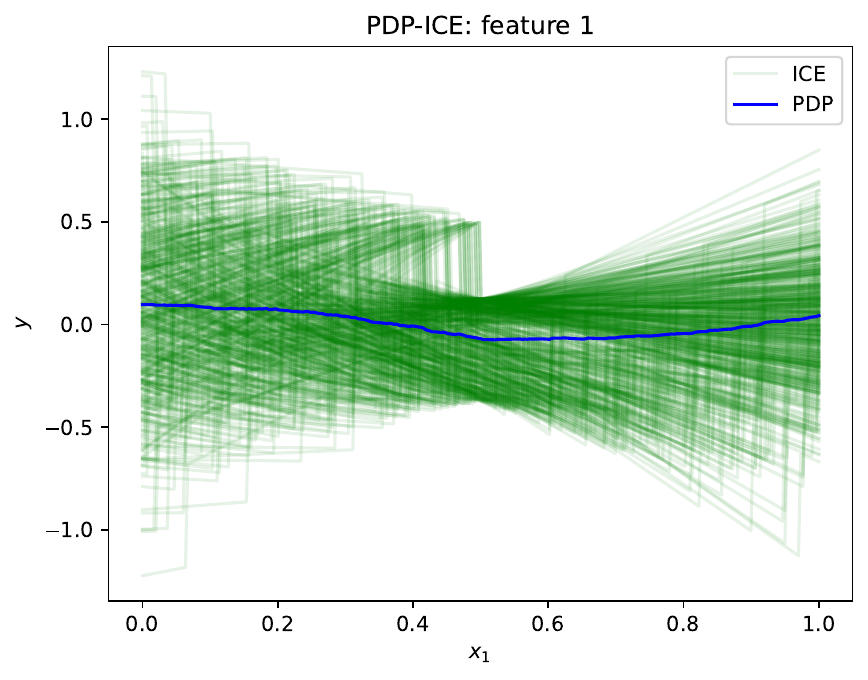}
    \includegraphics[width=0.3\linewidth]{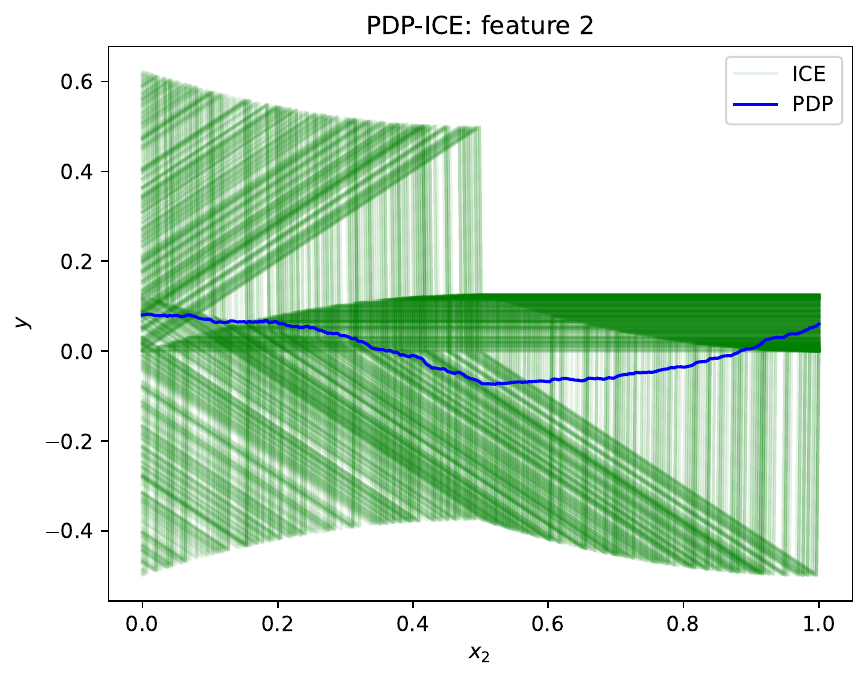}
    \includegraphics[width=0.3\linewidth]{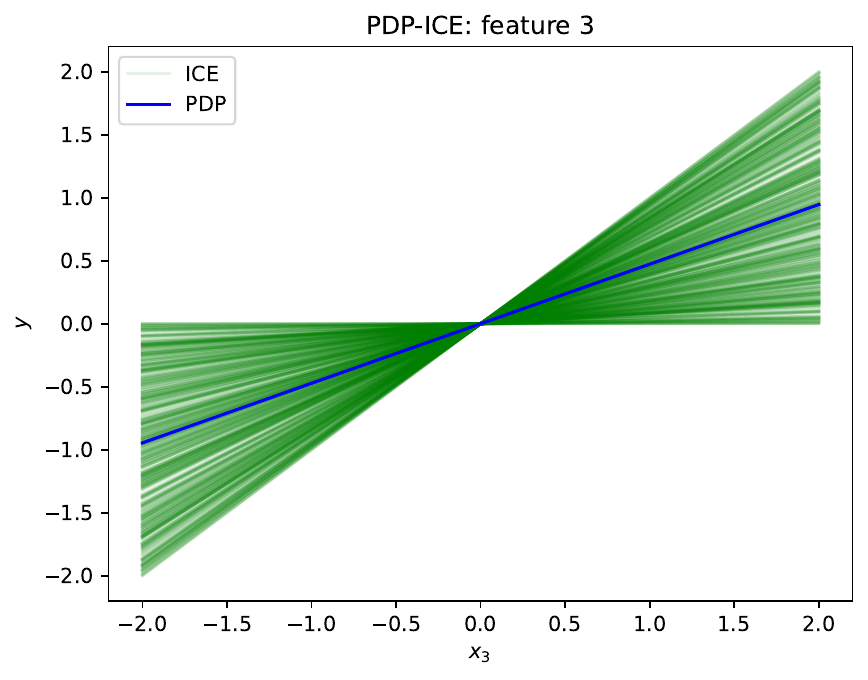}
    \caption{Case (b)}
\end{figure}

\paragraph{Conclusion.}

The example confirms our previous knowledge that PDP-ICE provide erroneous
effects in cases with correlated features.
The feature effect computed by PDP and the heterogeneity illustrated by
ICE are correct only for feature $x_3$,
because it is independent from the other features.
For features the correlated features $x_1, x_2$,
both PDP and ICE provide misleading explanations.
In contrast, RHALE handles well all cases, providing accurate estimations for the feature effects
and the heterogeneity.

\subsection{Real World Experiment}
\label{sec:real-world-experiment}

\begin{figure}[h]
  \centering
  \includegraphics[width=.43\textwidth]{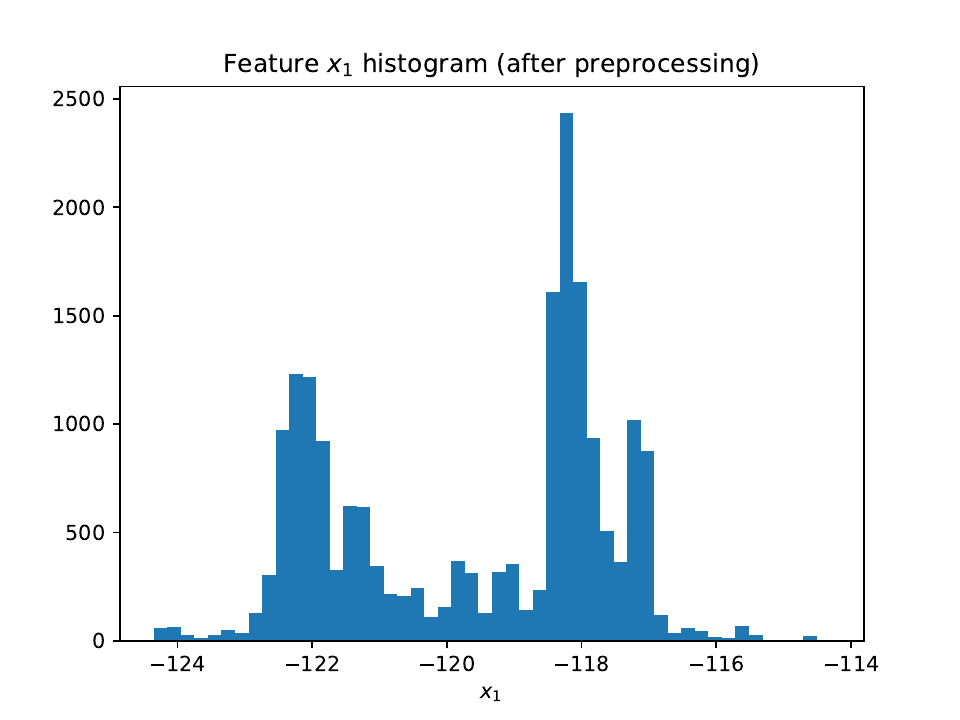}
  \includegraphics[width=.43\textwidth]{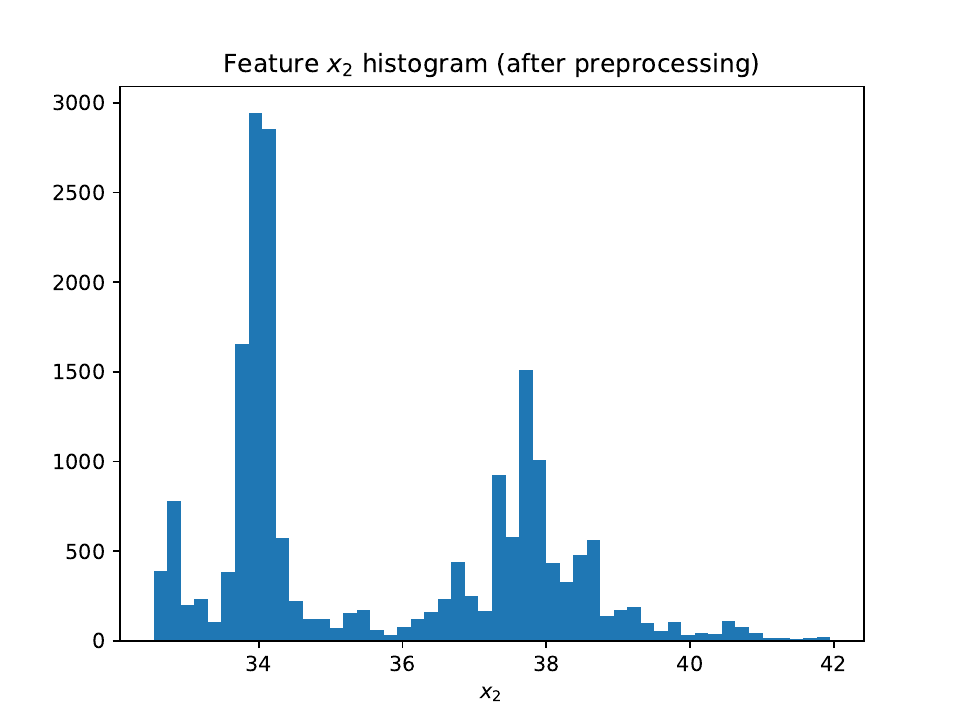}\\
  \includegraphics[width=.43\textwidth]{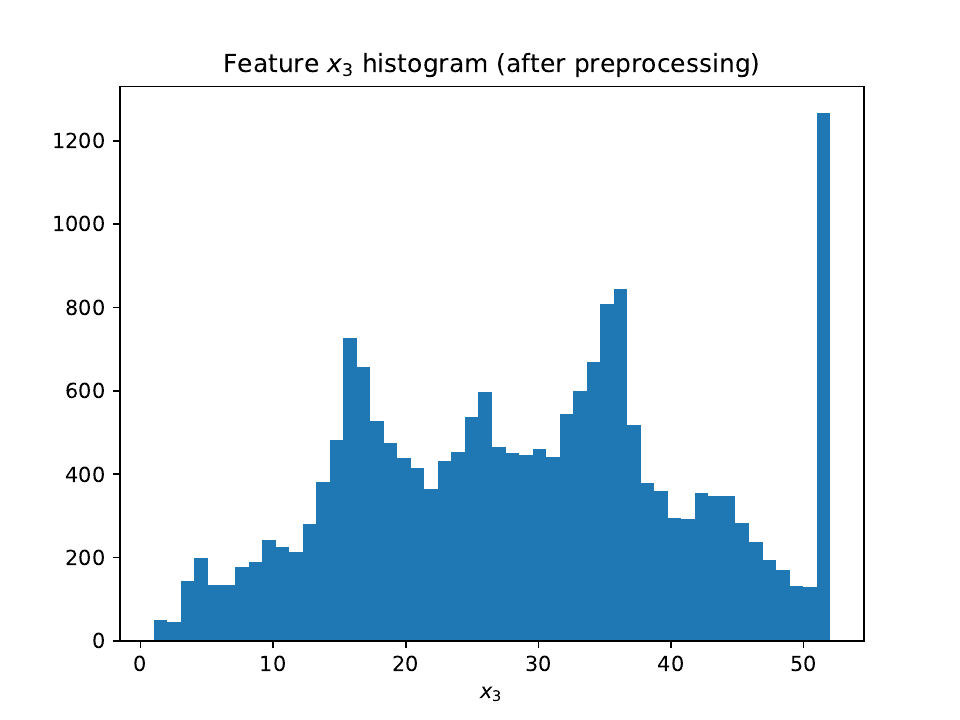}
  \includegraphics[width=.43\textwidth]{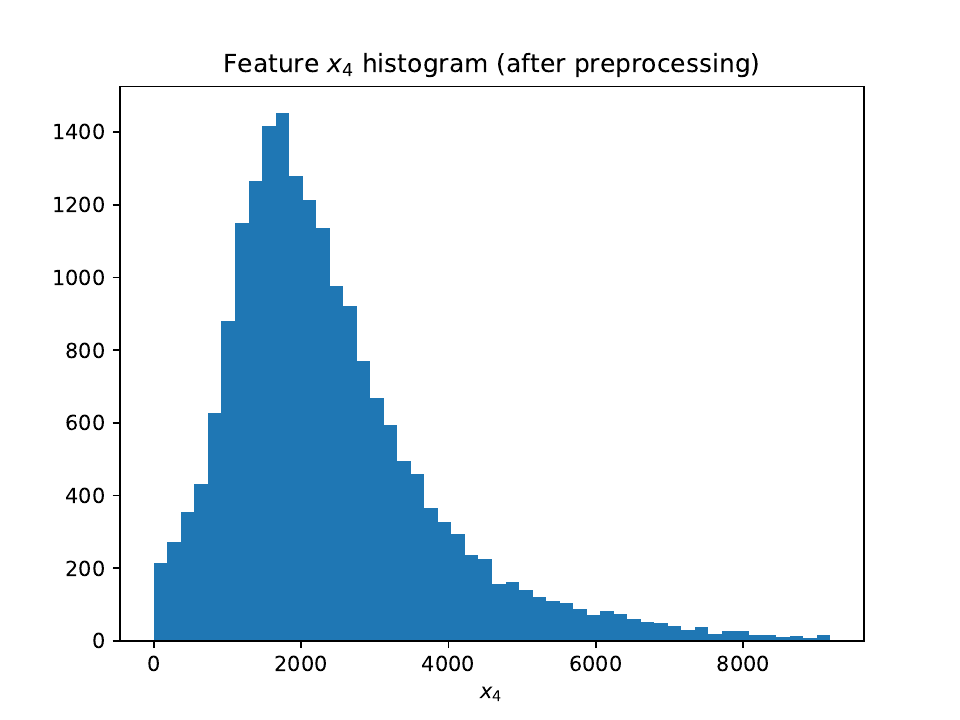}\\
  \includegraphics[width=.43\textwidth]{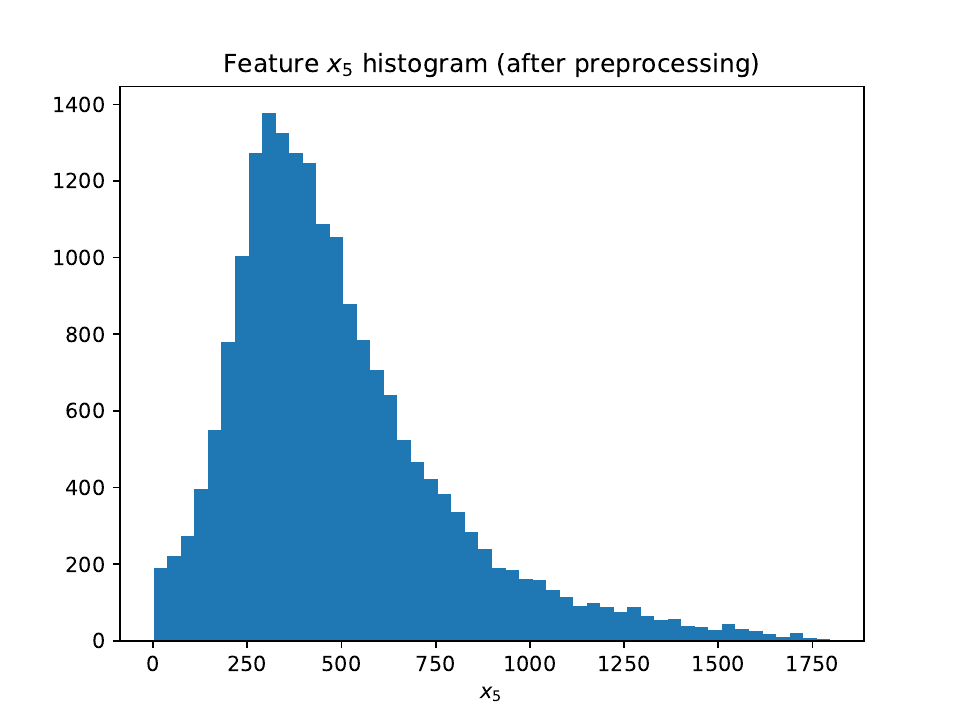}
  \includegraphics[width=.43\textwidth]{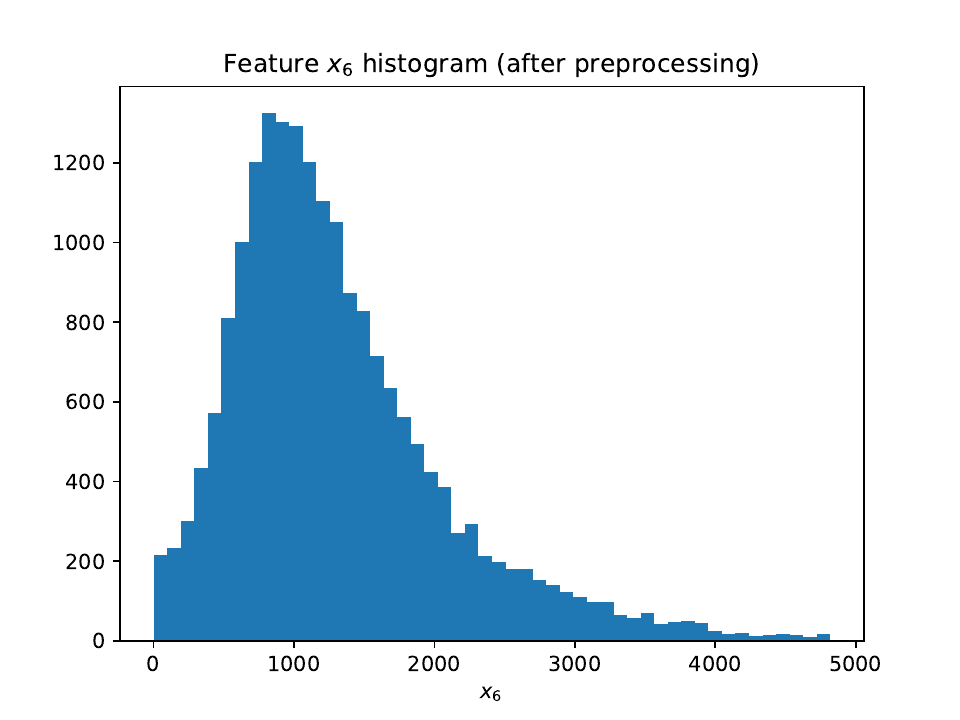}\\
  \includegraphics[width=.43\textwidth]{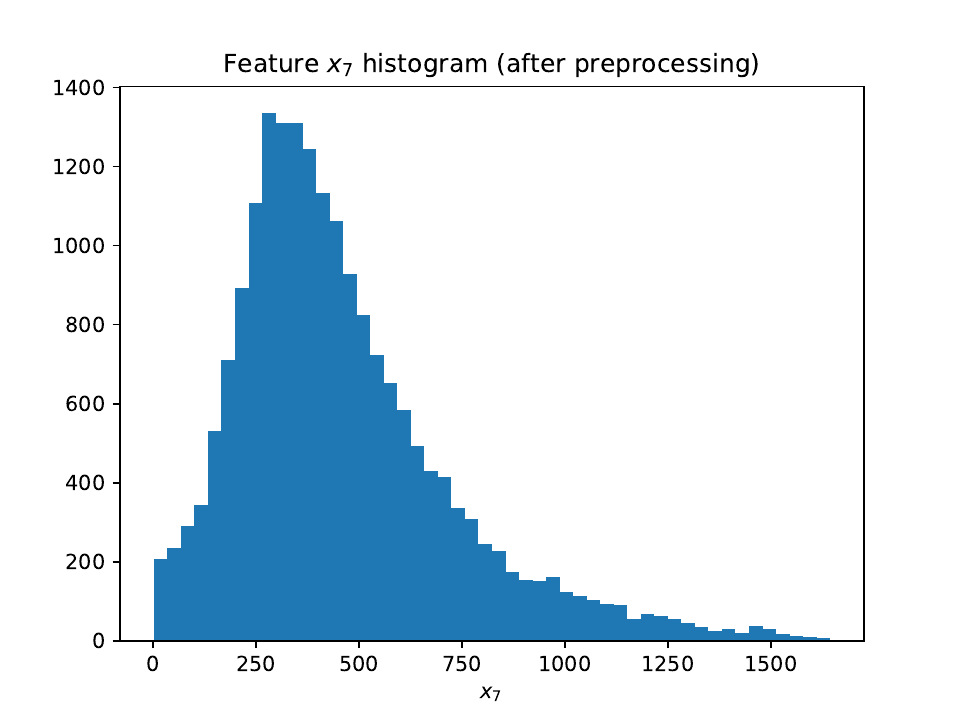}
  \includegraphics[width=.43\textwidth]{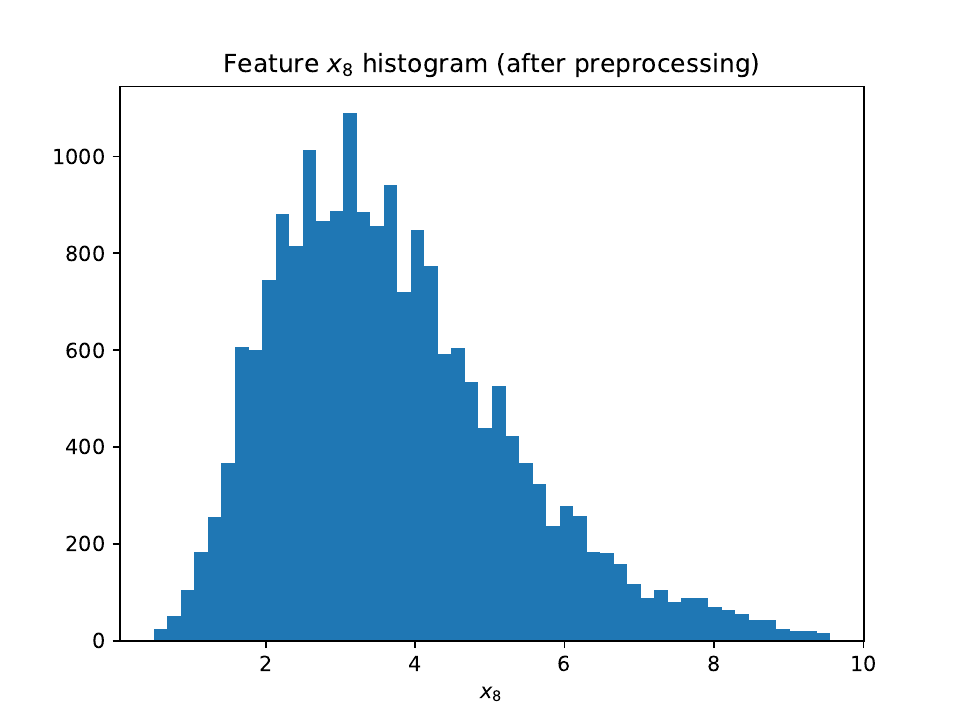}\\
  \caption{The Histogram of each feature in the California Housing Dataset.}
  \label{fig:feature-histograms}
\end{figure}

In this section, we provide further details on the real-world
example.
The real-world example uses the California Housing Dataset,
which contains 8 numerical features. We exclude instances with missing
or outlier values. If we denote as \(\mu_s\) (\(\sigma_s\)) the
average value (standard deviation) of the \(s\)-th feature, we
consider outliers the instances of the training set with any feature
value over three standard deviations from the mean, i.e.
\(|x_s^i - \mu_s| > \sigma_s\). This preprocessing step discards
\(884\) instances, and \(N=19549\) remain. We provide their
description with some basic descriptive statistics in
Table~\ref{tab:features-description} and their histogram in
Figure~\ref{fig:feature-histograms}.

\begin{figure}[h]
  \centering
  \includegraphics[width=.24\textwidth]{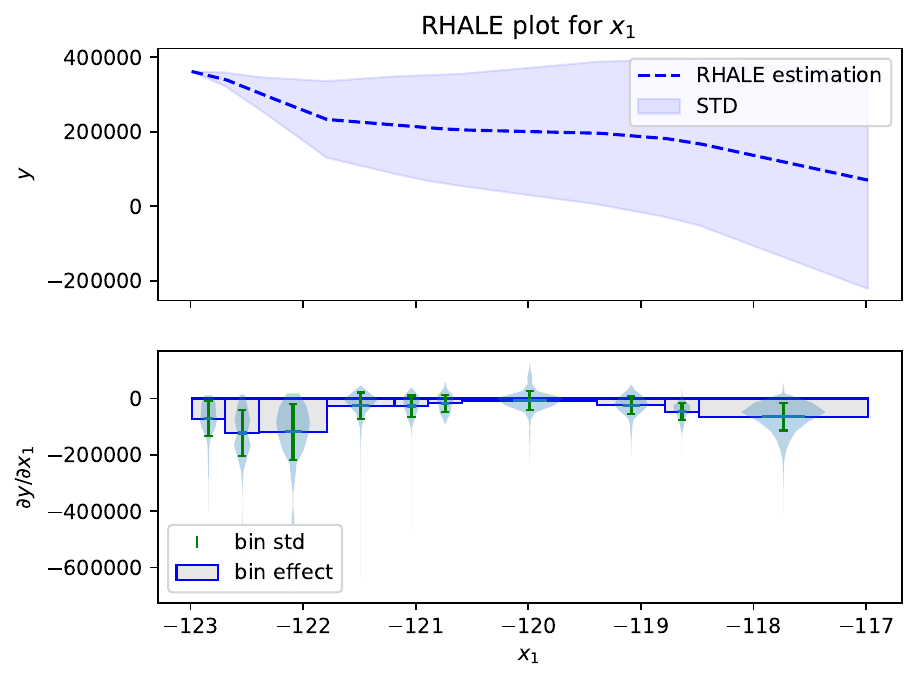}
  \includegraphics[width=.24\textwidth]{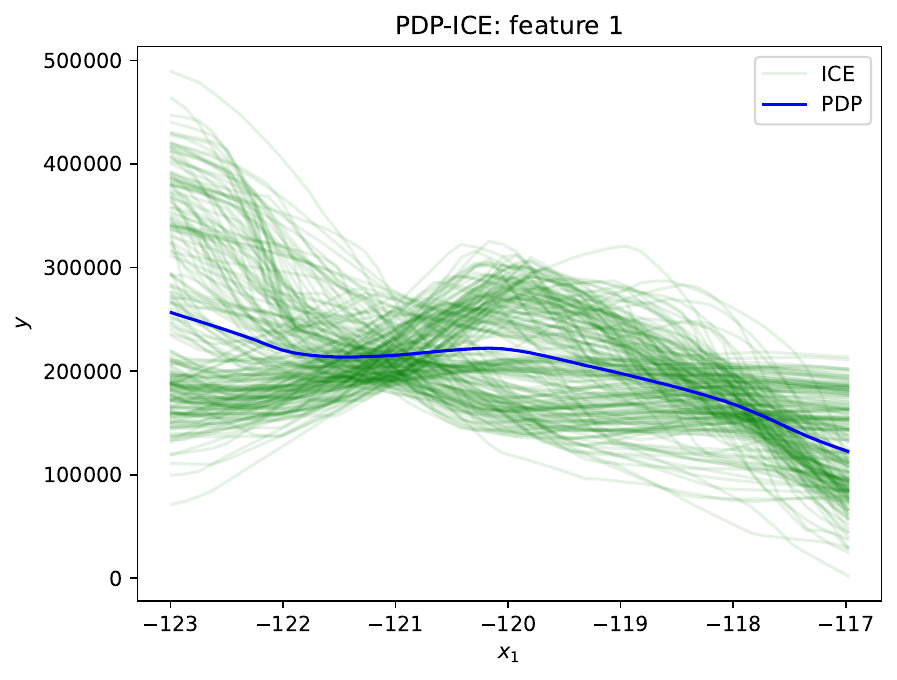}
  \includegraphics[width=.24\textwidth]{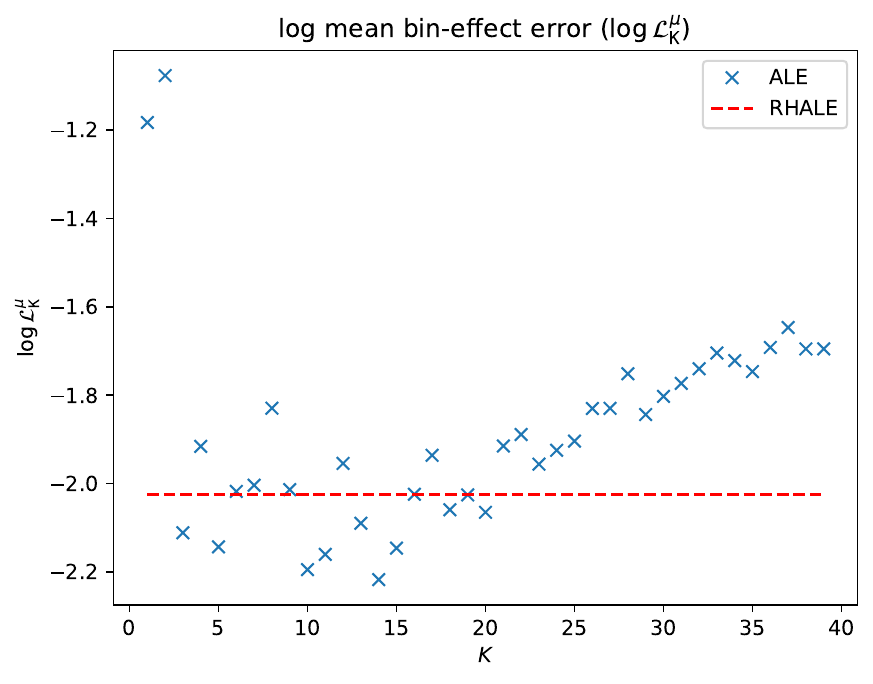}
  \includegraphics[width=.24\textwidth]{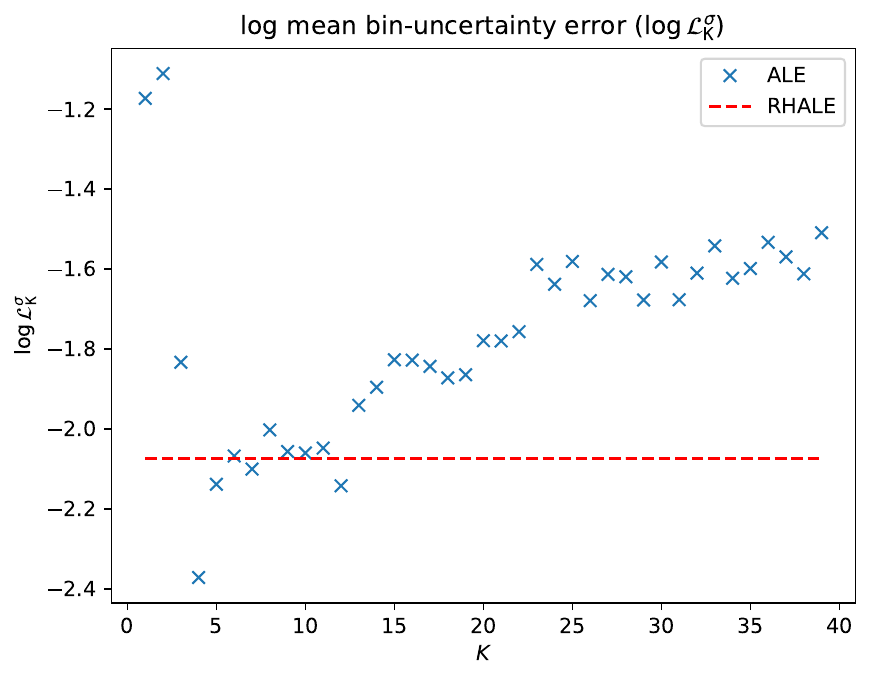}\\
  \includegraphics[width=.24\textwidth]{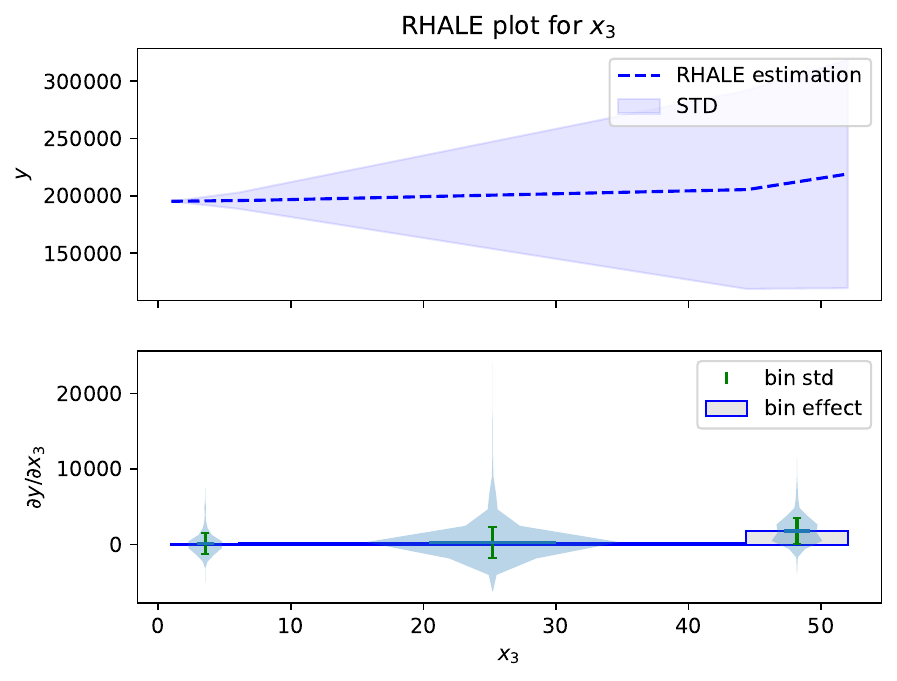}
  \includegraphics[width=.24\textwidth]{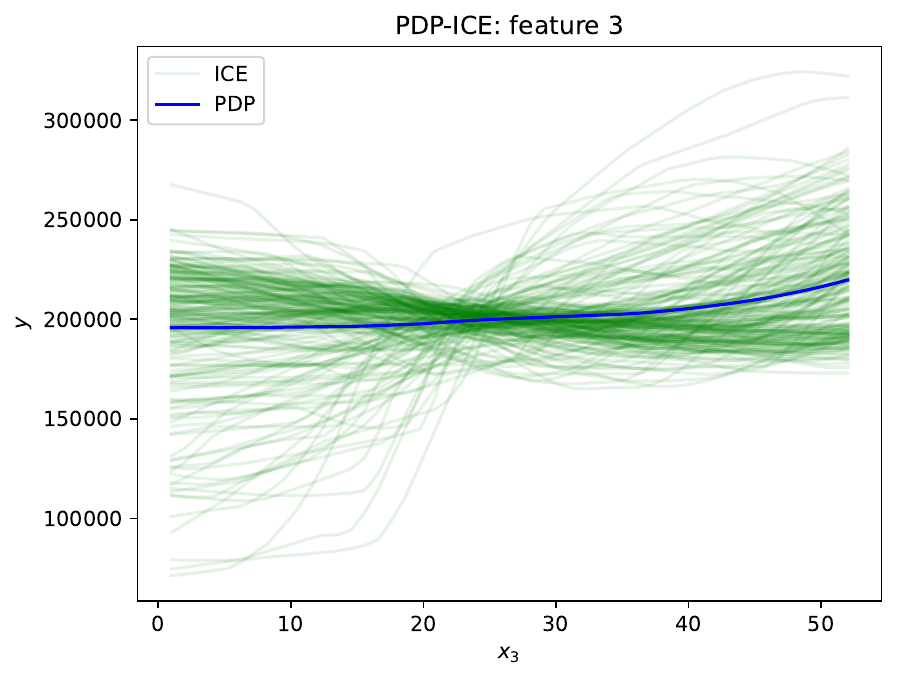}
  \includegraphics[width=.24\textwidth]{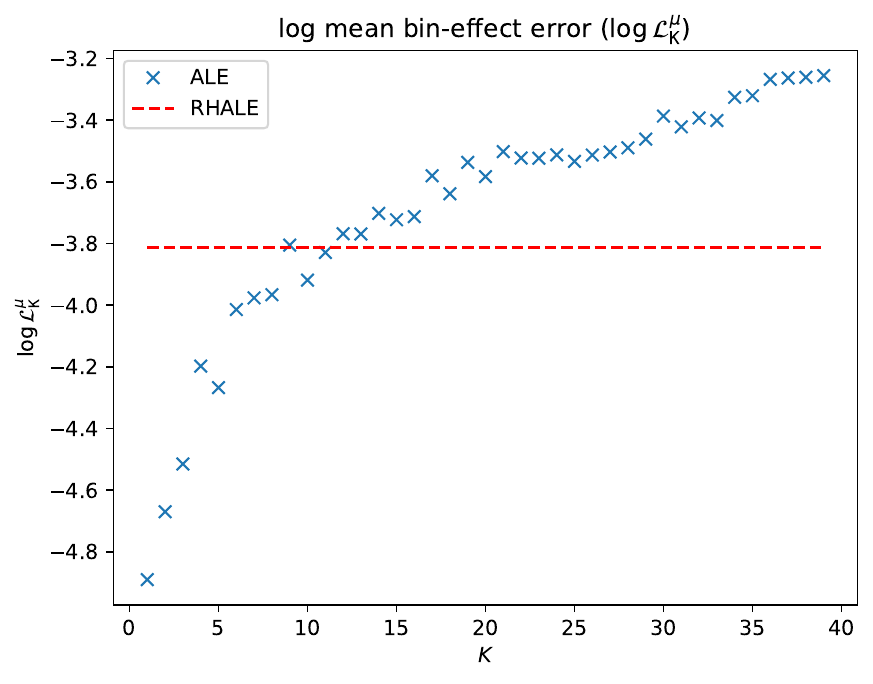}
  \includegraphics[width=.24\textwidth]{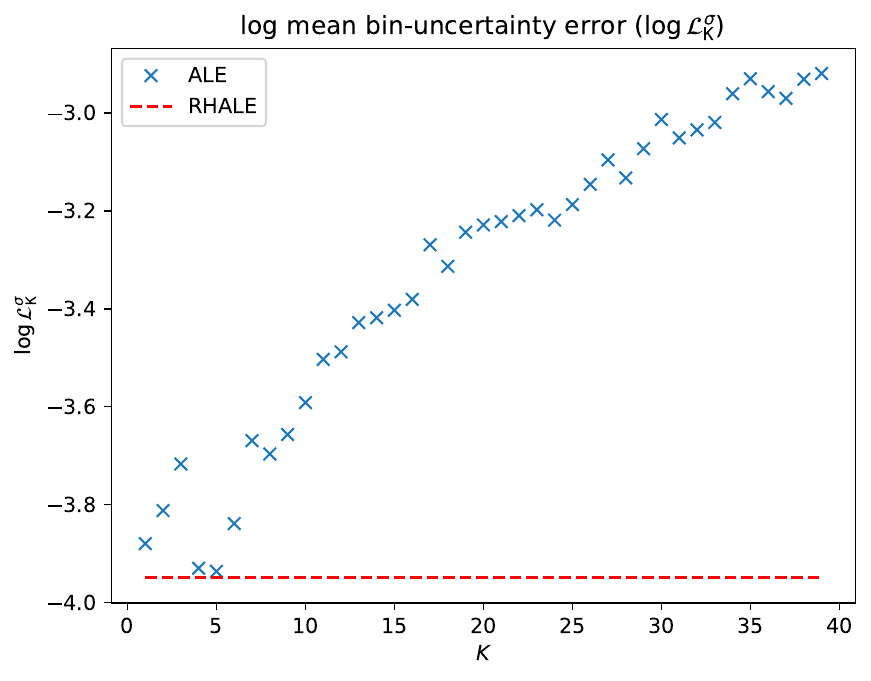}\\
  \includegraphics[width=.24\textwidth]{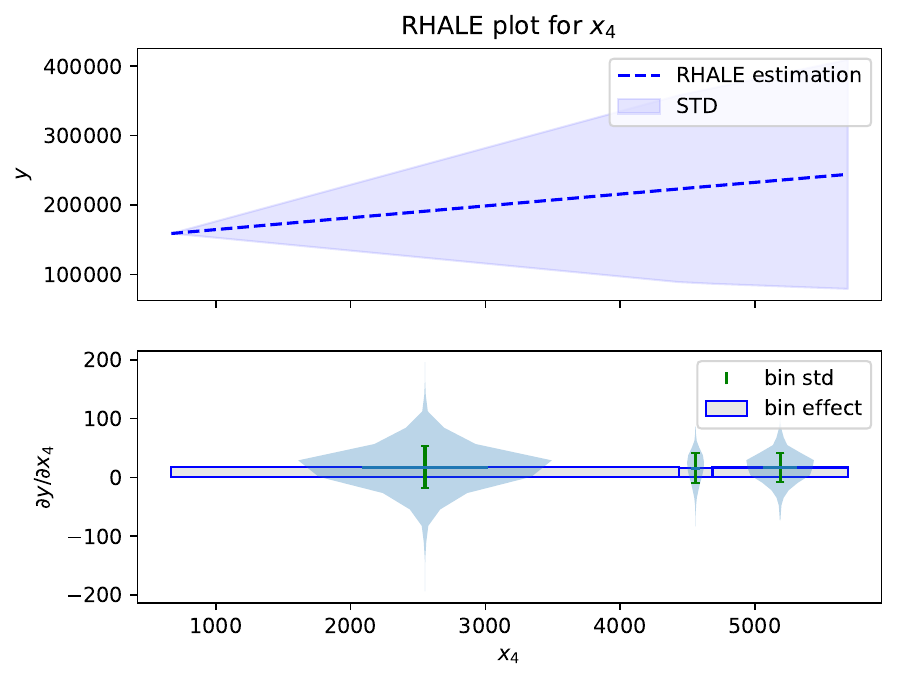}
  \includegraphics[width=.24\textwidth]{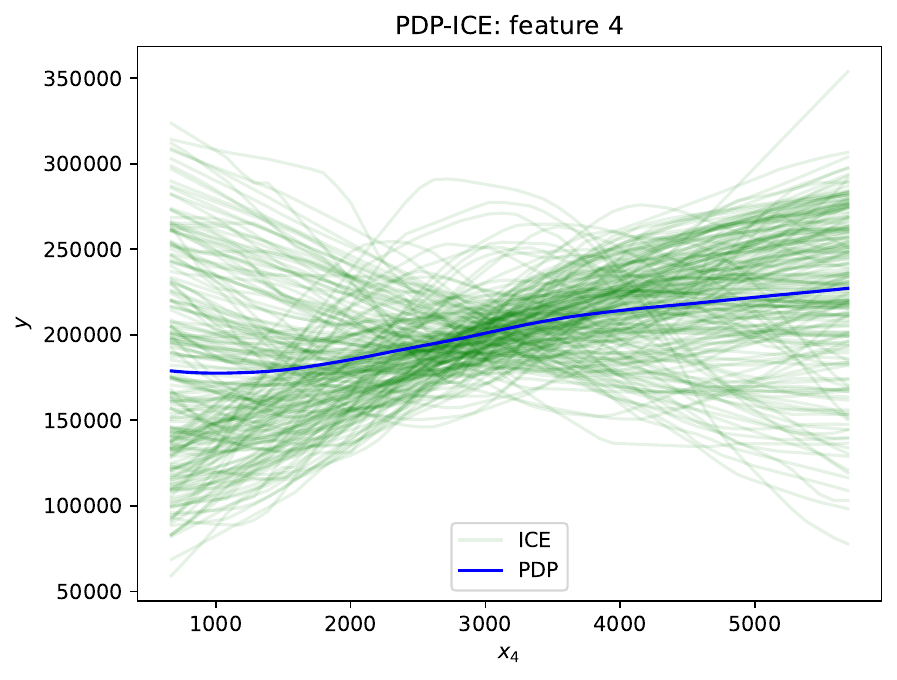}
  \includegraphics[width=.24\textwidth]{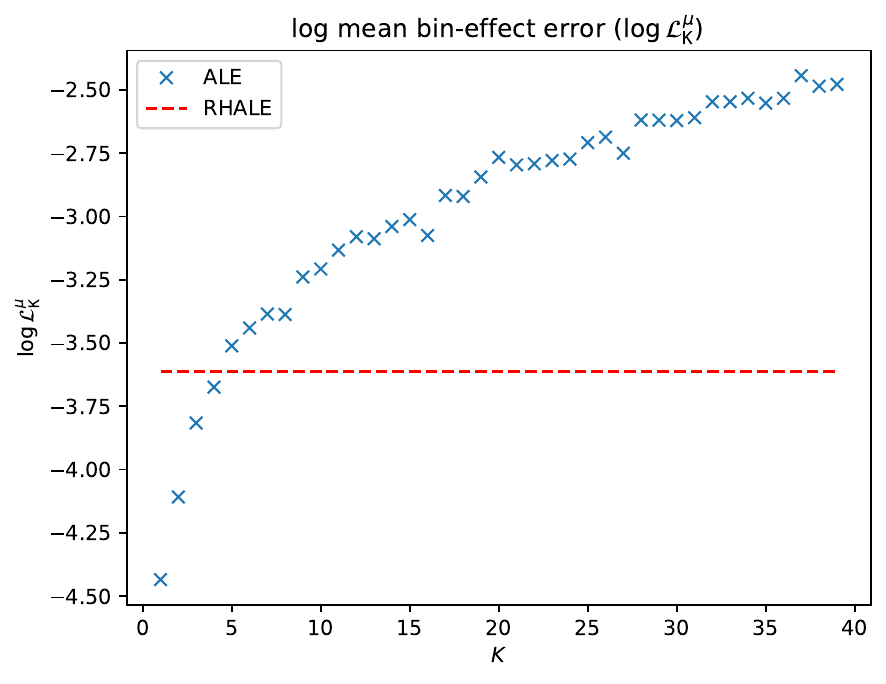}
  \includegraphics[width=.24\textwidth]{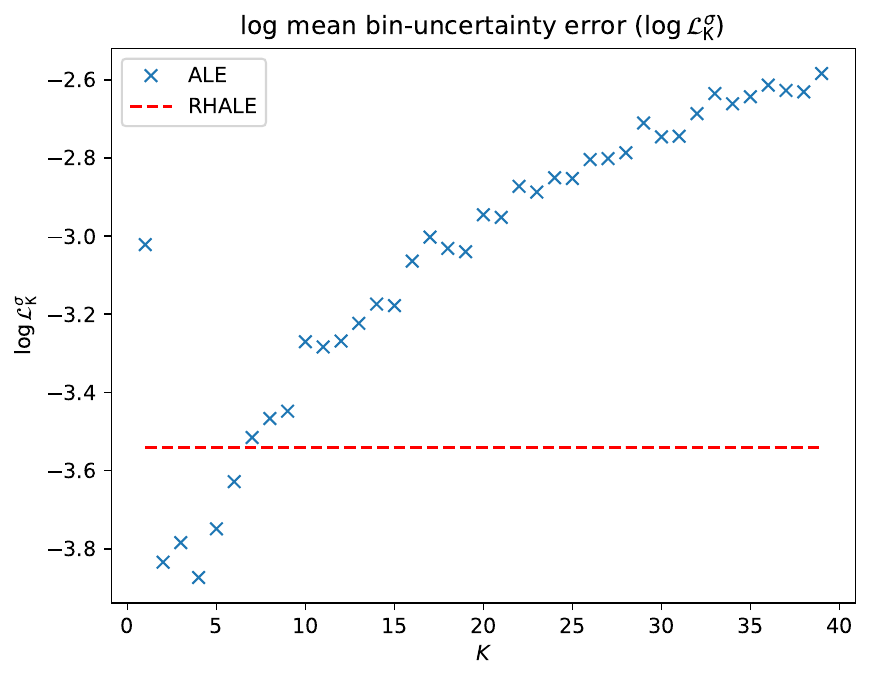}\\
  \includegraphics[width=.24\textwidth]{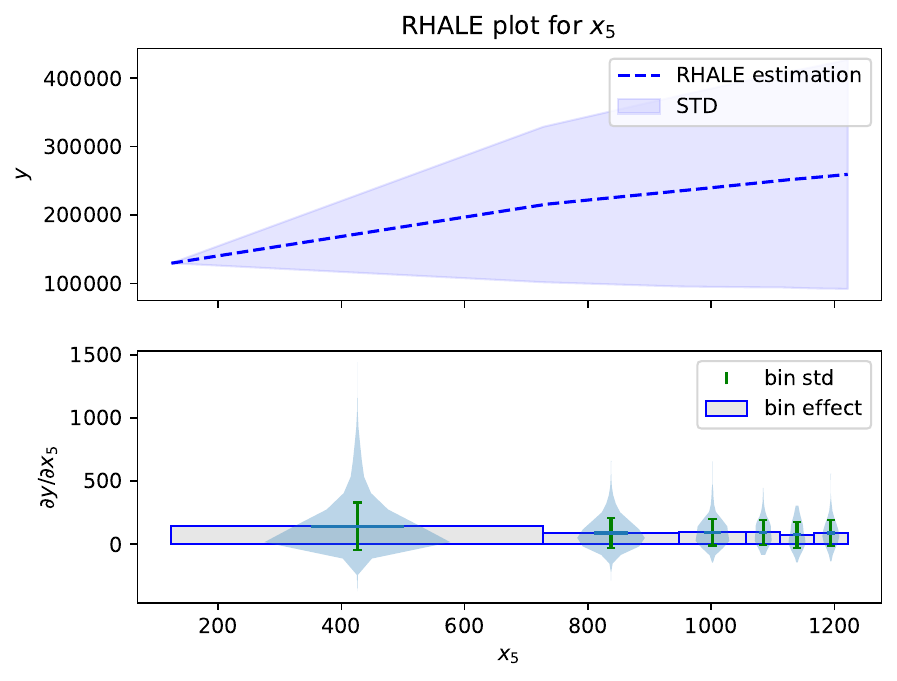}
  \includegraphics[width=.24\textwidth]{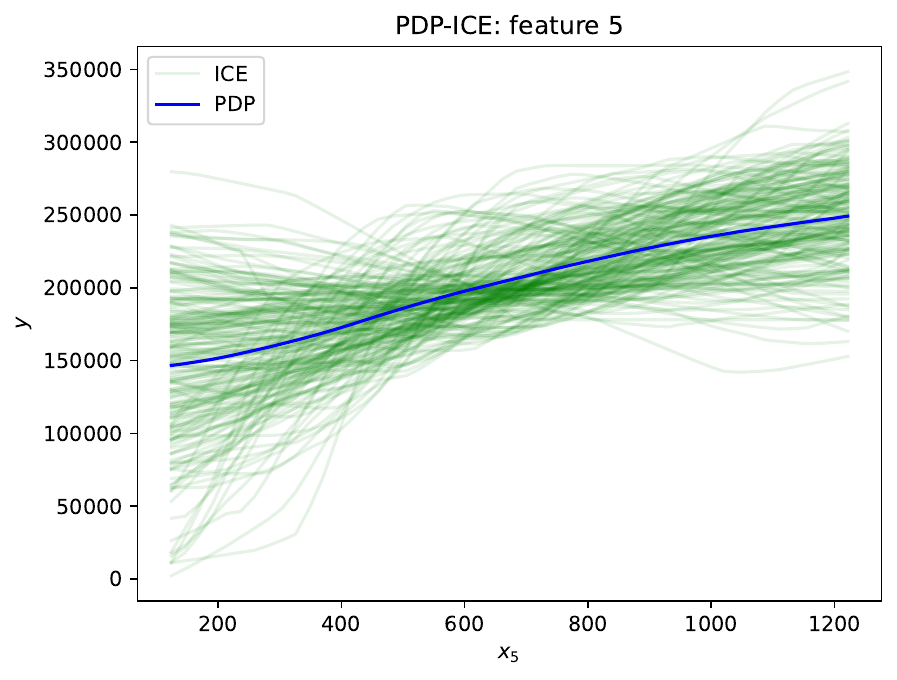}
  \includegraphics[width=.24\textwidth]{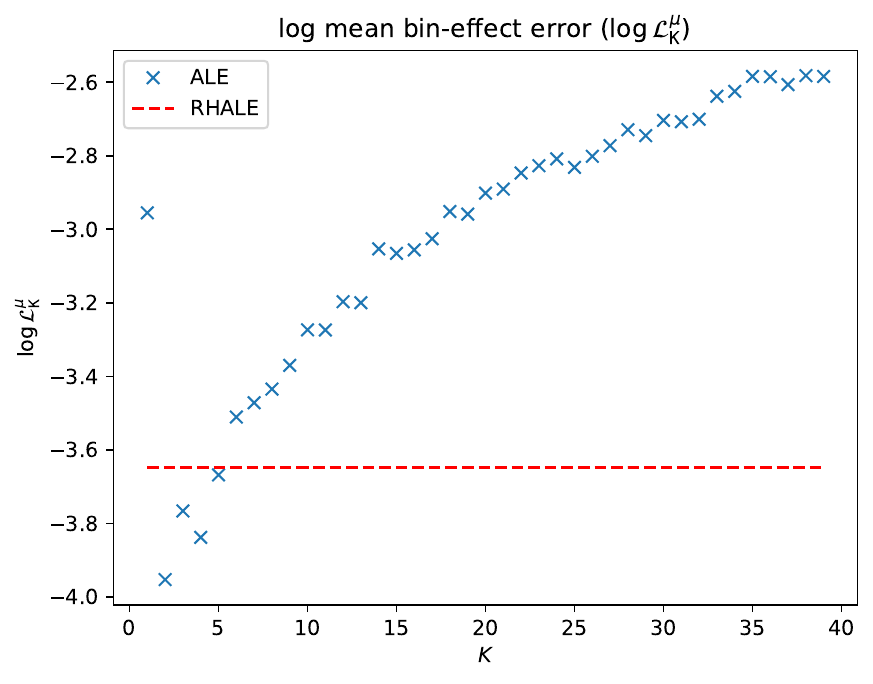}
  \includegraphics[width=.24\textwidth]{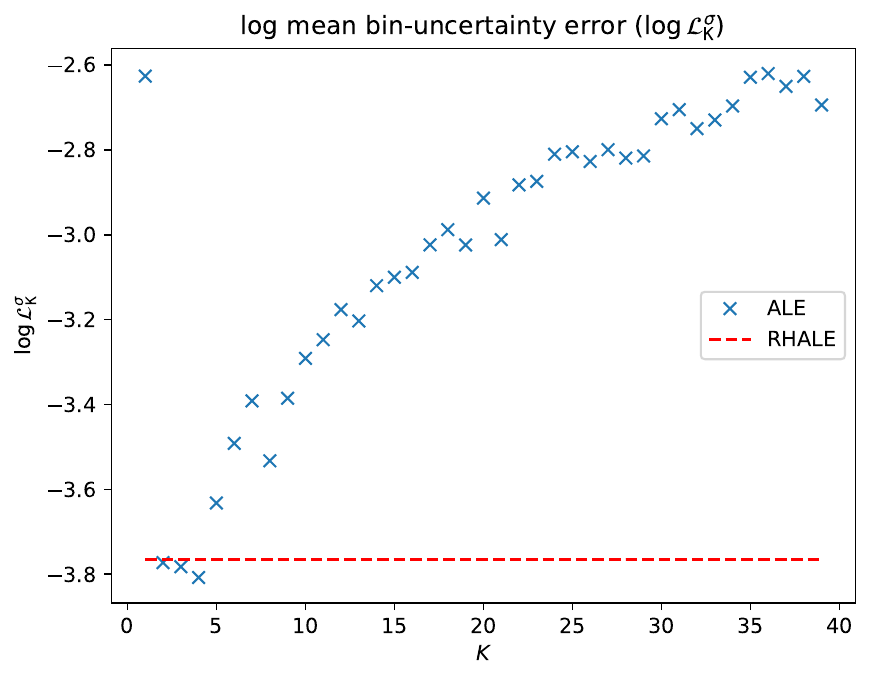}\\
  \includegraphics[width=.24\textwidth]{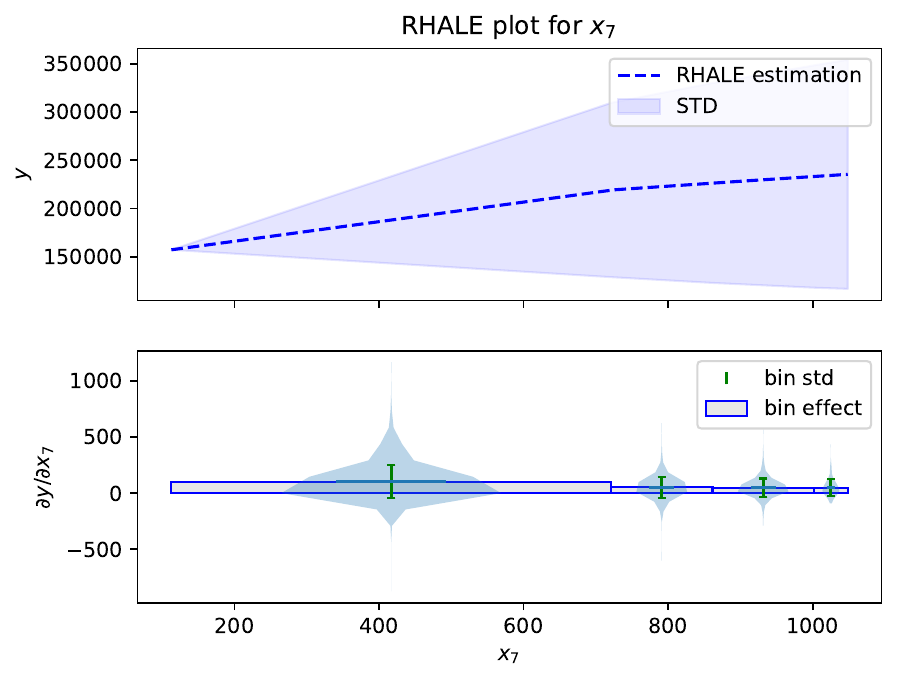}
  \includegraphics[width=.24\textwidth]{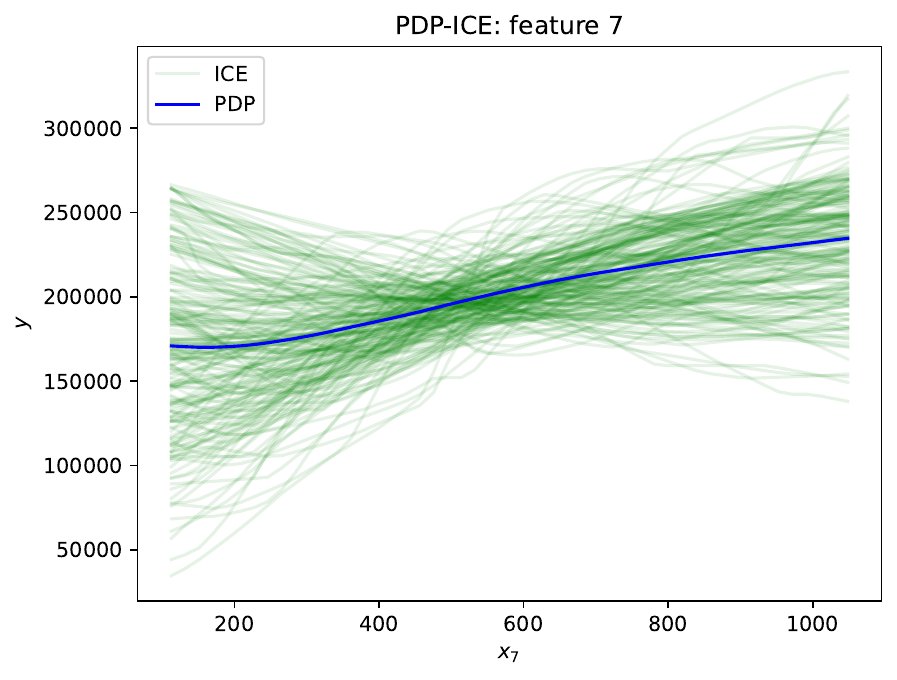}
  \includegraphics[width=.24\textwidth]{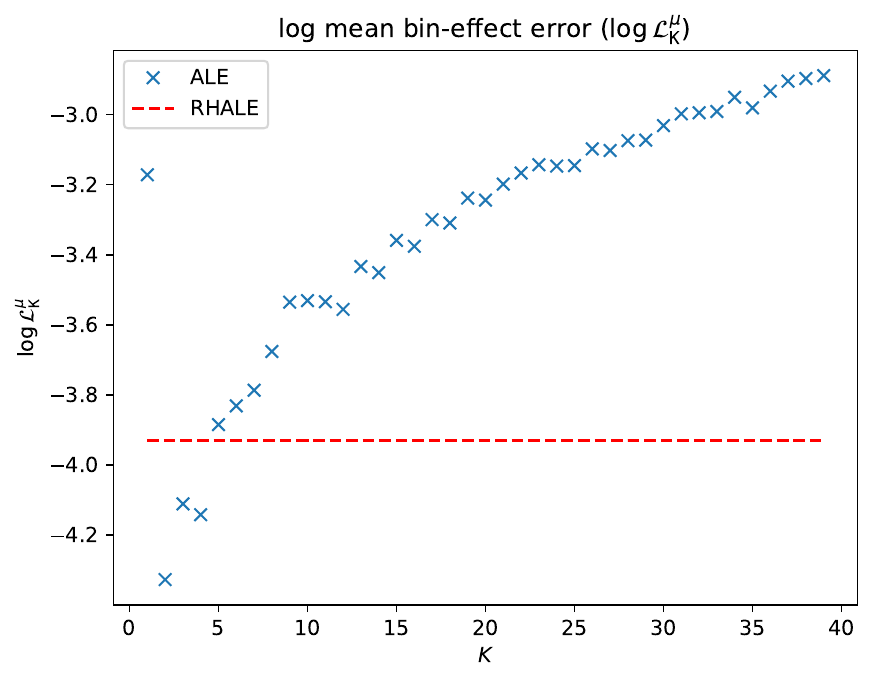}
  \includegraphics[width=.24\textwidth]{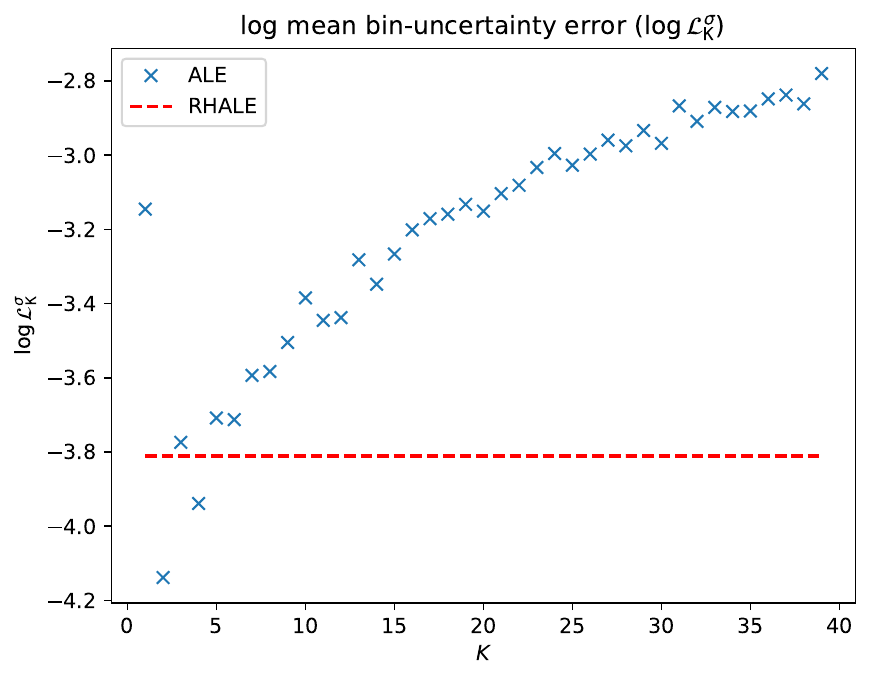}\\
  \caption{From left to right: (a) RHALE plot, (b) PDP-ICE plot, (c) RHALE vs fixed-size \(\mathcal{L}^{\mu}\) and (d) RHALE vs fixed-size \(\mathcal{L}^{\sigma}\). From top to bottom, features \(x_1, x_3, x_4, x_5, x_7, x_8\).}
  \label{fig:ex-real-1-app}
\end{figure}

In Figure 7 of the main paper, we provided the RHALE vs PDP-ICE plots
for features \(x_2\) (latitude), \(x_6\) (total number of people) and
\(x_8\) (median house value). In figure 8, we compared RHALE with
fixed-size approximation, for the same features. In
Figure~\ref{fig:ex-real-1-app}, we provide the same information for the
rest of the features; \(x_1\) (longitude), \(x_3\) (median age of
houses), \(x_4\) (total number of rooms), \(x_5\) (total number of
bedrooms) and \(x_7\) (total number of households). The observation of
these features leads us to similar conclusion. First, RHALE and
PDP-ICE plots compute similar effects and level of heterogeneity and
RHALE's approximation is (almost) as good as the best fixed-size
approximation. More specifically, we observe that RHALE's variable size
bin splitting correctly creates wide bins for features
\(x_3, x_4, x_5, x_7\), where the feature effect plot is (piecewise)
linear, while using narrow bins for feature \(x_2\) where the feature
effect is not linear.

\begin{table}
  \caption{Description of the features apparent in the California-Housing Dataset}
  \label{tab:features-description}
  \centering
  \begin{tabular}{ c|c|c|c|c|c| }
    \hline
    & Description & \(min\) & \(\max\) & \(\mu\) & \(\sigma\) \\
    \hline
    \(x_1\) & longitude & \(-124.35\) & \(-114.31\) & \(-119.58\) & \(2\) \\
    \(x_2\) & latitude  & \(32.54\) & \(41.95\) & \(35.65\) & \(2.14\) \\
    \(x_3\) & median age of houses & \(1\) & \(52\) & \(29.01\) & \(12.42\) \\
    \(x_4\) & total number of rooms & \(2\) & \(9179\) & \(2390.79\) & \(1433.83\) \\
    \(x_5\) & total number of bedrooms & \(2\) & \(1797\) & \(493.86\) & \(291\) \\
    \(x_6\) & total number of people & \(3\) & \(4818\) & \(1310.91\) & \(771.78\) \\
    \(x_7\) & total number of households & \(2\) & \(1644\) & \(460.3\) & \(267.34\) \\
    \(x_8\) & median income of households & \(0.5\) & \(9.56\) & \(3.72\) & \(1.60\) \\
    \hline
    \(y\) & median house value & \(14.999\) & \(500000\) & \(206864.41\) & \(115435.67\) \\
    \hline
  \end{tabular}
\end{table}

\bibliography{bibliography-supplement}